%% file: main.tex
\documentclass{article}

\PassOptionsToPackage{numbers, compress}{natbib}

\usepackage[main, final]{neurips_2026}

\title{{FlexTab}: A Flexible Encoder-Decoder Architecture for In-Context Learning Across Diverse Tabular Tasks}

\usepackage[textsize=tiny]{todonotes}

\usepackage[export]{adjustbox}

\author{%
Marek Polewczyk\textsuperscript{\normalfont$*\textrm{1}$}\quad%
Maximilian Schambach\textsuperscript{\normalfont$*\textrm{1}$}\quad%
Marco Spinaci\textsuperscript{\normalfont$*\textrm{2}$}\\
\textbf{Sam Thelin}\textsuperscript{\normalfont$\textrm{1}$}\quad%
\textbf{Johannes H\"ohne}\textsuperscript{\normalfont$\textrm{1}$}\\[2mm]
\textsuperscript{\normalfont$\textrm{1}$}SAP SE \quad \textsuperscript{\normalfont$\textrm{2}$}SAP France\\
\texttt{\{firstname.lastname\}@sap.com}
}

\usepackage{colortbl}
\usepackage[utf8]{inputenc} %
\usepackage[T1]{fontenc}    %
\usepackage{hyperref}       %
\hypersetup{colorlinks=true, linkcolor=black, citecolor=black, urlcolor=black}

\usepackage{url}            %
\usepackage{booktabs}       %
\usepackage{amsfonts}       %
\usepackage{nicefrac}       %
\usepackage{microtype}      %
\usepackage{xcolor}         %
\usepackage{enumitem}

\usepackage{graphicx}
\usepackage{subcaption}
\usepackage{float}

\usepackage{amsmath}
\usepackage{amssymb}
\usepackage{mathtools}
\usepackage{amsthm}

\usepackage[capitalize,noabbrev]{cleveref}

\usepackage{siunitx}
\usepackage{xspace}
\usepackage{multirow}

\theoremstyle{plain}

\theoremstyle{definition}

\theoremstyle{remark}

\usepackage{xspace}
\usepackage{xcolor}
\newcommand{\URL}[1]{\href{#1}{#1}}
\newcommand{\ROWT}{\texttt{[ROW]}\xspace}
\newcommand{\CLST}{\texttt{[CLS]}\xspace}
\newcommand{\TRGT}{\texttt{[MASK]}\xspace}
\definecolor{mynangrey}{rgb}{0.5, 0.5, 0.5}
\definecolor{mygrey}{rgb}{0.75, 0.75, 0.75}
\newcommand{\NAN}{\textcolor{mynangrey}{N/A}}

\NewDocumentCommand\NORM{om}{
  \IfNoValueTF{#1}
    {\ensuremath{\lVert #2 \rVert}}
    {\ensuremath{\lVert #2 \rVert_{#1}}}
}

\NewDocumentCommand\ABS{om}{
  \IfNoValueTF{#1}
    {\ensuremath{\lvert #2 \rvert}}
    {\ensuremath{\lvert #2 \rvert_{#1}}}
}

\NewDocumentCommand\EXPVAL{om}{
  \IfNoValueTF{#1}
    {\ensuremath{\mathbb{E}\!\left[#2\right]}}
    {\ensuremath{\mathbb{E}_{#1}\!\left[#2\right]}}
}

\NewDocumentCommand\IND{d<>o}{
  \IfNoValueTF{#1}
    {\IfNoValueTF{#2}{}{\ensuremath{_{#2}}}}
    {\IfNoValueTF{#2}
      {\ensuremath{_{\textrm{#1}}}}
      {\ensuremath{_{\textrm{#1}, #2}}}}
}
\NewDocumentCommand\SUP{d<>o}{
  \IfNoValueTF{#1}
    {\IfNoValueTF{#2}{}{\ensuremath{^{\left(#2\right)}}}}
    {\IfNoValueTF{#2}
      {\ensuremath{^{\textrm{#1}}}}
      {\ensuremath{^{\textrm{#1}, \left(#2\right)}}}}
}

\NewDocumentCommand\MIN{o}{
  \IfNoValueTF{#1}{\min}{\min\limits_{#1}\,}
}
\NewDocumentCommand\MAX{o}{
  \IfNoValueTF{#1}{\max}{\max\limits_{#1}\,}
}

\makeatletter
\DeclareRobustCommand\onedot{\futurelet\@let@token\@onedot}
\def\@onedot{\ifx\@let@token.\else.\null\fi\xspace}
\makeatother
\newcommand{\custompar}[1]{\noindent\textbf{#1:\;}}

\newcommand\wrt{w.\kern0.05em r.\kern0.05em t\onedot}

\graphicspath{{./figures}} 

\begin{document}

\maketitle
\phantomsection\def\thefootnote{*}\footnotetext{Equal contribution.}\def\thefootnote{\arabic{footnote}}

\input{sections/01-abstract}
\input{sections/02-introduction}
\input{sections/03-proposed}

\input{sections/04-experiments}

\input{sections/05-results}

\input{sections/06-conclusion}

{
\small
\bibliography{refs}
\bibliographystyle{acl_natbib}

}

\newpage

\appendix
\crefalias{section}{appendix}
\crefalias{subsection}{appendix}
\crefalias{subsubsection}{appendix}
\input{sections/07-appendix}

\end{document}

%% file: sections/01-abstract.tex
\begin{abstract}
We introduce FlexTab, a flexible encoder-decoder architecture for in-context learning on tabular data that pairs a single, task-agnostic encoder with a suite of task-specific decoders. Unlike existing tabular in-context learners, which entangle feature representations with a specific prediction target, our design produces \textit{target-agnostic} row embeddings that can be leveraged across a wide range of downstream tasks within a table-native in-context learning setup. We demonstrate this flexibility on seven distinct problems: classification, regression, anomaly detection, clustering, entity matching, retrieval, and entity classification in relational databases. Both the encoder and the task-specific decoders are trained on tasks generated from a large corpus of real-world, unlabeled tables. FlexTab achieves state-of-the-art performance on classification, regression, anomaly detection, entity matching, and retrieval, while remaining competitive with specialized models on entity classification in a relational setting. These results demonstrate that a single shared encoder, paired with task-specific decoders, can serve as an effective general-purpose backbone for diverse tabular prediction problems. The inference code and checkpoints are publicly available at: \url{https://github.com/SAP-samples/flextab}.
\end{abstract}

%% file: sections/02-introduction.tex
\section{Introduction}
\label{sec:introduction}

Recently, tabular in-context learners such as TabPFN, TabICL, Mitra, and ConTextTab~\cite{tabpfnv1,tabpfnv2,tabicl,qu2026tabiclv2,mitra,contexttab} have been shown to outperform conventional per-dataset trained and tuned models across a range of tabular prediction tasks.
They outperform established baselines such as XGBoost, CatBoost, and LightGBM~\cite{xgboost,catboost,lightgbm}, as well as recent state-of-the-art deep-learning approaches such as RealMLP and TabM~\cite{realmlp,tabm}, and even automated machine learning solutions such as AutoGluon~\cite{autogluon}.
As such, tabular in-context learners have broken the long-standing paradigm of training, tuning, and deploying individual models per dataset or predictive problem.
Furthermore, they open up many new possibilities in both research and industry applications, for example, enabling on-the-fly predictions in agentic workflows, scientific discovery, and other fast-iteration scenarios.

However, their development and application have mostly focused on standard supervised prediction tasks, namely classification and regression.
While some variations exist, in particular in the TabPFN ecosystem (e.g., embeddings, interpretability, anomaly detection, missing value imputation, data generation, and hypothesis testing) as well as problem-specific adaptations (e.g., causal inference, time-series forecasting, or graph predictions), these are mostly created in a post-hoc fashion by recasting the problem at hand as a single-table classification or regression problem, or require significant architectural adaptations and full retraining.
Moreover, many of these community contributions have not yet been thoroughly studied or compared with existing methods in the literature.

While effective in many cases, this general approach exhibits the following limitations:
(a) the necessity to reformulate every (sub)problem as a single-table classification or regression task or perform significant modifications to an existing model architecture;
(b) the tight coupling between internal latent representations and predictive modeling, which causes features to always be conditioned on targets; and
(c) the inability to leverage synergies between different tasks.

We argue that these limitations cannot be jointly overcome within the design principle of current tabular in-context learners, due to the inherent entanglement of internal target-aware representations and predictive decoding.
To this end, we propose a \textit{flexible} tabular in-context learning architecture, FlexTab, based on an encoder-decoder setup that clearly separates internal latent representations from predictive paths by using a variety of problem-specific decoders. 
This strictly decouples target-agnostic feature representation from task-specific predictive modeling.

We train our approach on a large collection of real-world tables and evaluate on a diverse set of tasks. Beyond the well-studied classification and regression settings, these include less common tasks:
\begin{itemize}[leftmargin=1.2em, labelsep=0.5em, itemsep=2pt, topsep=2pt, parsep=0pt]
    \item \textbf{Anomaly detection:} identifying rows in a table that are considered outliers: unsupervised (no labels are provided), one-class (the context rows are all considered inliers), and semi-supervised (only a few rows are labeled as either inlier or anomaly).
    \item \textbf{Clustering:} grouping rows into coherent clusters without access to any labels, so that rows within a cluster are similar to each other.
    \item \textbf{Entity matching:} deciding for pairs of rows from two different tables whether they describe the same real-world entity.
    \item \textbf{Retrieval:} selecting context rows that are most informative for a given query row, which are then used as context for prediction.
    \item \textbf{Entity classification in relational databases:} predicting the target of a row by additionally leveraging related rows in other tables linked via primary-foreign key relations.
\end{itemize}
Together, these tasks highlight the flexibility of our approach in both single- and multi-table scenarios. We achieve competitive or state-of-the-art performance across all investigated scenarios, and we also highlight the need for better benchmarks and baselines in under-studied areas such as tabular anomaly detection, clustering, and matching.

%% file: sections/03-proposed.tex
\begin{figure*}
    \centering
    \includegraphics[width=\linewidth]{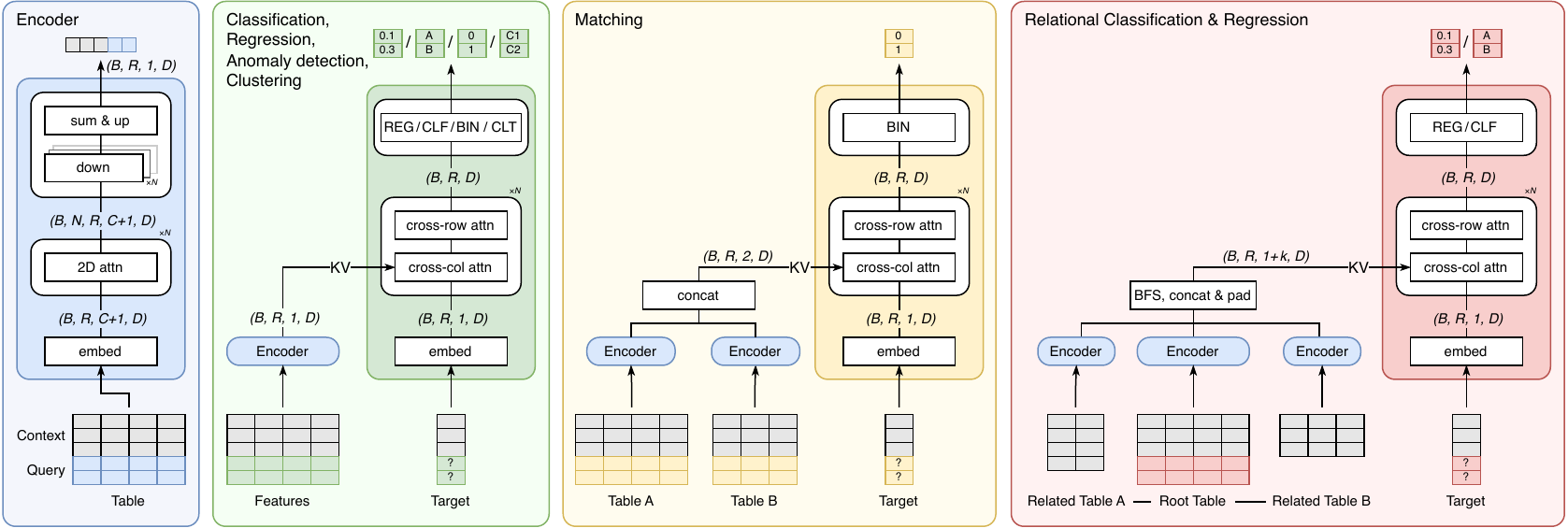}
    \caption{A simplified schematic overview of our proposed architecture, separating an in-context encoder from task-specific decoders with regression (REG), classification (CLF), binary classification (BIN) and clustering (CLT) heads. Note that details such as normalization and feed-forward layers as well as skip connections are omitted.}
    \label{fig:architecture}
\end{figure*}

\section{Tabular In-Context Learning}
\label{sec:architecture}
\subsection{Background}
Tabular in-context learners were primarily pioneered by works on prior-fitted networks and (conditional) neural processes~\cite{pfn,neuralprocesses}, resulting in the first version of TabPFN~\cite{tabpfnv1}.
Building on its core in-context learning principles, a multitude of works extended and refined TabPFNv1, ranging from changes in the synthetic pretraining priors~\cite{tabicl,mitra}, to incorporating retrieval~\cite{tabdpt}, adapting the attention mechanism from row-wise token attention to cell-wise alternating 2D-attention~\cite{tabpfnv2}, improving equivariance~\cite{equitabpfn}, accuracy, and speed~\cite{tabicl,qu2026tabiclv2}, and handling semantic features~\cite{contexttab}, to name a few.
Most of the prior art focuses on tabular classification and regression tasks.
While some works exist that transfer or apply tabular in-context learners to non-tabular domains such as time series forecasting~\cite{timeseriespfn}, reinforcement learning~\cite{tabpfnRL}, or the graph domain~\cite{ofgraphsandtables,yandexturningtfmsintogfms,graphpfn}, applications to further tabular tasks remain scarce.
Several works have investigated causal inference~\cite{dopfn,causalpfn}, as well as unsupervised clustering~\cite{marszalek2025zeus,tabclustpfn}, or anomaly detection~\cite{marszalek2026tactic,wei2026iclad}, including very recent preprints.
However, these approaches require fundamental changes to the model architecture and/or the pretraining data.
We argue that tabular foundation models should natively handle and generalize to a multitude of tabular tasks beyond classification and regression, without the need to recast problems as single-table classification or regression tasks, and without substantial architectural adaptations.
While some additional applications are supported in the community-maintained extensions of TabPFN (e.g.,\ anomaly detection) and some may be approached using embeddings extracted from current models (e.g.,\ matching via row embeddings), these approaches have not been experimentally investigated and come with their respective shortcomings.
For example, embeddings from current models such as TabPFN, TabICL, or ConTextTab are \textit{target-aware}.
Obtaining embeddings from rows or tables without a target (for example, for clustering, matching, or retrieval) requires the use of dummy targets, e.g.,\ by adding random ones, which hurts performance. To overcome these limitations, we propose a flexible architecture, built on the core principles of tabular in-context learning, that natively handles different tabular tasks and can easily be extended to additional ones in the future.

\subsection{Proposed Architecture}
Our proposed architecture is depicted in \cref{fig:architecture} in a simplified way, with more details presented in \cref{app:arch-details}.
At its core, our architecture separates general (target-agnostic) context and query row embeddings from task-specific decoding.
The target decoding is performed by separate task-specific decoders that contextualize the targets with the feature row embeddings via cross-attention.
The encoder and decoders are built around 2D-attention blocks, with alternating cross-row (vertical) and cross-column (horizontal) attention mechanisms -- analogous to the design of TabPFN, Mitra, and ConTextTab -- but separate the target prediction stream into different decoders to make the feature encodings target-agnostic and thus usable across a range of prediction problems.

\custompar{Encoder} 
The encoder builds on the general setup of the ConTextTab architecture~\cite{contexttab}, adopting the same embedding strategy with separate numerical, categorical, and time embedding layers described in \cref{app:arch-details}.
Given input tables with batch size $B$, $R$ rows, and $C$ columns, the data is embedded into a tensor of shape $(B, R, C, D)$, where $D$ denotes the hidden dimension. 
A learnable \ROWT token (replicated across rows) is appended to the embeddings along the column axis, resulting in a tensor of shape $(B, R, C+1, D)$, similar to \CLST tokens in BERT-like models~\cite{bert,RoBERTa} and TabICL~\cite{tabicl,qu2026tabiclv2}, where these tokens are used to generate row-wise (but target-aware) embeddings.
After embedding, cell tokens and \ROWT tokens are jointly contextualized through alternating cross-column and cross-row attention layers.
When the input table consists of both context and query rows, cross-row attention is performed via cross-attention in such a way that context tokens only attend to other context tokens, and query tokens do not attend to each other. 
Similarly, cross-column attention is structured so that \ROWT tokens attend to feature tokens, while feature tokens cannot attend back to the \ROWT tokens. 

Rather than relying solely on the final-layer representations, we collect the \ROWT token outputs from every encoder layer. Each layer's representation is passed through a dedicated linear projection, after which the resulting representations are summed across layers. The aggregated representation is then passed through a shared linear layer, producing the final encoder output. This design is markedly different from conventional encoder-decoder architectures, where typically only the last encoder layer's output is used~\cite{vaswani2017attention}.
We perform thorough ablations on this choice in \cref{sec:ablation-results}.
In our design, the encoder excludes any target values, resulting in general-purpose, target-agnostic row representations that are used as the cross-attention $KV$ in the task-specific decoders.

\custompar{Decoders} 
The different decoders consist of the same 2D-attention blocks as the encoder, with the crucial difference that cross-column attention is realized via cross-attention to the encoder outputs. 
As in common tabular ICL architectures, query targets are decoded from a learnable \TRGT token. 
The key distinction is that cross-row attention is applied exclusively to the target column, while the encoder's \ROWT tokens serve as attention $KV$ inputs to all cross-column attention layers, with the target column acting as the attention $Q$. In this way, target tokens attend to the separately encoded features via cross-attention.
Again, context tokens can only attend to each other and query tokens can only attend to context.
In the main classification and regression decoder, each target attends to a single \ROWT token, in which case the cross-attention reduces to a simple linear projection; however, in specialized decoders (e.g., for matching or relational entity classification tasks), the target may attend to multiple tokens that originate from separate tables with potentially different schemas.
While not the focus of this investigation, our architecture also naturally extends to incorporate other external embeddings, similar to multi-modal LLMs.
For example, the decoder could attend to embeddings produced by text embedding models or other knowledge sources, enabling the decoder to integrate heterogeneous data sources beyond tabular inputs.

\custompar{Classification and Regression} 
The main design follows the description above, with separate prediction heads for regression and classification but otherwise shared decoder weights.
The classification head is identical to that used in ConTextTab.
However, unlike ConTextTab, we treat regression as a classification task and predict logits over quantile bins instead of producing a direct point estimate optimized via an $L^2$ loss.
This simplifies the implementation and training, and enables quantile-based uncertainty estimates. 
In the default setting, we match the number of transformer blocks $N$ to that of the encoder, but also perform ablations on asymmetric setups in \cref{sec:ablation-results}.
Additional details on the target embedding and prediction heads are provided in \cref{app:arch-details}.

\custompar{Anomaly Detection} By definition, anomaly detection is a form of binary classification with imbalanced, noisy, and potentially missing labels. As such, we treat it with the same architecture as classification, but with a decoder trained on binary targets simulating an anomaly detection task. As is customary, we distinguish three different anomaly detection setups~\cite{wei2026iclad,marszalek2026tactic}:

(1)~\emph{Unsupervised}, where context and query are assumed to have the same distribution of normal and anomalous points, but no explicit label is provided for any of the context rows.
(2)~\emph{One-class} (a.k.a.\ novelty detection), where the context is known to contain only normal points; and 
(3)~\emph{Semi-supervised}, where context and query have the same distribution, but only a percentage of the anomalies in context are labeled, while the majority of data is unlabeled. We train a single decoder for all three tasks.

\custompar{Clustering} We convert the above supervised classification framework to an unsupervised clustering one by masking the target column of all rows with \TRGT, including the context. 
The final head layer outputs an embedding vector per row, which is used for clustering.

\custompar{Matching}
We consider cross-table row matching as a pairwise binary classification problem.
Row pairs are treated as matching targets, and the two tables are assumed to have the same number of rows.
In particular, we do not consider the steps of a full matching pipeline such as blocking~\cite{konda2016magellan} for creating pairwise candidate tables, but focus on the prediction step, as is common in the literature.
To apply a tabular in-context learning model to pairwise cross-table matching, one could recast the task as a single-table problem by concatenating the rows into one table.
However, we propose to first encode each table separately, contextualizing each table independently using only its own rows, and then decode the binary matching target using cross-attention to the two separate row embeddings in the decoder stream, as depicted in \cref{fig:architecture}.
The binary classification head is architecturally identical to the default classification head, but with the restriction to two target classes.

\custompar{Entity Classification in Relational Databases}
To further demonstrate the flexibility of our architecture in multi-table scenarios, we examine entity classification tasks in relational databases. 
At inference time, we first select a root table that contains the target column and split it into context and query rows. 
For each row, we perform a breadth-first search (BFS) across the database up to a maximum of $n$ hops to identify related rows. Typically, \num{2} hops are sufficient to reach all tables within the examined databases. The child rows, obtained via primary-to-foreign key relationships, are subsampled and restricted to rows whose timestamps precede the root row.
Then, for each table, rows are contextualized and encoded jointly using our encoder, yielding row embeddings for all rows in all tables. 
Finally, for each row in the task table, we pass the concatenated and padded row embeddings of all rows discovered via BFS to the decoder.
The decoder thus learns to identify patterns over sequences of embeddings corresponding to structurally related rows, enabling relational reasoning across tables.
A key advantage of this approach is that it decouples representation learning at the table level from relational reasoning at the decoder level. By encoding rows within each table jointly, we obtain high-quality, context-aware embeddings that capture intra-table information. 
At the same time, the decoder can flexibly attend over a large and diverse set of \ROWT tokens, allowing it to aggregate information from multiple tables and relation paths.

%% file: sections/04-experiments.tex
\section{Training Details and Task Generation}
\label{sec:experiments}

\custompar{Pretraining}
First, we pretrain the encoder together with the classification and regression decoder on conventional tabular prediction tasks, similar to previous tabular in-context learners.
We then freeze the encoder and train the additional task-specific decoders on top using their respective task generation and loss objectives.
We do this because the generation of classification and regression tasks for large-scale training of tabular in-context learners is the most well-studied and has proven effective, including using real-world data instead of synthetically generated tasks~\cite{tabdpt,contexttab,GeneralizationCanEmerge}.
In principle, it is also possible to pretrain the encoder in a self-supervised fashion separately, or to train the full architecture end-to-end. 
Throughout, for training, we use a large corpus of ca.\ \num{300}\,k unlabeled tables from an internal, proprietary dataset compiled from different public sources, from which we generate the decoder-specific tasks.
We report more details and perform a contamination study of this corpus in \cref{app:pretrain-data}.
By default, we use 12 layers and a hidden dimension of \num{768}, with \num{12} heads of \num{64} dimensions, and a feed-forward dimension of \num{3072}, for a total of ca.\ \num{175}\,M encoder parameters. We train on 8 H100 GPUs, using a micro-batch size of 1 and accumulating gradients over \num{256} micro-batches, with the AdamW optimizer and a learning rate of $10^{-4}$. We run the initial pretraining for approximately \num{30}\,M steps (roughly 7 days), and the training of the additional decoders takes between \num{7}\,M and \num{15}\,M steps each, depending on the task.

\custompar{Classification and Regression}
We generate classification and regression tasks from randomly sampled real-world tables in a similar fashion to prior work~\cite{contexttab}. 
At each step, we randomly select \num{1000} rows, then sample between \num{50} and \num{900} rows as queries, and use the rest as context. 
Subsequently, we randomly sample feature and target columns. 
For both classification and regression, we use a cross-entropy loss.

\custompar{Anomaly Detection}
Anomaly detection tasks are sampled similarly to classification tasks, with a single column selected as target.
If this column is numeric, it is transformed to categorical by selecting ``optimal'' cuts via K-means, with $2 \leq k \leq 10$ chosen via incremental goodness-of-variance-fit thresholding. 
Then, we map each category to binary (0 = normal data, 1 = anomaly) by marking rarer categories as anomalies; the fraction of anomalies is randomly chosen, not exceeding \SI{40}{\percent} of the rows. 
To mimic different types of anomalies in real applications, we use multiple techniques to split the data into context and query: random splitting, selecting only normal points as context, or excluding some anomaly classes from the context. While the model output is binary, context rows have an additional label 2 for unknown labels (in unsupervised or semi-supervised settings).

\custompar{Clustering} 
The sampling strategy follows that of the classification tasks, but target values are completely masked. 
The decoder outputs per-row embedding vectors tuned for unsupervised clustering.
During training, we compute the cosine similarity between such vectors and apply element-wise binary cross-entropy loss to these similarities. For inference, we apply standard K-means on the embedding vectors.
More details on the data sampling and training loss are provided in \cref{sec:clustering_architecture}.

\custompar{Matching}
To generate matching tasks from real-world tables, we use two equally weighted strategies:
a same-row matching construction and a hidden target-based matching construction.
For the former, we create synthetic matching pairs by splitting a randomly drawn table into two with a random overlap. We then shuffle the tables. Pairs that originate from the same original row are considered matches; others are considered non-matches.
For the second approach, we do the same but consider rows as matching if they agree on a held-out target column.
We describe this sampling procedure in more detail in \cref{app:arch-details}.
The training loss is a cross-entropy loss on the predicted binary targets.

\custompar{Entity Classification in Relational Databases}
To simulate the incorporation of information from multiple tables, we generate examples using column-wise partitions of single tables. At each training step, we draw a table and split it into a variable number of chunks, with each chunk containing at least two columns. 
These chunks are batched and processed independently by the encoder, producing separate \ROWT embeddings for each chunk. 
As a result, each original table row is represented by multiple \ROWT tokens, and this setup enables the decoder to learn how to aggregate information from multiple tables to predict the target, which is selected in the same way as for the standard classification and regression decoder. 
To regularize pretraining, we apply perturbations: 
\SI{20}{\percent} of \ROWT embeddings are randomly duplicated within each row, \SI{20}{\percent} are randomly zeroed out, and Gaussian noise is added to \SI{20}{\percent} of the embeddings.

%% file: sections/05-results.tex
\input{tables/classif_reg}

\section{Results}
\label{sec:results}

\custompar{Classification and Regression}
We evaluate on a mix of conventional, numeric-heavy benchmarks as well as semantic ones.
For the conventional benchmarks, we evaluate on the single-fold TabArena-Lite~\cite{tabarena} and TALENT-Tiny~\cite{talent}.
For the semantic ones, we use CARTE~\cite{CARTE} and TextTab~\cite{texttabbench}.
Throughout, we report mean accuracy and soft-clipped $R^2$, as well as mean rank.
We compare against a range of best-in-class baselines, including SOTA in-context learners (TabPFNv2.6, TabICLv2, and ConTextTab), as well as per-dataset trained ones (CatBoost and RealMLP), tuned and ensembled via \num{5}-fold inner cross-validation.
Further, we compare against the AutoGluon system, using its ``extreme'' preset, which itself stacks many of the previous models.
We provide additional evaluations (win-ratios, Elo scores, and critical difference diagrams) in \cref{app:add-results} and more information on the baselines in \cref{app:baselines}. Furthermore, we refer to \cref{app:runtime-analysis} for an analysis of the model runtime in comparison to other baselines.

The main results, evaluating FlexTab (with and without ensembling) and baselines, are shown in \cref{tab:tabular-benchmarks}. 
We exclude AutoGluon from the overall rank comparison due to its stacking nature, making it difficult to highlight architectural differences. Overall, our model performs competitively across non-semantic benchmarks while setting a new standard on the semantically rich CARTE and TextTab benchmarks, where it ranks best. 
This highlights the general efficacy of our approach. 
Crucially, compared to TabPFN and ConTextTab, this shows that feature context compression into single row tokens can achieve state-of-the-art results.
We note that this difference is statistically significant in the case of semantic-heavy benchmarks, as shown in the critical difference diagrams in \cref{app:add-results-clf-reg}.

\custompar{Anomaly Detection}
We evaluate on the popular ADBench benchmark \cite{han2022adbench}. We select up to \num{10000} points per dataset and create context and query splits following the same procedure as in ICLAD~\cite{wei2026iclad}. 
We use AUROC as the main metric. 
In addition to top-performing models from ADBench, we include the foundation model FoMo-0D~\cite{shen2025fomod} (unsupervised and one-class) and the contemporary TACTIC approach~\cite{marszalek2026tactic}. 
In the semi-supervised setting, we found that the strongest baselines are in-context learning classification models (e.g., TabPFNv2.6 and TabICLv2).
Results are reported in \cref{tab:outlier_detection_unsup_oneclass} for the unsupervised and one-class evaluations and in \cref{tab:outlier_detection_semisup} for the semi-supervised case. In this last setting, we also include ``semi-supervised novelty detection''~\cite{JMLR:v11:blanchard10a}, with the same annotated inliers as one-class, but where the remaining data is also available to the model.

\emph{Unsupervised:} Our anomaly-detection-tuned variant, FlexTab-AD, shows better performance than most baselines, only slightly behind the recent SOTA TACTIC model, highlighting the effectiveness of our architecture for this task.
However, based on the critical difference diagram reported in \cref{app:add-results-outlier}, we come to the same conclusion as discussed in the original paper \cite{han2022adbench}: in the unsupervised setting, virtually all current methods are statistically indistinguishable.
This calls for both continued model research and, equally important, the curation of more nuanced and diverse benchmarks.

\emph{One-class:} The performance of FlexTab-AD is competitive with all baselines, surpassing the recent TACTIC. However, the classical KNN approach remains statistically not worse than any of the more modern approaches.

\emph{Semi-supervised novelty detection:} When the unlabeled points are presented to the model, one can apply classical supervised methods effectively, and most resulting algorithms vastly outperform one-class approaches, coherently with the results in \cite{JMLR:v11:blanchard10a}. Our dedicated FlexTab-AD achieves the overall best performance in this setting.

\emph{Semi-supervised:} All in-context learning models significantly outperform the traditional baselines. 
FlexTab-AD is the best-performing method when very few anomalies are labeled, but models trained for supervised classification perform better once enough labeled rows are available. 
As shown in the critical difference diagram in the Appendix, TabICLv2 significantly outperforms other supervised methods such as TabPFN and ConTextTab.
However, FlexTab natively handles all three investigated cases, whereas current in-context learners can natively be applied only to the semi-supervised case.

\input{tables/outlier_detection}

\custompar{Clustering} 
While our approach is generally effective, we defer a thorough evaluation and discussion to \cref{sec:clustering_results} due to training and validation instabilities of our approach as well as when reproducing SOTA baselines such as TabClustPFN~\cite{tabclustpfn}.

\custompar{Matching}
For benchmarking, we use a total of five matching datasets from the literature.
Note that most public matching datasets cannot be considered tabular but are rather free-text entity matching tasks where cross-row statistics or context do not play any role.
Hence, we focus on tabular matching tasks with at least two columns of relevant information that may benefit from contextualization.
In particular, we evaluate on the synthetic Febrl4 dataset~\cite{febrl}, as well as Fodors-Zagats, Bikes, eBooks, and Movies sourced from the DeepMatcher study~\cite{mudgal2018deepmatcher} and the Magellan Data Repository~\cite{magellandata}, which we found to be the most structured.
Due to the relatively small number of labeled examples, we perform \num{5}-fold cross-validation and report average match-class F1 scores.

We compare our approach against tabular in-context learners by recasting cross-table matching as a single-table binary classification problem, as well as native matching baselines that are trained per dataset. We report the Hybrid DeepMatcher variant~\cite{mudgal2018deepmatcher} as well as the BERT-based DITTO~\cite{ditto}, but show additional ones in \cref{app:add-results-matching}.
For completeness, we also compare against simple embedding-similarity-based approaches, i.e.,\ by using the cosine similarity of the two row embeddings binarized with a threshold tuned on each dataset's (and each fold's) train split.

The results are depicted in \cref{tab:matching_f1}.
We observe that all tabular in-context learners, when we recast matching as a single-table problem, perform overall quite poorly.
While TabPFN and TabICL seem to perform reasonably well on the synthetic exact-match Febrl4 task, their performance varies considerably on more semantically heavy tasks, such as Fodors-Zagats or Movies.
ConTextTab and FlexTab, which natively handle semantic columns, show slightly more robust performance, but overall fall behind even simple baselines such as the n-gram-based embedding similarity matching.
With a dedicated matching decoder, trained on matching tasks, we can see that FlexTab-Match performs very strongly.
Our in-context approach even outperforms dedicated per-dataset-trained deep learning baselines such as those from the DeepMatcher study or DITTO.
Overall, our approach performs best in terms of average F1 and ranks almost on par with DITTO, highlighting its effectiveness.

Moreover, we observe that the embeddings obtained from our encoder clearly outperform those from TabPFN when used for matching, highlighting the benefit of the general-purpose, target-agnostic row embeddings produced by our architecture.
Nevertheless, for many matching tasks, simple n-gram hash embeddings perform very well, while their performance strongly degrades on others (e.g.,\ Bikes and Movies).
Overall, we can already achieve reasonable results with simple cosine-similarity-based matching using FlexTab's row embeddings, which can be useful in large-scale matching scenarios.
However, to achieve the best performance, the native matching decoder is preferred.

We show extended results, including all DeepMatcher baselines, as well as additional embeddings and combining TabPFN and TabICL with TF-IDF features, in \cref{app:add-results-matching}.
There, we also ablate the pretraining strategy used, creating synthetic matching tasks from real-world tables.

\input{tables/matching}

\custompar{Entity Classification in Relational Databases}
As the de facto standard in relational prediction benchmarking, we evaluate our model on the 12 RelBench~\cite{relbench} entity classification tasks, following a setup similar to that of established multi-table foundation models such as Griffin~\cite{griffin} and Relational Transformer~\cite{relational_transformer}, and report mean AUROC in \cref{tab:relational}. 
We additionally include results for KumoRFM~\cite{kumorfm}, a SOTA closed-source industry model, as well as the recent RDBLearn approach~\cite{rdblearn}.
Aside from ours, we report results taken from the respective publications and KumoRFM-2~\cite{kumorfm2}.

Our model's performance is on par with that of Relational Transformer, showcasing the competitiveness and flexibility of our approach relative to recent architectures that are purpose-built for relational prediction.
Note that, unlike the other investigated relational-native approaches, ours is pretrained exclusively on chunks created from single tables, not on actual relational data.
By investigating FlexTab's performance with a varying number of auxiliary rows in \cref{app:add-results-relational}, we verify that a growing number of row tokens from related tables and larger context size monotonically improve performance, showcasing that the decoder can successfully leverage information contained in tokens from additional auxiliary tables.

The recent RDBLearn approach~\cite{rdblearn}, which uses a custom feature engineering technique, aggregating features across the relational tables and performing its prediction using a single-table in-context learner, demonstrates that relational-native models still do not show a clear gain over this simpler solution.
However, this approach is primarily applicable to databases where such aggregation preserves the relevant signal, and may not generalize to settings where textual values across different tables must be jointly considered. 
Overall, the field of database-native predictive models as well as benchmarking remains an open and active research area, requiring continued work on table encoding~\cite{meyer2025relate} and relational data modeling~\cite{dwivedi2025relationalgraphtransformer}.

\input{tables/relational}

\custompar{Retrieval} To more directly demonstrate the meaningfulness of FlexTab's embeddings, we conduct retrieval experiments and report evaluation results in~\cref{tab:retrieval-main-res} (and in more detail in \cref{tab:retrieval-extended-res} in \cref{app:add-results}) across varying context sizes and retrieval methods. In the default setting, we randomly sample training rows as context and show that performance improves as the context size grows. An alternative is a specialized retrieval method in which we first embed all rows with the FlexTab encoder by selecting $k$ random training rows as context and obtaining the row embeddings for all training and test rows. For each test row, we then retrieve the top $k$ context rows based on the L2 distance between embeddings (similar to existing approaches~\cite{tabdpt, localpfn}). Having retrieved the best training rows for each query, we use them as context to make the prediction. We find that retrieving just 1024 dedicated rows per query, via FlexTab embedding similarity, already yields better scores than selecting 8192 random rows.

In addition, in~\cref{tab:retrieval-main-res}, we show that retrieval with FlexTab embeddings outperforms the specialized retrieval based on other embedding methods. We compare against two schemes: a hashing vectorizer applied to the serialized row, and a hashing vectorizer applied to each text column combined with a standard scaler for numerical columns, where the vectorized columns are concatenated into the final embedding. This confirms that the general-purpose, target-unaware row embeddings of FlexTab effectively represent the underlying row’s content. Furthermore, we directly probe the embeddings via KNN and evaluate the final predictions. We show that KNN-5 over embedded rows using the FlexTab encoder performs significantly better than KNN-5 over embedded rows using the hashing and standardization methods described above, highlighting the quality of FlexTab's general-purpose embeddings.

\input{tables/retrieval}

\subsection{Architecture Ablations}
\label{sec:ablation-results}

We conduct two ablation studies to assess our design choices, focusing on (a) the strategy used to aggregate encoder representations and (b) the depth of the decoder. Due to the strong availability of high-quality benchmarks for classification and regression, we perform the studies on those. Results are reported in \cref{tab:tabular-benchmarks-ablation}, using the same benchmarks as previously. For simplicity, we evaluate FlexTab in the non-ensembled variant here.

\custompar{Encoder Aggregation}
We compare five strategies: combining projected encoder representations at dimensions \num{768}, \num{512}, and \num{256}; a simple mean aggregation over encoder layers; as well as using only the last encoder layer output directly.
In all cases, aggregations are performed over the contextualized \ROWT embeddings as previously described.
Surprisingly, when we used only the output of the last layer, as is common in the original Transformer encoder-decoder setup~\cite{vaswani2017attention}, we observed training collapse after roughly \num{2}\,M steps.
We then investigated the different aggregation strategies.
The learned aggregations, using individual projection layers, clearly outperform the non-parametric alternatives. The \num{768}-dimensional variant achieves the best overall rank. 
Mean pooling across encoder layers noticeably lags behind, indicating that uniform weighting discards useful layer-specific information. Overall, this suggests that intermediate layers carry beneficial complementary signals. 

\custompar{Decoder Size}
Next, we vary the decoder depth among 4, 8, and 12 layers while keeping the encoder fixed, investigating asymmetric setups. 
The 12- and 8-layer variants are essentially tied in overall rank. Shrinking the decoder to 4 layers causes a clear drop in performance. 
This gap highlights the need for adequate decoder capacity to handle both classification and regression effectively, but also demonstrates the flexibility of the approach to scale the decoder capacity independently of the encoder.
We adopt the 12-layer decoder as our default, given its consistency across benchmarks, but note that the 8-layer variant offers a competitive and more efficient alternative.

\input{tables/classif_reg_ablation}

%% file: tables/classif_reg.tex
\begin{table*}
\centering
\footnotesize
\caption{Classification and regression performance, depicting mean accuracy (Acc) for classification and (soft-clipped) $R^2$ score for regression tasks, in percent, as well as mean per-task rank.}
\label{tab:tabular-benchmarks}
\setlength{\tabcolsep}{3.5pt}
\begin{tabular}{l
S[table-format=2.1, detect-weight=true]
S[table-format=2.1, detect-weight=true]
S[table-format=2.1, detect-weight=true]
S[table-format=2.1, detect-weight=true]
S[table-format=2.1, detect-weight=true]
S[table-format=2.1, detect-weight=true]
S[table-format=2.1, detect-weight=true]
S[table-format=2.1, detect-weight=true]
S[table-format=2.1, detect-weight=true]
S[table-format=2.1, detect-weight=true]
S[table-format=2.1, detect-weight=true]
S[table-format=2.1, detect-weight=true]
S[table-format=2.1, detect-weight=true]
}
\toprule
Model & \multicolumn{1}{c}{All} & \multicolumn{3}{c}{CARTE} & \multicolumn{3}{c}{TabArena-Lite} & \multicolumn{3}{c}{TALENT-Tiny} & \multicolumn{3}{c}{TextTab} \\
  \cmidrule(lr){2-2} \cmidrule(lr){3-5} \cmidrule(lr){6-8} \cmidrule(lr){9-11} \cmidrule(lr){12-14}
 & {Rank} & {Rank} & {Acc} & {R2} & {Rank} & {Acc} & {R2} & {Rank} & {Acc} & {R2} & {Rank} & {Acc} & {R2} \\
\midrule

AutoGluon & \NAN & \NAN & 78.8 & 74.0 & \NAN & 88.6 & 80.7 & \NAN & 88.9 & 84.7 & \NAN & 83.5 & 67.0 \\
\arrayrulecolor{mygrey}\midrule\arrayrulecolor{black}
\bfseries{FlexTab [bag=8] (ours)} & \bfseries 2.7 & \bfseries 1.6 & \bfseries 77.3 & \bfseries 74.4 & 3.6 & 87.1 & 79.5 & 3.2 & 88.0 & \bfseries 85.8 & \bfseries 2.8 & 83.8 & 62.7 \\
\bfseries{FlexTab [bag=1] (ours)} & 3.2 & 2.3 & 77.0 & 73.6 & 3.8 & 87.0 & 79.1 & 3.4 & 87.9 & 85.5 & 3.2 & 83.7 & 61.4 \\
RealMLP & 3.7 & 4.1 & 74.3 & 69.0 & 3.4 & 88.3 & 79.9 & 3.4 & \bfseries 88.6 & 85.0 & 3.8 & 82.3 & \bfseries 67.7 \\
ConTextTab & 3.7 & 3.0 & 77.1 & 72.4 & 4.2 & 87.6 & 77.8 & 4.0 & 87.6 & 83.2 & 3.8 & \bfseries 84.4 & 58.8 \\
TabPFNv2.6 & 3.8 & 6.3 & 70.7 & 59.9 & \bfseries 1.9 & \bfseries 88.7 & \bfseries 80.5 & 2.5 & 88.4 & 84.9 & 4.5 & 81.5 & 63.8 \\
TabICLv2 & 3.9 & 6.5 & 70.5 & 55.3 & 2.2 & \bfseries 88.7 & 79.9 & \bfseries 2.2 & 88.3 & 85.3 & 4.1 & 82.5 & 60.2 \\
CatBoost & 4.0 & 4.3 & 76.3 & 68.3 & 3.8 & 88.2 & 79.0 & 4.2 & 87.1 & 83.5 & 3.3 & 83.7 & 65.4 \\
Random forest & 6.0 & 6.8 & 71.4 & 63.4 & 5.5 & 87.6 & 76.1 & 5.9 & 85.6 & 78.0 & 5.7 & 79.8 & 60.7 \\
\bottomrule
\end{tabular}
\end{table*}

%% file: tables/outlier_detection.tex
\begin{table}
\begin{minipage}[t]{0.32\textwidth}
  \centering
  \footnotesize
   \setlength{\tabcolsep}{2pt}
    \caption{Unsupervised and one-class anomaly detection: mean rank (Rk) and AUROC (in \%).}
    \label{tab:outlier_detection_unsup_oneclass}
    \begin{tabular}{l
    S[table-format=2.1, detect-weight=true]
    S[table-format=2.1, detect-weight=true]
    S[table-format=2.1, detect-weight=true]
    S[table-format=2.1, detect-weight=true]}
    \toprule
    Model & \multicolumn{2}{c}{Unsup.} & \multicolumn{2}{c}{One-Class} \\
      \cmidrule(lr){2-3} \cmidrule(lr){4-5}
     & {Rk} & {Avg} & {Rk} & {Avg} \\
    \midrule
KNN \cite{ramaswamy2000efficient} & 4.6 & 69.7 & \bfseries 3.3 & 81.4 \\
\bfseries{FlexTab-AD} & 4.5 & 73.8 & 3.7 & \bfseries 82.0 \\
TACTIC \cite{marszalek2026tactic} & \bfseries 3.9 & \bfseries 74.9 & 4.7 & 80.3 \\
LOF \cite{LOF} & 5.4 & 64.3 & 4.3 & 79.7 \\
OCSVM \cite{scholkopf1999support} & 4.9 & 71.4 & 5.2 & 76.3 \\
IForest \cite{liu2008isolation} & 4.5 & 72.9 & 5.8 & 75.9 \\
PCA \cite{shyu2003novel} & 5.1 & 69.5 & 5.3 & 76.6 \\
ECOD \cite{ECOD} & 5.1 & 71.2 & 7.2 & 70.6 \\
FoMo-0D \cite{shen2025fomod} & 7.0 & 58.0 & 5.4 & 78.3 \\
    \bottomrule
    \end{tabular}
\end{minipage}%
\hfill
\begin{minipage}[t]{0.66\textwidth}
  \centering
  \footnotesize  
   \setlength{\tabcolsep}{2pt}  
  \caption{Semi-supervised novelty and semi-supervised anomaly detection: mean rank (Rk) and AUROC (in \%) for different levels of labeled anomaly ratios.}
    \label{tab:outlier_detection_semisup}
    \setlength{\tabcolsep}{2pt}
\begin{tabular}{l
S[table-format=2.1, detect-weight=true]
S[table-format=2.1, detect-weight=true]
S[table-format=2.1, detect-weight=true]
S[table-format=2.1, detect-weight=true]
S[table-format=2.1, detect-weight=true]
S[table-format=2.1, detect-weight=true]
S[table-format=2.1, detect-weight=true]
S[table-format=2.1, detect-weight=true]
S[table-format=2.1, detect-weight=true]
S[table-format=2.1, detect-weight=true]}
\toprule
Model & \multicolumn{2}{c}{Novelty} & \multicolumn{2}{c}{1\%} & \multicolumn{2}{c}{5\%} & \multicolumn{2}{c}{25\%} & \multicolumn{2}{c}{100\%} \\
  \cmidrule(lr){2-3} \cmidrule(lr){4-5} \cmidrule(lr){6-7} \cmidrule(lr){8-9} \cmidrule(lr){10-11}
 & {Rank} & {Avg} & {Rank} & {Avg} & {Rank} & {Avg} & {Rank} & {Avg} & {Rank} & {Avg} \\
\midrule
TabICL \cite{qu2026tabiclv2} & 3.1 & 87.9 & 3.4 & 82.6 & 3.0 & \bfseries 88.0 & \bfseries 2.5 & \bfseries 91.9 & \bfseries 2.2 & \bfseries 94.4 \\
\bfseries{FlexTab-AD} & 2.7 & \bfseries 90.7 & \bfseries 2.7 & \bfseries 83.6 & \bfseries 2.9 & 87.3 & 3.5 & 90.7 & 4.9 & 93.2 \\
TabPFN \cite{tabpfnv2} & \bfseries 2.6 & 90.2 & 4.9 & 78.4 & 4.2 & 85.9 & 3.4 & 90.8 & 2.8 & 94.0 \\
\bfseries{FlexTab} & 5.6 & 80.8 & 3.1 & 82.7 & 3.3 & 87.4 & 3.6 & 91.1 & 3.9 & 93.6 \\
ConTextTab \cite{contexttab} \hspace{-2mm}& 5.2 & 85.1 & 3.7 & 81.7 & 3.8 & 86.9 & 4.1 & 90.9 & 3.7 & 93.6 \\
XGBOD \cite{zhao2018xgbod} & 5.8 & 84.3 & 6.1 & 75.7 & 5.9 & 83.2 & 5.9 & 89.2 & 5.4 & 93.2 \\
DevNet \cite{devnet} & 5.6 & 85.7 & 6.9 & 69.3 & 6.9 & 77.0 & 6.7 & 85.4 & 6.7 & 91.0 \\
DeepSAD \cite{DeepSAD} & 6.4 & 83.3 & 6.9 & 72.5 & 7.5 & 77.7 & 7.6 & 83.7 & 7.4 & 89.2 \\
FEAWAD \cite{FEAWAD} & 8.0 & 77.1 & 7.4 & 67.9 & 7.7 & 75.2 & 7.7 & 83.9 & 8.0 & 89.3 \\
\bottomrule
\end{tabular}
\end{minipage}
\end{table}

%% file: tables/matching.tex
\begin{table*}
\footnotesize
\centering
\caption{Matching benchmark F1 scores (in \%, positive class) obtained as 5-fold cross-validation averages for each dataset. The average per-group rank is denoted as GRk whereas the average rank across all investigated models is denoted as Rk.}
\label{tab:matching_f1}
\begin{tabular}{l l S[table-format=3.1, detect-weight]  S[table-format=3.1, detect-weight]  S[table-format=3.1, detect-weight]  S[table-format=3.1, detect-weight]  S[table-format=3.1, detect-weight]  S[table-format=3.1, detect-weight]  S[table-format=2.1, detect-weight]  S[table-format=2.1, detect-weight]}
\toprule
 & & \multicolumn{5}{c}{Benchmark} & \multicolumn{3}{c}{Average} \\
\cmidrule(lr){3-7} \cmidrule(lr){8-10}
 & Model & {Febrl4} & {F.-Zagat} & {Bikes} & {eBooks} & {Movies} & {F1} & {GRk} & {Rk} \\
\midrule
\multirow{5}{*}{\textcolor{gray}{\textit{\shortstack[l]{In-context\\learners}}}}
 & TabPFN & 97.3 & 29.4 & 81.8 & 67.9 & 0.0 & 55.3 & 3.2 & 7.2 \\
 & TabICL & 95.0 & 36.9 & 82.4 & 73.1 & 0.0 & 57.5 & 2.8 & 6.8 \\
 & ConTextTab & 83.2 & 23.2 & 82.3 & 60.3 & 79.0 & 65.6 & 3.6 & 6.8 \\
 & \textbf{FlexTab (ours)} & 79.8 & 12.7 & 84.3 & 72.6 & 61.0 & 62.1 & 3.6 & 7.0 \\
 & \textbf{FlexTab-Match (ours)} & \bfseries 99.3 & \bfseries 92.5 & \bfseries 86.8 & \bfseries 87.3 & \bfseries 90.2 & \bfseries 91.2 & \bfseries 1.0 & \bfseries 2.2 \\
\arrayrulecolor{gray}\midrule\arrayrulecolor{black}
\multirow{3}{*}{\textcolor{gray}{\textit{\shortstack[l]{Embedding\\similarity}}}}
 & NGramHash Embeddings & \bfseries 99.7 & \bfseries 88.0 & \bfseries 69.8 & \bfseries 88.1 & 22.9 & 73.7 & \bfseries 1.4 & \bfseries 4.4 \\
 & TabPFN Embeddings & 71.8 & 41.2 & 50.8 & 56.4 & 24.0 & 48.9 & 2.8 & 8.6 \\
 & \textbf{FlexTab Embeddings (ours)} & 98.5 & 81.5 & 54.4 & 84.9 & \bfseries 78.9 & \bfseries 79.6 & 1.8 & 5.4 \\
\arrayrulecolor{gray}\midrule\arrayrulecolor{black}
\multirow{2}{*}{\textcolor{gray}{\textit{\shortstack[l]{Per-dataset\\trained}}}}
 & Deepmatch Hybrid & 99.8 & 87.1 & 87.7 & 88.5 & \bfseries 87.6 & \bfseries 90.2 & 1.6 & 2.2 \\
 & DITTO & \bfseries 100.0 & \bfseries 94.7 & \bfseries 95.5 & \bfseries 89.5 & 44.9 & 84.9 & \bfseries 1.2 & \bfseries 2.0 \\
\bottomrule
\end{tabular}
\end{table*}

%% file: tables/relational.tex
\begin{table*}
\footnotesize
\centering
\caption{Entity classification in relational databases prediction performance: Zero-shot test AUROC (\%) for $12$ RelBench binary classification tasks. The target task is never seen during pretraining.}
\label{tab:relational}
\begin{tabular}{l rr rr rr r rr rr r r}
\toprule
& \multicolumn{2}{c}{f1}
& \multicolumn{2}{c}{avito}
& \multicolumn{2}{c}{event}
& \multicolumn{1}{c}{trial}
& \multicolumn{2}{c}{amazon}
& \multicolumn{2}{c}{stack}
& \multicolumn{1}{c}{hm}
& \\
\cmidrule(lr){2-3}\cmidrule(lr){4-5}\cmidrule(lr){6-7}\cmidrule(lr){8-8}\cmidrule(lr){9-10}\cmidrule(lr){11-12}\cmidrule(lr){13-13}
Model
  & dnf & top3
  & clicks & visits
  & repeat & ignore
  & out
  & user & item
  & eng & badge
  & churn
  & Mean \\
\midrule
Griffin                         & 57.7 & 82.5 & 45.9 & 60.7 & 71.9 & 83.3 & 51.0 & 62.3 & 69.0 & 77.5 & 73.5 & 60.2 & 66.29 \\
RT                              & 81.2 & 89.3 & 59.5 & 61.8 & 73.2 & 77.5 & 51.8 & 64.0 & 70.9 & 75.7 & 80.1 & 62.8 & 70.65 \\
\textbf{FlexTab-Multi (ours)}   & 74.6 & 83.3 & 51.9 & 58.5 & \bfseries 77.4 & 79.2 & 65.1 & 62.1 & 76.5 & 81.6 & 80.6 & 60.3 & 70.93 \\
RDBLearn                        & 70.9 & 79.7 & \bfseries 69.0 & \bfseries 65.5 & 75.0 & 82.5 & \bfseries 71.6 & \bfseries 67.6 & \bfseries 82.1 & \bfseries 89.5 & \bfseries 85.4 & \bfseries 68.1 & 75.58 \\
KumoRFM-1                       & \bfseries 82.4 & \bfseries 91.1 & 64.9 & 64.1 & 76.1 & \bfseries 89.2 & 70.8 & 67.3 & 79.9 & 87.1 & 80.0 & 67.7 & \bfseries 76.72 \\
\bottomrule
\end{tabular}%
\end{table*}

%% file: tables/retrieval.tex
\begin{table*}
\centering
\footnotesize
\caption{Influence of embedding methods on retrieval (top) and results of probing the embeddings via KNN (bottom), measured in terms of classification and regression scores.}
\label{tab:retrieval-main-res}
\setlength{\tabcolsep}{3.0pt}
\begin{tabular}{lS[table-format=2.1]S[table-format=2.1]S[table-format=2.1]S[table-format=2.1]S[table-format=2.1]S[table-format=2.1]S[table-format=2.1]S[table-format=2.1]S[table-format=2.1]S[table-format=2.1]S[table-format=2.1]S[table-format=2.1]S[table-format=2.1]}
\toprule
Model Name & \multicolumn{1}{c}{All} & \multicolumn{3}{c}{CARTE} & \multicolumn{3}{c}{TabArena} & \multicolumn{3}{c}{TALENT} & \multicolumn{3}{c}{TextTab} \\
  \cmidrule(lr){2-2} \cmidrule(lr){3-5} \cmidrule(lr){6-8} \cmidrule(lr){9-11} \cmidrule(lr){12-14}
 & {Rank} & {Rank} & {Acc} & {R2} & {Rank} & {Acc} & {R2} & {Rank} & {Acc} & {R2} & {Rank} & {Acc} & {R2} \\
\midrule

\multicolumn{14}{l}{\textit{Influence of embedding methods on retrieval}} \\
\addlinespace[2pt]
FlexTab [k=1024, flextab-emb] & \bfseries 1.7 & \bfseries 1.9 & 77.3 & \bfseries 75.5 & \bfseries 1.4 & \bfseries 87.6 & \bfseries 80.4 & 2.1 & 87.7 & \bfseries 86.3 & \bfseries 1.5 & 84.2 & 70.0 \\
FlexTab [k=1024, column-hash] & 1.9 & 2.3 & 77.2 & 75.1 & 1.7 & 86.7 & 80.2 & \bfseries 1.8 & \bfseries 88.0 & 86.2 & 1.8 & 84.1 & \bfseries 70.3 \\
FlexTab [k=8192, random] & 2.2 & 2.8 & 77.0 & 73.6 & 1.8 & 87.0 & 79.1 & 1.8 & 87.9 & 85.5 & 2.6 & 83.7 & 61.4 \\
FlexTab [k=1024, row-hash] & 2.5 & 2.4 & \bfseries 77.7 & 75.0 & 2.4 & 86.3 & 78.1 & 3.0 & 86.4 & 81.6 & 2.0 & \bfseries 84.3 & 68.6 \\

\midrule
\multicolumn{14}{l}{\textit{Probing the embeddings via KNN}} \\
\addlinespace[2pt]

FlexTab [k=8192, random] & \bfseries 1.1 & \bfseries 1.0 & \bfseries 77.0 & \bfseries 73.6 & \bfseries 1.2 & \bfseries 87.0 & \bfseries 79.1 & \bfseries 1.0 & \bfseries 87.9 & \bfseries 85.5 & \bfseries 1.2 & \bfseries 83.7 & \bfseries 61.4 \\
KNN-5 [flextab-emb] & 2.2 & 2.1 & 72.2 & 63.2 & 2.4 & 84.9 & 67.2 & 2.2 & 82.8 & 72.1 & 2.2 & 78.6 & 51.8 \\
KNN-5 [column-hashing] & 2.9 & 3.4 & 67.1 & 51.2 & 2.6 & 82.2 & 66.7 & 2.7 & 78.5 & 68.7 & 3.0 & 76.4 & 45.0 \\
KNN-5 [row-hashing] & 3.5 & 3.4 & 66.7 & 48.9 & 3.4 & 77.4 & 18.7 & 3.9 & 56.0 & 1.2 & 3.5 & 76.7 & 28.3 \\

\bottomrule
\end{tabular}
\end{table*}

%% file: tables/classif_reg_ablation.tex
\begin{table}
\centering
\footnotesize
\caption{Performance across two ablation groups compared to the base model: (Top) Encoder row-level embedding aggregations; (Bottom) Encoder sizes}
\label{tab:tabular-benchmarks-ablation}
\resizebox{\textwidth}{!}{%
\begin{tabular}{l l
S[table-format=2.1, detect-weight=true]
S[table-format=2.1, detect-weight=true]
S[table-format=2.1, detect-weight=true]
S[table-format=2.1, detect-weight=true]
S[table-format=2.1, detect-weight=true]
S[table-format=2.1, detect-weight=true]
S[table-format=2.1, detect-weight=true]
S[table-format=2.1, detect-weight=true]
S[table-format=2.1, detect-weight=true]
S[table-format=2.1, detect-weight=true]
S[table-format=2.1, detect-weight=true]
S[table-format=2.1, detect-weight=true]
S[table-format=2.1, detect-weight=true]
}
\toprule
 & Model & \multicolumn{1}{c}{All} & \multicolumn{3}{c}{CARTE} & \multicolumn{3}{c}{TabArena-Lite} & \multicolumn{3}{c}{TALENT-Tiny} & \multicolumn{3}{c}{TextTab} \\
  \cmidrule(lr){3-3} \cmidrule(lr){4-6} \cmidrule(lr){7-9} \cmidrule(lr){10-12} \cmidrule(lr){13-15}
 & & {Rank} & {Rank} & {Acc} & {R2} & {Rank} & {Acc} & {R2} & {Rank} & {Acc} & {R2} & {Rank} & {Acc} & {R2} \\
\midrule

\multirow{5}{*}{\textcolor{gray}{\textit{\shortstack[l]{Encoder\\Aggregation}}}}
 & FlexTab (dim 768) & \bfseries 1.7 & \bfseries 1.5 & \bfseries 77.0 & \bfseries 73.6 & \bfseries 1.8 & 87.0 & \bfseries 79.1 & \bfseries 1.7 & \bfseries 87.9 & \bfseries 85.5 & 1.8 & 83.7 & 61.4 \\
 & FlexTab (dim 512) & 1.8 & 2.0 & 76.1 & 73.3 & 1.8 & 86.9 & 77.1 & 1.7 & 87.7 & 85.5 &  \bfseries 1.5 & \bfseries 84.0 & \bfseries 61.9 \\
 & FlexTab (dim 256) & 2.3 & 2.8 & 76.2 & 72.8 & 2.0 & \bfseries 87.5 & 78.7 & 2.2 & 87.2 & 85.2 & 1.9 & 83.5 & 61.5 \\
 & FlexTab (mean) & 2.9 & 3.0 & 75.1 & 72.4 & 2.5 & 86.2 & 78.6 & 3.1 & 86.1 & 84.2 & 3.3 & 82.6 & 59.0 \\
 & FlexTab (last enc layer) & \multicolumn{13}{l}{\hspace{0.55em}Training collapsed} \\

\midrule

\multirow{3}{*}{\textcolor{gray}{\textit{\shortstack[l]{Decoder\\Size}}}}
 & FlexTab (12 layers) & \bfseries 1.5 & 1.8 & 76.1 & 73.3 & \bfseries 1.4 & 86.9 & \bfseries 77.1 & \bfseries 1.3 & \bfseries 87.7 & \bfseries 85.5 & \bfseries 1.2 & \bfseries 84.0 & \bfseries 61.9 \\
 & FlexTab (8 layers) & 1.5 & \bfseries 1.5 & \bfseries 76.7 & \bfseries 73.6 & 1.5 & \bfseries 87.3 & 76.9 & 1.7 & 87.1 & 85.0 & 1.4 & 83.9 & 61.4 \\
 & FlexTab (4 layers) & 2.4 & 2.4 & 76.5 & 72.4 & 2.3 & 87.0 & 74.7 & 2.5 & 85.7 & 82.8 & 2.5 & 83.4 & 58.9 \\

\bottomrule
\end{tabular}
}
\end{table}

%% file: sections/06-conclusion.tex
\section{Conclusion and Outlook}
\label{sec:conclusion}

We presented FlexTab, a flexible encoder-decoder architecture for in-context learning on tabular data. 
By decoupling a shared, target-agnostic encoder from a family of task-specific decoders, FlexTab natively supports a broad spectrum of tabular problems -- including classification, regression, anomaly detection, clustering, entity matching, and entity classification over relational databases -- within a single unified framework. 
The entire model is trained on tasks generated from a large corpus of real-world, unlabeled tables, avoiding reliance on synthetic priors or task-specific pretraining data.
Across our experiments, FlexTab achieves competitive or state-of-the-art performance on all evaluated tasks, demonstrating that general-purpose row embeddings combined with task-specific predictive modeling can serve as an effective foundation for diverse tabular prediction problems.

\custompar{Limitations}
Despite these encouraging results, our approach has several limitations. 
First, compressing each row into a single embedding may introduce an information bottleneck, particularly for wide tables or tasks requiring fine-grained cell-level reasoning. We explore this in more detail in \cref{app:add-results}. 
Second, like other models trained on large-scale real-world data, our model evaluation may suffer from contamination.
We performed a contamination study and found light contamination of evaluation datasets used during pretraining.
We refer to the Appendix, where we discuss both issues in detail.

\custompar{Outlook}
Several promising research directions remain open. 
A natural next step is to replace the currently supervised pretraining of the encoder with a fully self-supervised objective, for example following the LeJEPA or data2vec approaches~\cite{balestriero2025lejepa,baevski2022data2vec}.
Moreover, jointly training all decoders on top of a shared, possibly higher-capacity encoder may unlock synergies across tasks that our staged training does not currently exploit. 
Given the structural similarities among the different decoders, a further refinement would be to consolidate them, for instance through a mixture-of-experts design, reducing parameter redundancy while preserving task specialization. 
Our architecture is not limited to the tasks studied here; it extends to settings such as domain transfer, few-shot adaptation, zero-shot prediction, recommendation, and more. 
As shown by our matching and relational entity classification results, the specific decoders in those cases can pick up signals from multiple embeddings, opening up a range of applications that reason across multiple modalities -- similar to multi-modal LLMs but with an in-context learning focus and table-native design.
We view the breadth of tasks unified under a single encoder as a first step toward multi-task training at scale, which we believe is a promising and still largely unexplored route to scaling laws in the tabular domain.

%% file: sections/07-appendix.tex
\begin{figure*}
    \centering
    \includegraphics[width=\linewidth]{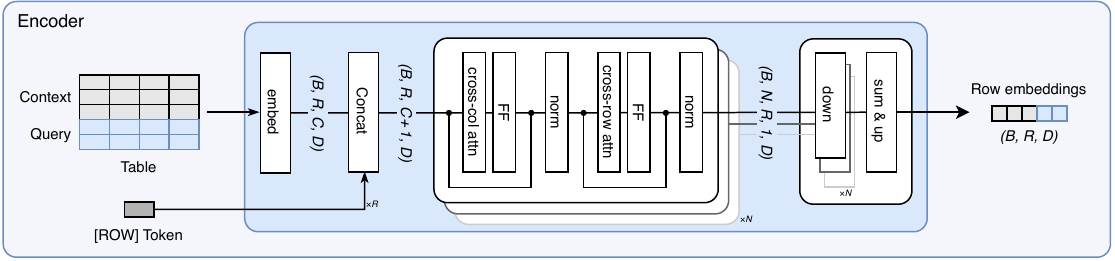}
    \caption{Detailed schematic depiction of the encoder architecture. Note that cross-row attention is performed in a way such that query rows cannot attend to each other, analogous to existing architectures. The learnable \texttt{[ROW]} tokens are repeated along the row axis and then appended to the table embeddings along the column axis. Although depicted in parallel, the transformer blocks are executed sequentially using the output of the previous layer serving as the input to the next, as is common. However, the $N$ individual \texttt{[ROW]} token outputs of each transformer block are separately down-projected, then summed and up-projected to obtain the final row embeddings.}
    \label{fig:encoder-detail}
\end{figure*}

\section{Architectures and Pretraining Details}
\label{app:arch-details}

\subsection{Encoder}

We follow the embedding strategy of ConTextTab~\cite{contexttab}, encoding each cell according to its modality: text, date, or numeric. Text cells are embedded with a pretrained \texttt{all-MiniLM-L6-v2} model, and a learnable linear layer then projects the resulting vector to the FlexTab transformer's input dimension of $d = 768$. Dates are decomposed into day, month, and year; each component is embedded separately, and the three vectors are summed. Numeric columns are discretized into 50 quantile bins, following the procedure described in the decoder's regression section below, and are encoded by piecewise linear interpolation over a set of learned reference vectors. In addition, column headers serve as a form of positional encoding: each header is embedded with a text embedding model, passed through its own learnable linear layer, and added to every cell in its column. Finally, the summed embeddings are normalized with layer normalization.

A more detailed depiction of our proposed encoder backbone is shown in \cref{fig:encoder-detail}.
While depicted in parallel, the 2D Transformer blocks are executed sequentially.
The visualization further illustrates that each block's output is separately (down-)projected, then summed and (up-)projected to obtain the final row embeddings of the input rows.

In all cases, cross-row attention is performed in such a way that context rows can only attend to each other and query rows can only attend to context rows, preventing any query row leakage.
This is achieved efficiently using a cross-attention formulation.
Given an input sequence of shape $(B, S, D)$ with a batch size $B$, sequence length $S$, and hidden dimension $D$, where the first $C$ elements in the sequence are context examples and the last $Q$ elements are query examples, we achieve the desired attention pattern by using the context-truncated sequence of shape $(B, C, D)$ as attention $KV$ and the full sequence of shape $(B, S, D) = (B, C+Q, D)$ as attention $Q$.
If no context/query split is present, all sequence elements are effectively treated as context, and the formulation falls back to full self-attention between all rows.

\subsection{Decoder}

\begin{figure*}
    \centering
    \includegraphics[width=0.8\linewidth]{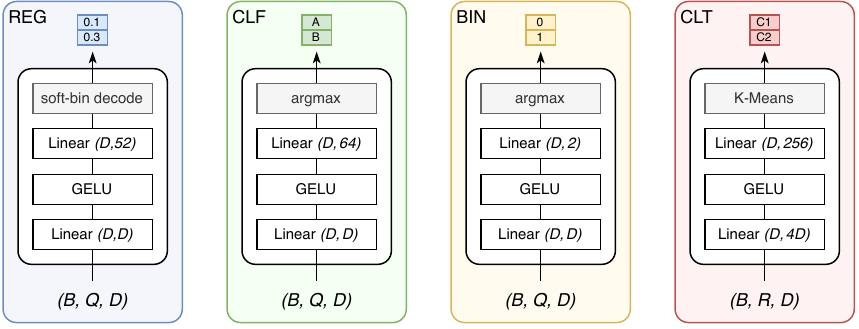}
    \caption{Details of the used heads for all different investigated decoders, including the regression (REG) head, classification (CLF) head, binary classification (BIN) head, and clustering (CLT) head. Grey blocks denote inference-only blocks, whereas for training, the raw logit outputs are used and passed to the loss function.}
    \label{fig:heads-detail}
\end{figure*}

All decoders are built around the same 2D attention block, with intermediate feed-forward and normalization layers as depicted in \cref{fig:encoder-detail}. Details on the individual heads are depicted in \cref{fig:heads-detail}.
In the main classification and regression decoder, each target attends to a single \ROWT token. Due to the softmax operation, a standard Multi-Head Attention applied over a KV sequence of length 1 produces an attention weight equal to 1.0. This ultimately reduces to a linear projection of V. Combined with the skip connection, the cross-column attention block output is therefore equivalent to the query input plus a linear projection of V (\ROWT token). However, this formulation allows the model to attend to multiple \ROWT tokens without any change to the architecture.

\subsection{Classification and Regression}
In this section, we describe the decoder architecture used for both classification and regression tasks. The decoder consists of the main backbone and two separate MLP-based heads: one dedicated to classification and the other to regression.

\custompar{Classification}
For classification, we employ a standard cross-entropy loss applied to the output of an MLP with a fixed output dimension of $64$, corresponding to the maximum number of supported classes. As a result, at inference time, the model cannot predict more classes than this predefined number without resorting to less efficient approaches such as hierarchical classification (e.g., as used in TabPFN’s many-class extension or TabICL).
To ensure the model can associate semantic meaning with class indices (e.g., class ``0''), it must learn an internal representation of these identifiers. To facilitate this, we introduce a dedicated input encoding specifically for the target column, in addition to the encodings used for feature columns. Although this design choice breaks class order permutation equivariance, we find it to be effective in typical few-class classification settings.

\custompar{Regression}
We encode numerical values via piecewise linear interpolation over a set of learned reference vectors.
Specifically, given the training data (context rows), we compute $n{=}50$ quantiles at orders $\frac{1}{2n}, \frac{3}{2n}, \dots, \frac{2n-1}{2n}$
and extend them with a (clipped) minimum and maximum, yielding $n{+}2$ reference points $q_0 < q_1 < \dots < q_{n+1}$.
Each reference point $q_i$ is associated with a learnable vector $v_i$.
A value $x \in [q_i, q_{i+1}]$ is then encoded as $(1{-}\lambda)\,v_i + \lambda\,v_{i+1}$, where $\lambda = \frac{x - q_i}{q_{i+1} - q_i}$;
values outside $[q_0, q_{n+1}]$ are clipped to the boundary.

For the regression target, this encoding is used on the input side during training.
On the output side, regression is cast as classification over the $n{+}2$ reference points:
the model produces a probability distribution $(p_0, \dots, p_{n+1})$ via a softmax, and the final prediction is the expectation $\sum_i p_i\, q_i$.

\subsection{Matching}
For the matching decoder training, we employ two separate pretraining and task generation strategies:

First, from a random table, we generate two column-wise splits with a random overlap of \textit{at most} 50\%.
These act as the matching Table A and Table B depicted in \cref{fig:architecture}.
With a random probability between 5\% and 33\% (drawn and fixed for the current table), we then mark each row pair as a match by drawing from the Bernoulli distribution.
This marking serves as the binary target label of the current sample.
We then keep the order of rows marked as matched in place, while randomly shuffling the remaining ones.
This strategy resembles cases of pairwise row matches with partially identical column entries.
That is, in most cases, the prediction can be made by exact matching, $\hat{y} = (A_i == B_j)$ for some column $A_i$ in Table A and $B_j$ in Table B respectively. 
However, note that this cannot trivially be exploited in our architecture as both rows are first encoded into latent \ROWT tokens which are contextualized over separate statistics and table schemas.
Hence, we found this strategy to provide reasonably challenging matching tasks that closely resemble many real-world matching tasks with exact or fuzzy column matches.

As for the second strategy, we use a more indirect semantic matching target:
Again, using a randomly drawn table, we use the same strategy as in the classification training to generate a feature-target split $X, \tilde{y}$ of a corresponding classification target.
We then again create two column-wise splits from the feature matrix $X$ with \textit{at least} 50\% overlap.
The resulting tables are then shuffled.
The ground truth matching target is then defined as the agreement of the original classification target between the rows in Tables A and B, $y = (\tilde{y}_A == \tilde{y}_B)$.
That is, two rows are considered matching, if they share their underlying classification target.
This results in a much more indirect, semantic matching task as opposed to the direct column matching above.

\subsection{Clustering}\label{sec:clustering_architecture}
For unsupervised clustering, most of the decoder architecture is inherited from the classification decoder, in order to minimize the number of parameters to be trained from scratch. Target embeddings are also inherited by using the \TRGT token in all of the rows, including the context. The prediction head is replaced by an MLP trained from scratch, outputting embedding vectors of dimension 256 for each row, which are further normalized.

Training data is sampled similarly to classification or regression data, by choosing a random table and selecting one random column as the target, with the following additional criteria (applied independently to each column, of any data type): missing values are dropped; only values that appear at least 5 times are retained; among those, only the 10 most numerous classes are retained. If, after these steps are applied, at least 50\% of the original rows survive, this column is used as the target. A random subset of the remaining columns is used as features, and the target is only used to construct labels (never shown to the model itself).

During training, the row-embedding vectors obtained above are transformed into a square matrix containing the cosine similarities $s_{ij}$ between any two rows (rescaled by a constant factor of 10). These values are then compared against the ground truth (1 meaning ``same hidden class'', 0 meaning ``different class''), that is:
\begin{equation} \label{eq:clustering_loss}
\text{Loss} = -\sum_{\substack{i = 1, \dots, R\\ j \in \mathrm{context}}} \Big[ \log ( s_{ij}) \delta_{c_i=c_j} + \log(1-s_{ij}) \delta_{c_i \neq c_j} \Big], \quad s_{ij} = \sigma\Big(10 \cdot \frac{x_i \cdot x_j}{\|x_i\|\|x_j\|}\Big)
\end{equation}
Note that the loss is computed over all rows $i$ (both context and query). In this way, the same trained model can support both the inductive (using context+query, with query rows not influencing each other) and transductive settings (using context only; the latter is the only setting supported by, e.g., TabClustPFN).

At inference time, the normalized embeddings are fed to K-means. This is a natural choice: for normalized vectors, the cosine similarity is directly linked to the Euclidean distance. For example, if a set of points achieves zero loss as computed in \eqref{eq:clustering_loss} (which is only possible if there are exactly two clusters), then embeddings of points in the same cluster will have distance 0, and those of points in different clusters will have distance 2 (being normalized vectors pointing in opposite directions), so they will be clustered correctly by K-means (or other Euclidean-distance-based algorithms). Also, K-means requires an explicit $k$ to be provided; we only implement and test the ``oracle'' setting where this is provided by the user.

\begin{figure*}
    \centering
    \includegraphics[width=\linewidth]{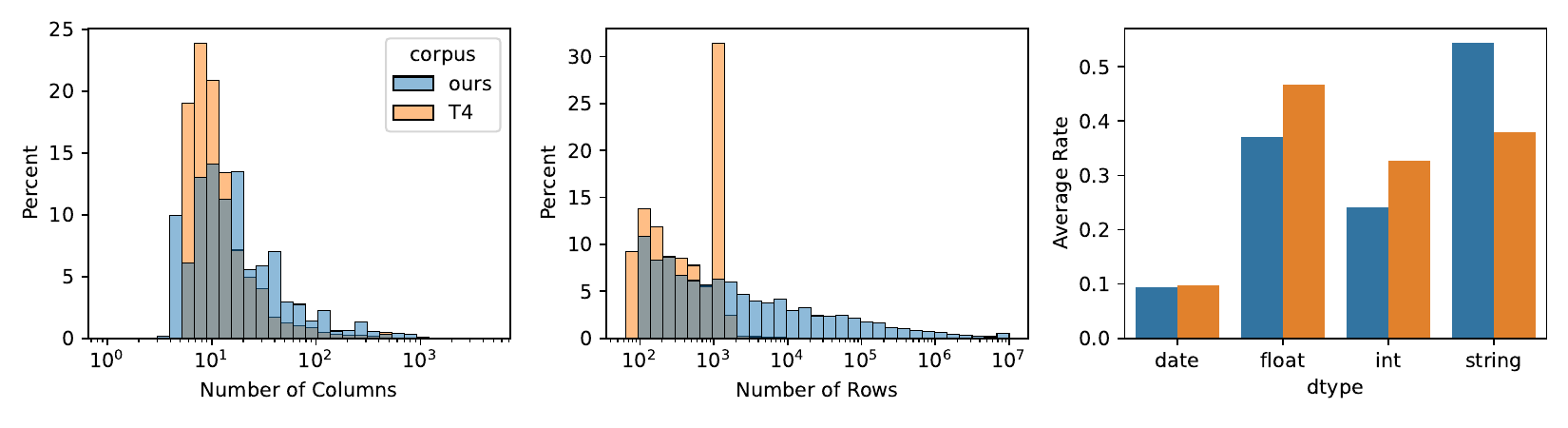}
    \caption{Column and row count distribution, as well as per-column data type rate of our pretraining data corpus and the T4 corpus. Our corpus consists of roughly 300\,k tables in total, sourced from various public sources. The T4 corpus was downsampled to 500\,k tables for its statistics calculation.}
    \label{fig:pretraining-distribution}
\end{figure*}

\begin{figure*}
    \centering
    \begin{subfigure}[t]{0.49\textwidth}
        \includegraphics[width=\linewidth]{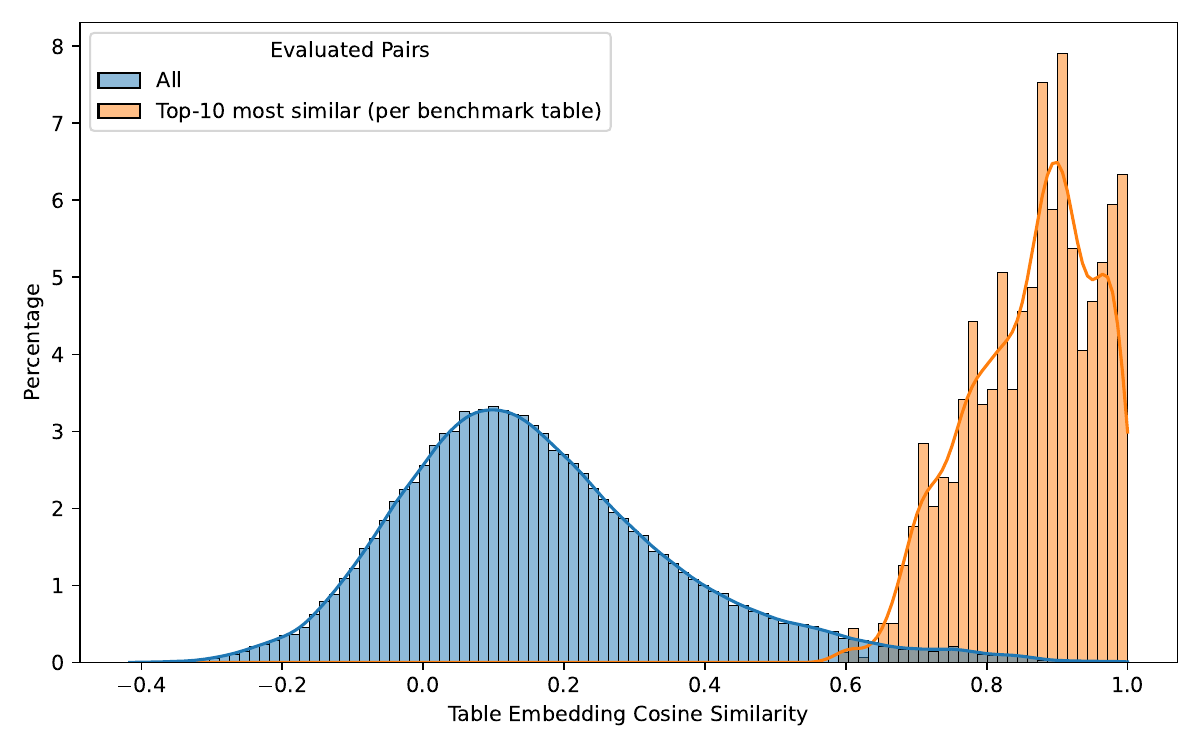}
        \caption{Cosine similarity distribution between the used pretraining tables and all evaluated benchmark tables}
        \label{fig:contamination-armadillo}
    \end{subfigure}
    \hfill
        \begin{subfigure}[t]{0.49\textwidth}
        \includegraphics[width=\linewidth]{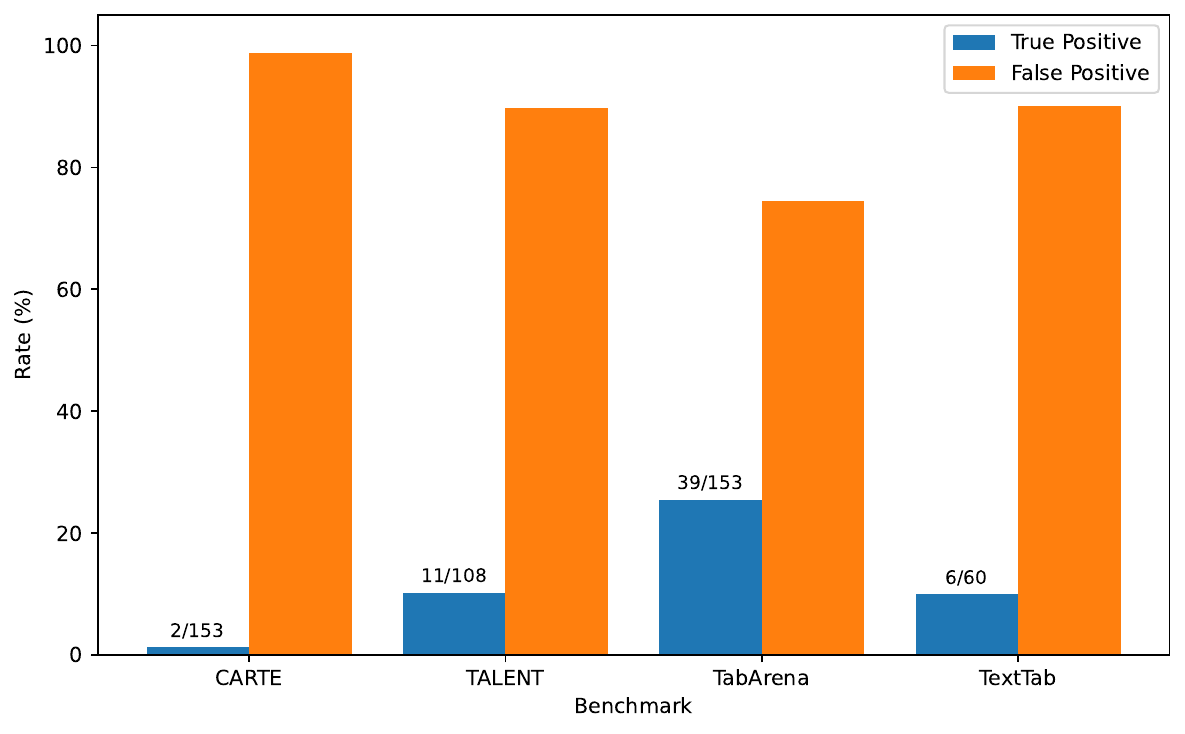}
        \caption{True and false positives among the top three most similar table pairs as estimated by Claude Opus 4.6}
        \label{fig:contamination-armadillo-true-pos}
    \end{subfigure}
    \caption{Table overlap and false positive rates based on Armadillo embeddings and false-positive estimated using Claude Opus 4.6. Out of those marked true positive, manual confirmation showed 48 confirmed true positive matches for 25 of the used evaluation tables.}

\end{figure*}

\subsection{Pretraining Data}
\label{app:pretrain-data}

\custompar{Data Statistics}
Our pretraining corpus consists of roughly \num{300}\,k raw, unlabeled tables.
These tables were sourced and curated at scale from various public sources with a permissive license. An overview of the size distribution is presented in \cref{fig:pretraining-distribution}, comparing the row, column, and data-type distributions of our corpus with those of The Tremendous TabLib Trawl (T4) collection~\cite{tabula8b}. We observe that our pretraining corpus is more broadly spread, particularly in terms of table width and height, and overall has a higher average rate of string columns. Note that these data type rates are averages of per-table rates. That is, we first measure the data type rate across columns of each table and then collect averages across tables.

\custompar{Contamination}
We perform a contamination study to estimate overlap between the pretraining corpus and the evaluated benchmark data for classification and regression. To that aim, we employ two different methodologies.

\emph{Row-wise Hashing:} This scalable approach looks at the full pretraining data, by flagging tables where rows match exactly (up to column permutation; table headers are also ignored). To that aim, we first convert all cells to strings, then encode them to 64 bits via the xxHash algorithm \cite{xxhash_github}. These hashes are then interpreted as \texttt{uint64} numbers, i.e., integers modulo $2^{64}$, and the final (permutation-invariant) fingerprint of each row is obtained by summing them together. We then check for duplicates by first filtering for tables having the same number of columns, sorting the fingerprints for all the pretraining data matching this condition, and finally looking for the (much fewer) fingerprints appearing in evaluation tables via binary search. Evaluation tables that match at least 10\% of their rows to a given table are flagged as leaking. This is the strongest lower bound, flagging tables that are reported verbatim (including both features and targets) in the training data. It is, however, not robust to some minor distortions, like type casting (e.g., \texttt{int} to \texttt{float}) or dropping one or more columns.

This exact equality highlighted that 8 of the 158 evaluation tables do appear in the pretraining data; 4 of these are part of the TabArena benchmark, and 4 more of the TALENT-Tiny one.

\emph{Table-level Embeddings:} As a more robust, but less scalable approach, we perform a pair-wise comparison between all used evaluation datasets and a maximum 10\,k-row sample of each pretraining dataset using table-level embeddings obtained via Armadillo~\cite{pugnaloni2025armadillo}.
Armadillo is a Graph Neural Network pretrained such that the cosine similarity between two table embeddings corresponds to their estimated overlap.
Note that the obtained embeddings are invariant against permutations of rows and columns, but trained for exact, non-fuzzy matches.

That is, we calculate table-level embeddings for all 158 tables used during evaluation (we focus on the classification and regression datasets here) and all roughly 300\,k tables used during pretraining.
We then calculate the pair-wise cosine similarity between all embeddings and collect the top 10 most-similar matches for each evaluation table. 
The resulting similarity distribution is shown in \cref{fig:contamination-armadillo}.
Overall, the embeddings obtained via Armadillo show a relatively large overlap, with a median of about \SI{14}{\percent}.
Among the most-similar matches, we observe a much higher similarity, indicating that almost all tables have at least one matching pretraining table with more than \SI{50}{\percent} estimated overlap, according to the Armadillo embeddings.
However, importantly, a manual inspection showed many non-matching pairs despite the high degree of estimated overlap.
Hence, as a further step, for each evaluation table, we passed the top three most similar found matches to an AI agent (Claude Code using Claude Opus 4.6) to estimate whether the pair is an actual match or a false positive.
The resulting distribution is depicted in \cref{fig:contamination-armadillo-true-pos}.
As a final verification step, we then manually went through all pairs marked as true-positives and found 48 pairs of those to be actually true positives.
However, even out of those, nine matches do not contain the original task's target column but are matches in the raw data instead.

Concluding, in total, we found matching pretraining tables for 25 of the used evaluation tables (including the 8 found with the previous methodology), in detail:
\begin{itemize}
    \item 2 from the CARTE benchmark (with one having only matches that do not contain the target)
    \item 15 from TabArena
    \item 2 from TextTab
    \item 6 from TALENT-Tiny (with one having only matches that do not contain the target)
\end{itemize}

In particular, the high number of numerical tasks in TabArena accounted for many false positives due to the high probability of any large mostly-numeric table to have a large cell-wise overlap.

While this overlap is considerable, we believe the impact on our reported results to be small:
Each table is sampled only a small number of times during training and subsequently augmented, subsampled, and used with a random prediction target.
Given the comparatively small model size and in-context learning training objective, we believe that current tabular in-context learners, including our proposed architecture, do not suffer or gain from in-weight data memorization in any meaningful way.
We can observe this with our model performing noticeably better than the state of the art on the CARTE benchmark, where we only found contamination for 2 of the used evaluation datasets, whereas the performance on TabArena is relatively worse despite the much larger amount of contamination found.

\begin{figure*}
    \centering
    \includegraphics[width=.85\linewidth, valign=c]{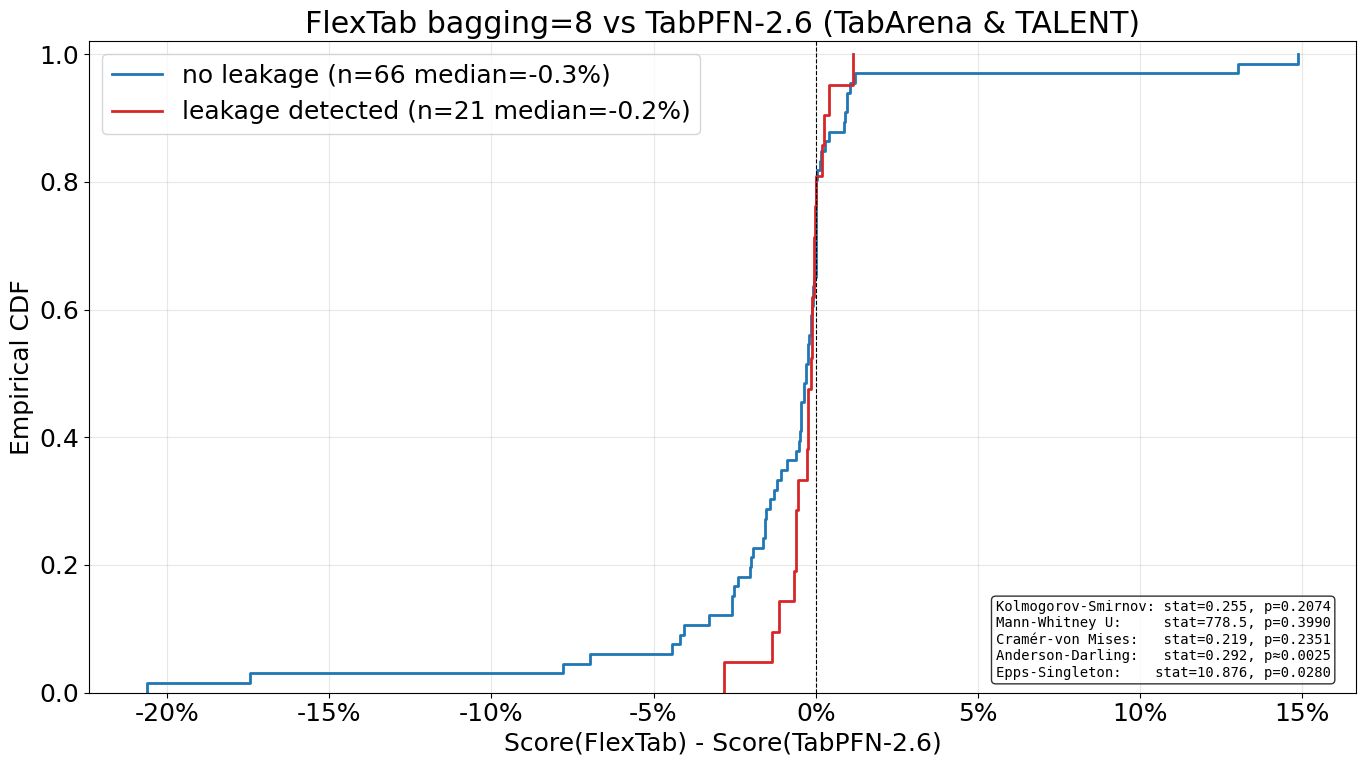}
    \caption{Cumulative distribution functions of the differences between FlexTab scores vs.\ TabPFNv2.6 scores on the contaminated (red) and non-contaminated (blue) subsets of numeric-heavy datasets.}
    \label{fig:cdf_contamination}
\end{figure*}

To confirm this hypothesis, we have performed a comparative analysis on the downstream performance on classification and regression tasks. We compared our FlexTab model against TabPFNv2.6 that, being trained on synthetic data, serves as a contamination-free baseline. We analyze separately the performance on contaminated and non-contaminated subsets of the evaluation datasets, compute the score difference between the two models, and investigate whether the two resulting distributions are statistically distinguishable. Given that the contamination is mostly present on the numeric-heavy datasets (TabArena and TALENT-Tiny) where our model is comparatively weaker and less on text-heavy ones (CARTE, TextTab) where it is stronger, this analysis only focuses on the former datasets, to avoid biasing the results. The cumulative distribution function of the differences are plotted in \cref{fig:cdf_contamination}. While the two distributions differ on the tails (as highlighted by the low $p$-value of the Anderson-Darling test), there is no consistent bias of FlexTab performing better on average on contaminated datasets. The median score delta only drops from a \SI{-0.2}{\percent} (non-contaminated) to a \SI{-0.3}{\percent} (contaminated) gap, well within statistical noise. This is confirmed by statistical tests that are more sensitive to the center of the distribution, like Kolmogorov-Smirnov ($p = 0.21$) and Cramér-von Mises ($p=0.24$), neither of which suggests a systematic difference between the two distributions.

Overall, while we have performed due diligence to estimate the amount of contamination in our pretraining dataset, estimating the impact on training and downstream performance is a challenging problem.
Generally, as the community moves forward, investigating tabular models trained on real-world data, the topic of benchmark contamination with pretraining data used to train large tabular models is an open and important research topic.

\subsection{Baselines}
\label{app:baselines}
Overall, we follow the evaluation protocol and baseline implementations of ConTextTab~\cite{contexttab}, but with updated baseline versions:

\custompar{TabPFN} We evaluate TabPFNv2.6 using the official PyPI package version 7.1.1.
For classification tasks with more than 10 classes, we use the many-class extension from \texttt{tabpfn-extensions} in version 0.2.2+ at the latest git commit \texttt{452708bdf2e644dd1b2ebe938862a9ea52ac7e21} which adds compatibility with TabPFNv2.6.
We use default initialization parameters, in particular, using 8 ensembled estimators for inference.

\custompar{TabICL} We evaluate the recent TabICLv2 model, using the PyPI package version 2.0.1.
As opposed to the original TabICL variant, TabICLv2 natively supports both classification and regression.
We use default initialization parameters, in particular, using 8 ensembled estimators for inference.

\custompar{ConTextTab} We evaluate the recent ConTextTab model using the official repository and checkpoints at git version tag v1.1.2\footnote{\URL{https://github.com/SAP-samples/contexttab/}}.

\custompar{AutoGluon} We evaluate AutoGluon version 1.5 in the \texttt{extreme} preset with a time limit of 1h, fitted on a 40-core node with 320\,GB of RAM and a single H100 GPU.
We also evaluated the \texttt{best} preset with a time limit of 4h but found the former to be slightly better performing, likely due to the better use of recent tabular in-context learners in the model pool.

\custompar{PytabKit} Again, we follow the setup in ConTextTab, but use the recent 1.7.3 version of the PyPI \texttt{pytabkit} package~\cite{realmlp} with updated search spaces following the TabArena setup~\cite{tabarena}. 
For RealMLP, we report the best-in-class tuned Caruana-ensembled results~\cite{tabarena}.
For the extended results, we also show the tuned-defaults (TD) variants.

\custompar{Scikit-learn} For the naive, KNN, and Random Forest baselines, we use the corresponding scikit-learn~\cite{sklearn} estimators with their default settings. We use scikit-learn version 1.5.2.

\custompar{LimiX} In the extended results, we also report the evaluation of LimiX (2M and 16M) using the corresponding retrieval configs.
We used the code from the official repository\footnote{\URL{https://github.com/limix-ldm-ai/LimiX}} at the latest commit hash \texttt{375b49eb1d377620943aa93e4654a63e542f0d2b}.

\subsection{Evaluation Metrics}
For all metric calculations, we use the corresponding scikit-learn implementations~\cite{sklearn}.

For classification and regression results, we report mean accuracy and soft-clipped $R^2$: due to the dominance of outliers, we soft-clip the negative $R^2$ values to the range $[-1, 0)$ via $\operatorname{tanh}$, which is smoothly differentiable and retains the relative ordering of negative scores~\cite{contexttab}.

For the calculation of the Elo scores, we utilize the implementation of TabArena~\cite{tabarena}, which itself is based on ChatBot Arena's implementation~\cite{chiang2024chatbot}.

For all ranking results (aside from those used in the critical difference diagrams), we use a robust ranking algorithm.
That is, models that lie within \num{0.1} percentage points (in terms of either Accuracy, AUROC, or $R^2$) are considered ties.
The same robustification also applies to our Elo score and win-ratio calculations.

For the critical difference (CD) diagrams, we use the \texttt{autorank} library~\cite{autorank}.
So as not to influence the statistical guarantees of the significance test used, we do not alter or robustify its internal rank calculation.
Hence, the ranks shown in the CD diagrams can deviate from those shown in the result tables or figures.

\subsection{Datasets}
\custompar{Classification and Regression}
We evaluate all models on CARTE, TextTab, TabArena-Lite, and TALENT-Tiny.
We use the same split creation protocol as ConTextTab, except for TabArena-Lite, for which we use the predefined splits.

\custompar{Matching}
We evaluate on the synthetic Febrl4 dataset from the Freely Extensible Biomedical Record Linkage project~\cite{febrl}, as well as Fodors-Zagats, Bikes, eBooks, and Movies sourced from the DeepMatcher study~\cite{mudgal2018deepmatcher} and the Magellan Data Repository~\cite{magellandata}.
Due to the relatively small size of these datasets with available ground truth match labels, we perform \num{5}-fold cross-validation over all available labeled examples and report average match-class F1 scores.

\custompar{Relational Prediction}
We use the RelBench benchmark from the official \texttt{relbench} PyPI package in the latest version 2.1.1 using the predefined splits.

\input{tables/classif_reg_extended}

\section{Additional Results}
\label{app:add-results}

\subsection{Classification and Regression}
\label{app:add-results-clf-reg}

\custompar{Additional Baseline Evaluations}
Additional results including LimiX, and other gradient-boosted tree variants and scikit-learn estimators, are shown in \cref{tab:tabular-benchmarks-extended}.
The overall results are in line with the selected presentation in the main paper and serve only to illustrate a complete picture.
Across all evaluated models and benchmarks, FlexTab ranks best overall and performs particularly well on semantic-heavy benchmarks.
This is in line with the results of ConTextTab, which our architecture builds upon, while narrowing the gap on numeric-heavy regression, such as on TabArena-Lite, but remaining behind in classification.

\begin{figure*}
    \centering
    \begin{subfigure}[b]{\textwidth}
        \hspace{14mm}\includegraphics[width=0.85\linewidth]{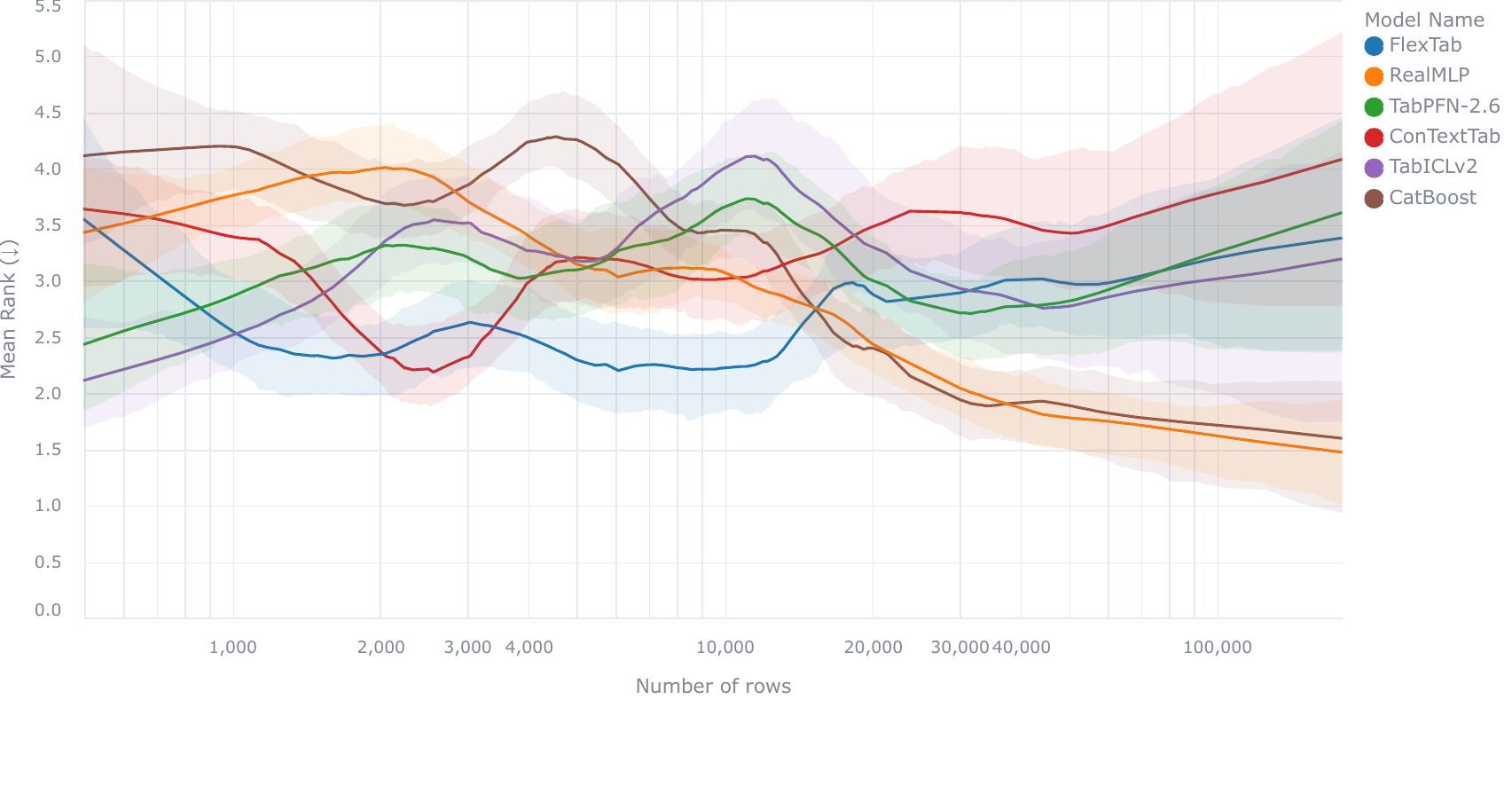}
        \vspace{-2em}
        \caption{Performance dependence on row count.}
    \end{subfigure}\\[3mm]
    \begin{subfigure}[b]{\textwidth}
        \hspace{14mm}\includegraphics[width=0.85\linewidth]{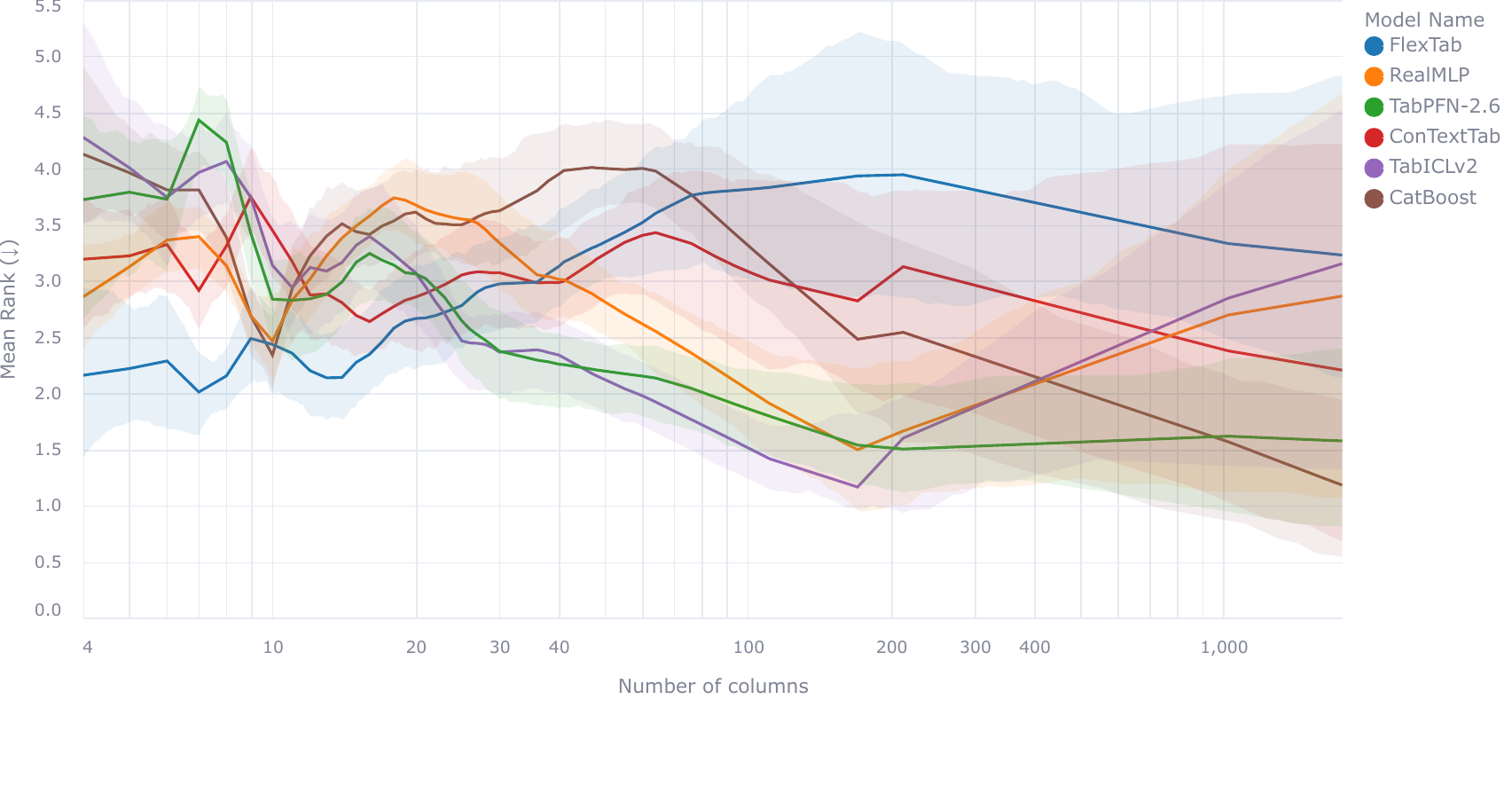}
        \vspace{-2em}
        \caption{Performance dependence on column count.}
    \end{subfigure}
\caption{Relation between number of training dataset rows (top) and columns (bottom) and model performance, obtained as a LOWESS regression in the plane $\log(n_{\mathrm{rows/cols}},\textnormal{rank})$. The confidence bands are the 80\% confidence intervals obtained via bootstrapping.}
\label{fig:rank_vs_num_rows_cols}
\end{figure*}

\custompar{Relation between Dataset Size and Model Performance}\label{app:additional-results-data-size}
We plot the average rank of each model as a function of the dataset size expressed in number of rows and the number of columns in \cref{fig:rank_vs_num_rows_cols} across all evaluated datasets. 

We observe that FlexTab performs particularly well in the 1\,k to 10\,k row count regime.
For smaller tables, TabICL and TabPFN tend to perform better, but we also note the large confidence intervals due to the comparatively small support set in those areas.
For larger tables, the in-context learners tend to converge to a similar performance and start being outperformed by CatBoost and RealMLP which are extensively trained and tuned for each dataset.

In terms of the dependence on the number of columns or features of each task, we observe that FlexTab excels at very narrow tables but shows decreasing performance for wider ones (relative to the other models).
Note again the potentially small support set and wide confidence bands; ranking scores can exaggerate such differences, and the absolute differences may not be as large.
Nevertheless, this points to a potential bottleneck in our architecture:
As we compress each row into a single embedding that is fed to the prediction decoder, wider tables naturally undergo a relatively larger compression than narrow ones.
This should be investigated in future work, for example, by increasing the hidden dimension or investigating hybrid approaches that utilize multiple tokens per row for wider tables.

\begin{figure*}
    \centering
    \begin{subfigure}[b]{\textwidth}
        \includegraphics[width=0.39\textwidth]{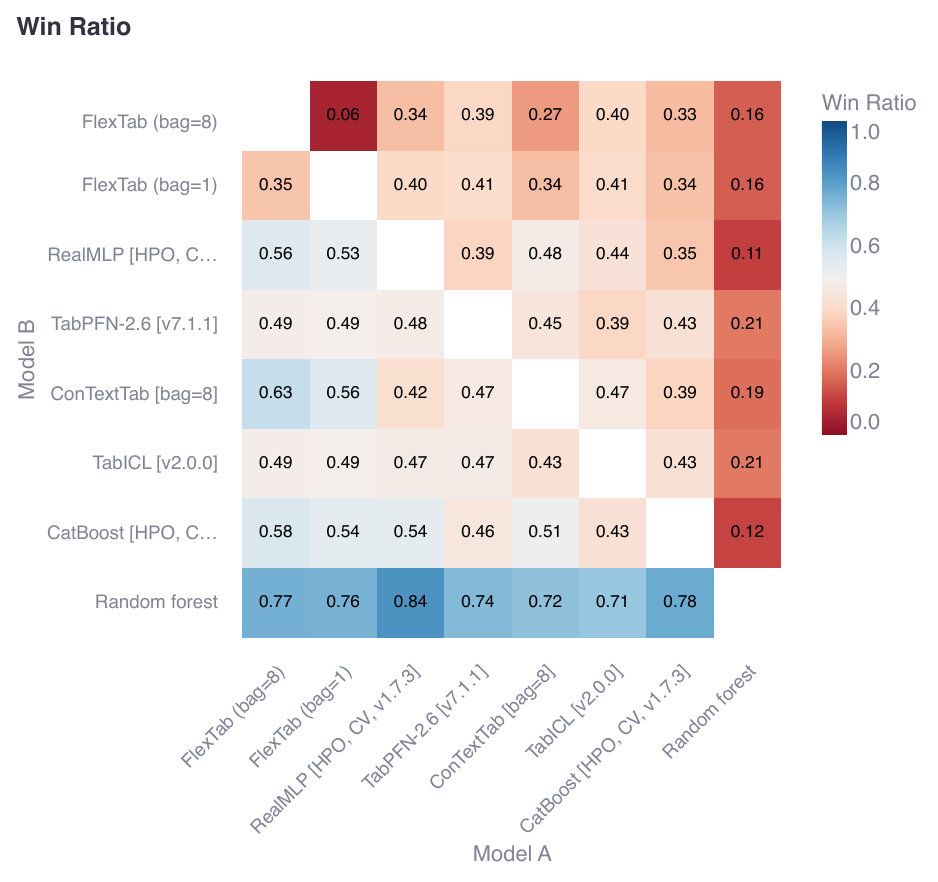}\hfill
        \includegraphics[width=0.59\textwidth]{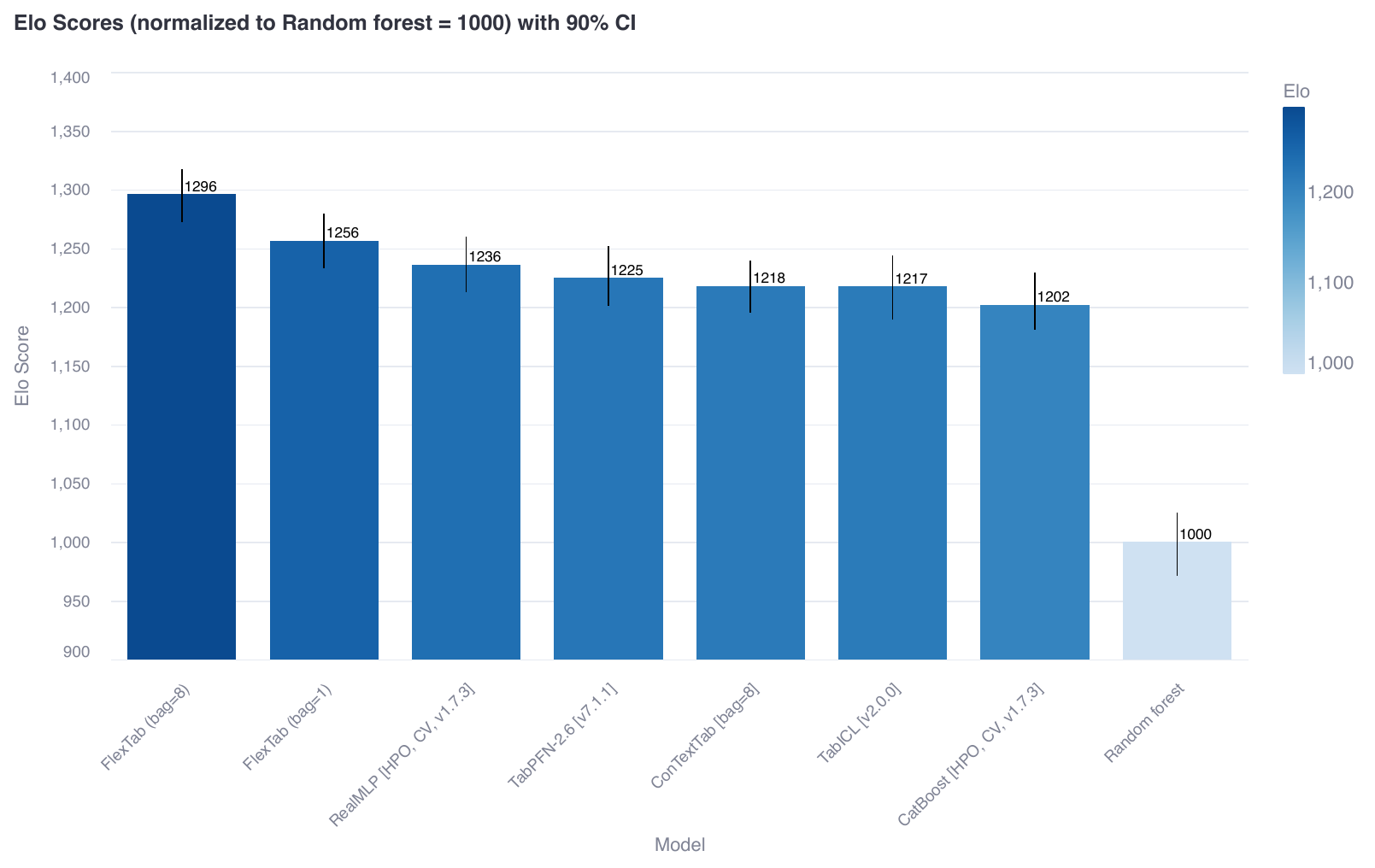}
        \caption{Win ratio matrix with Model A wins over Model B (left) and Elo scores (right).}
    \end{subfigure}\\[1mm]
    \begin{subfigure}[b]{0.8\textwidth}
        \includegraphics[width=\textwidth]{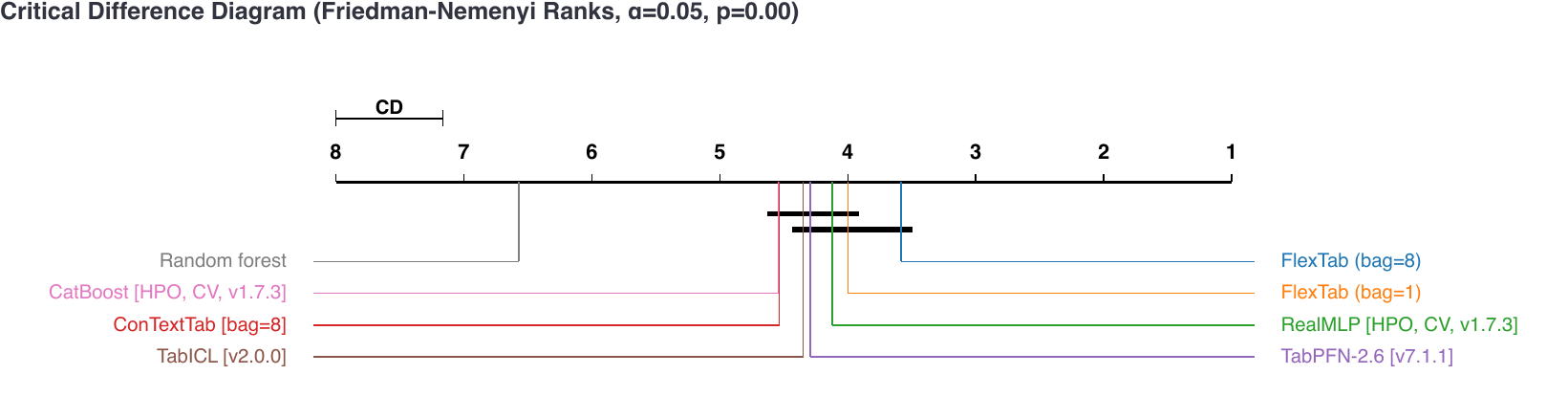}
        \caption{Critical difference (CD) diagram.}
    \end{subfigure}
\caption{Win ratio confusion matrix, Elo scores, and CD diagram of the main investigated models across all 158 datasets from CARTE, TabArena-Lite, TALENT-Tiny, and TextTab benchmarks.}
\label{fig:battle-scores-main}
\end{figure*}

\custompar{Win ratios, CD diagrams, and Elo scores} Additionally, we show the win ratio confusion matrix, CD diagrams, and Elo scores across all investigated benchmarks (CARTE, TabArena-Lite, TALENT-Tiny, and TextTab) in \cref{fig:battle-scores-main}. Wins are calculated based on accuracy for classification and $R^2$ for regression datasets. Ties are not counted as wins.

\begin{figure*}
    \centering
    \begin{subfigure}[b]{\textwidth}
        \includegraphics[width=0.39\textwidth]{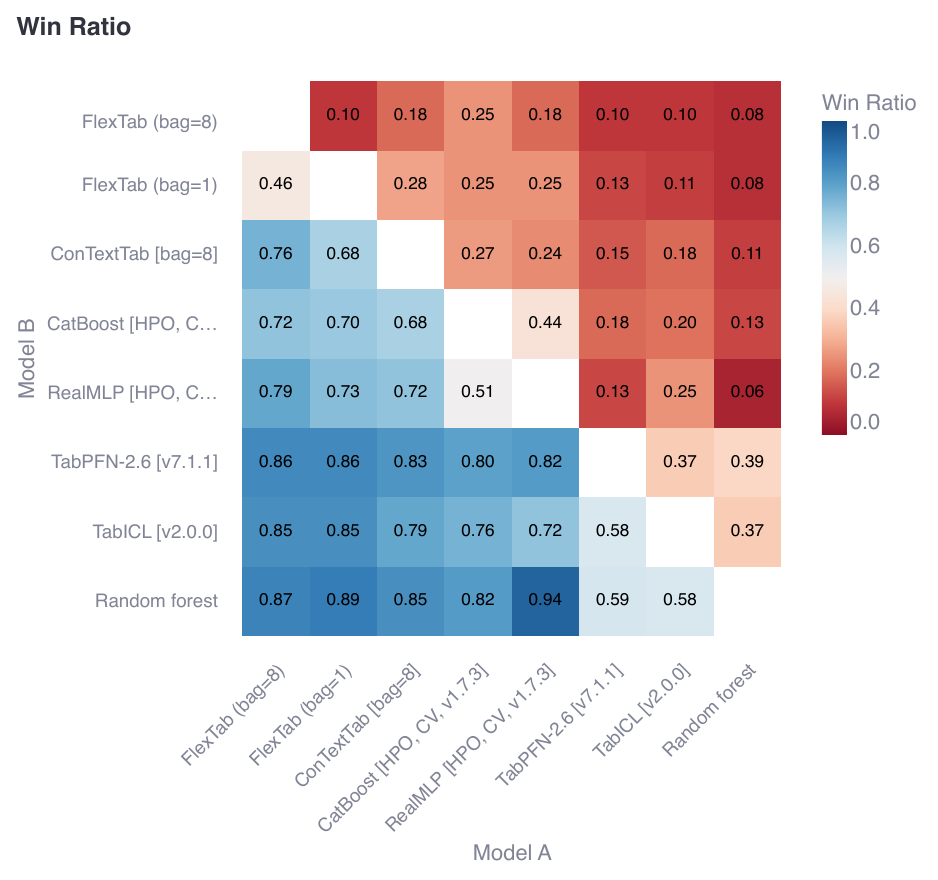}\hfill
        \includegraphics[width=0.59\textwidth]{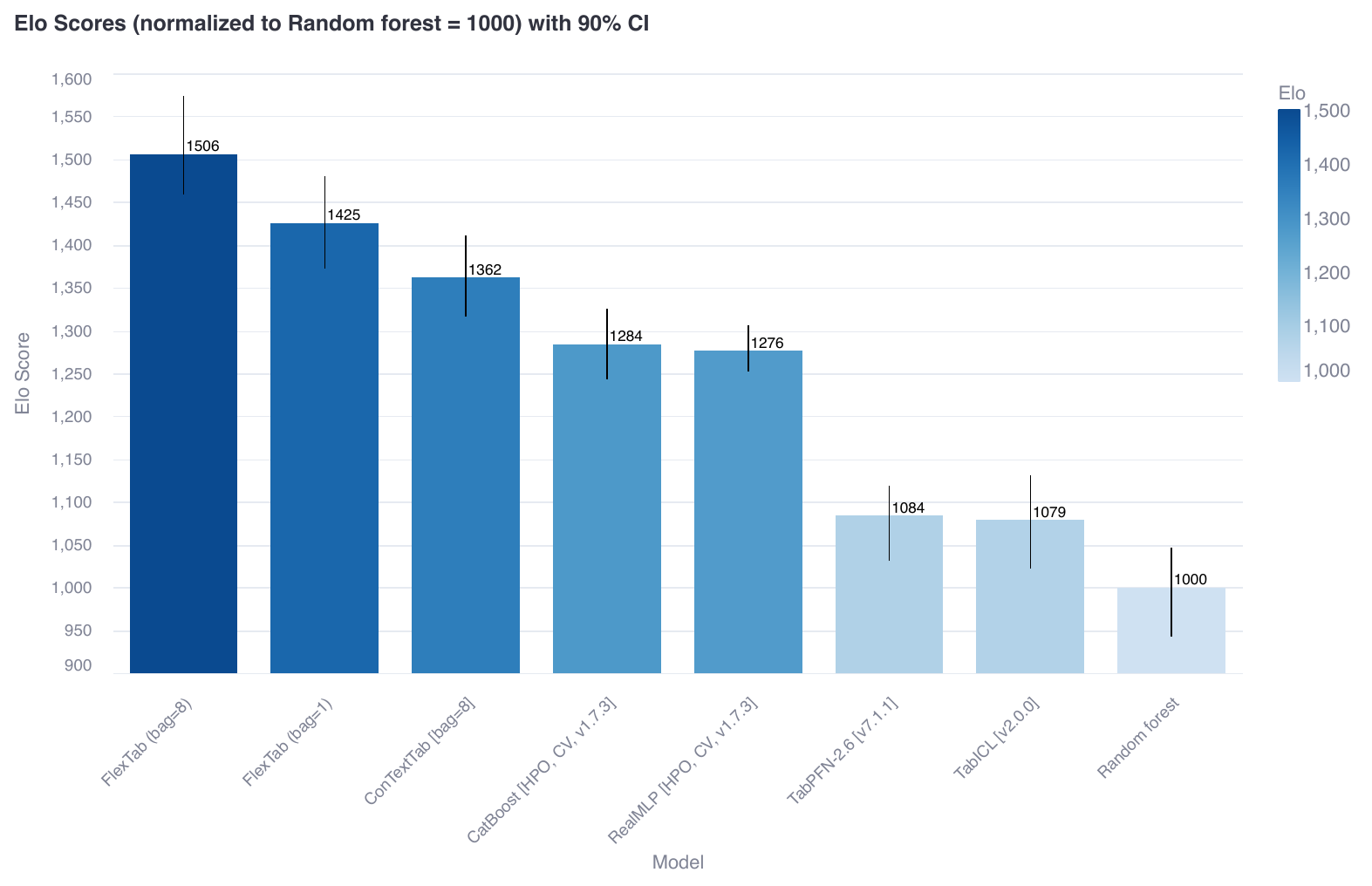}
        \caption{Win ratio matrix with Model A wins over Model B (left) and Elo scores (right).}
    \end{subfigure}\\[1mm]
    \begin{subfigure}[b]{0.8\textwidth}
        \includegraphics[width=\textwidth]{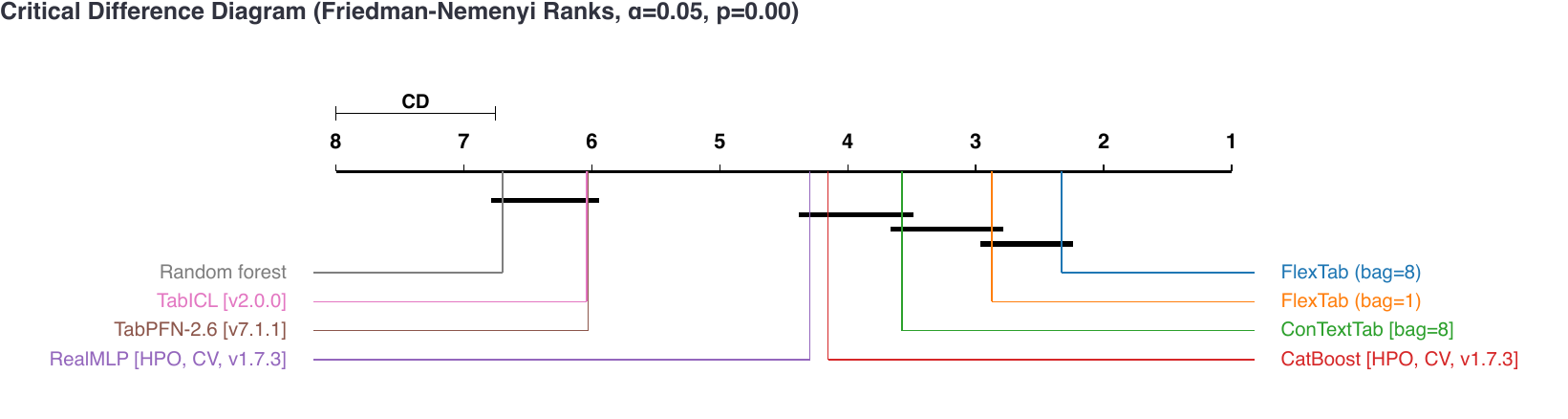}
        \caption{Critical difference (CD) diagram.}
    \end{subfigure}
\caption{Win ratio confusion matrix, Elo scores, and CD diagram of the main investigated models across all 71 semantic-heavy datasets from CARTE and TextTab benchmarks.}
\label{fig:battle-scores-semantic}
\end{figure*}

\begin{figure*}
    \centering
    \begin{subfigure}[b]{\textwidth}
        \includegraphics[width=0.39\textwidth]{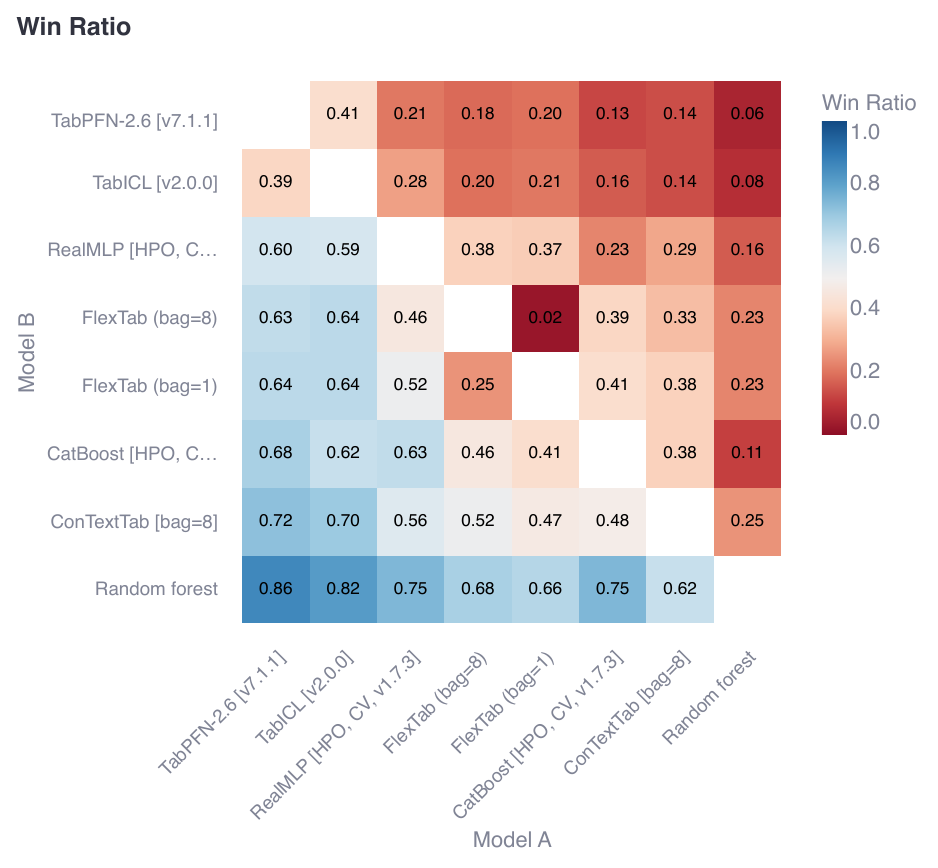}\hfill
        \includegraphics[width=0.59\textwidth]{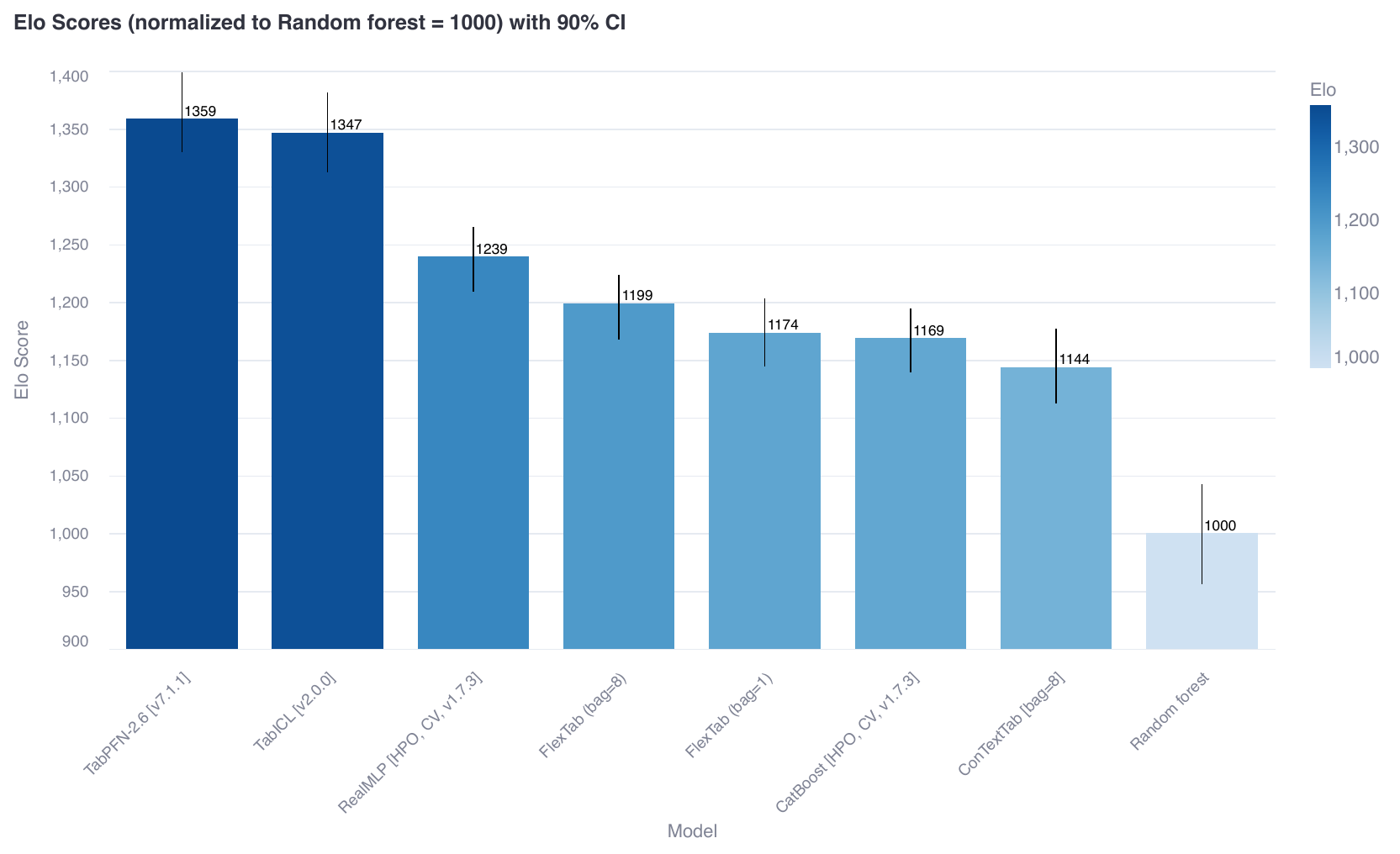}
        \caption{Win ratio matrix with Model A wins over Model B (left) and Elo scores (right).}
    \end{subfigure}\\[1mm]
    \begin{subfigure}[b]{0.8\textwidth}
        \includegraphics[width=\textwidth]{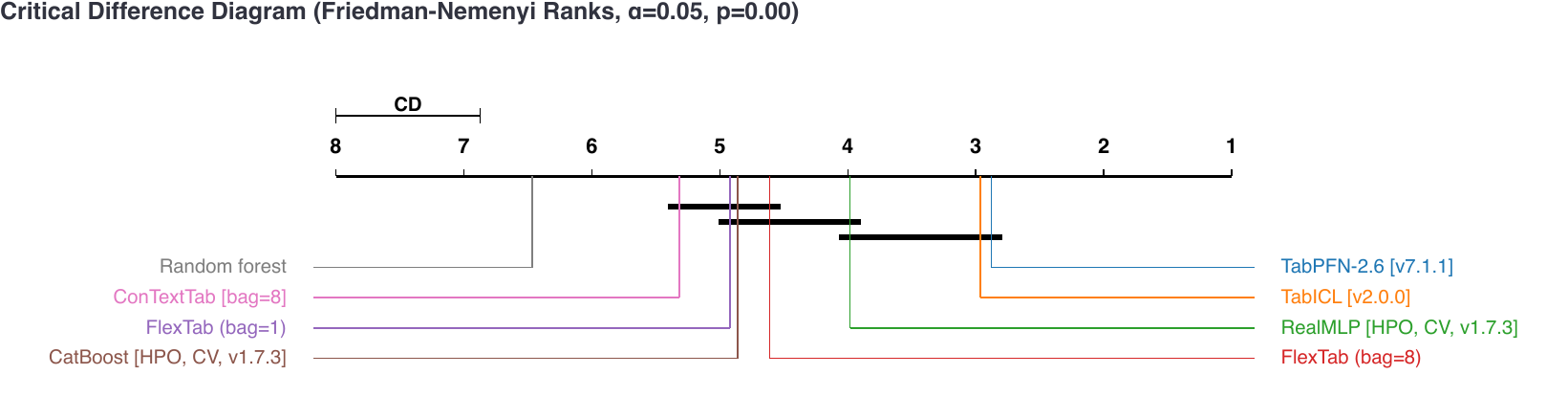}
        \caption{Critical difference (CD) diagram.}
    \end{subfigure}
\caption{Win ratio confusion matrix, Elo score, and CD diagram of the main investigated models across all 87 numeric-heavy datasets from TabArena-Lite and TALENT-Tiny benchmarks.}
\label{fig:battle-scores-numeric}
\end{figure*}

For fine-grained insights, we also plot per-benchmark-type diagrams:
\begin{itemize}
    \item for semantic-heavy benchmarks (CARTE and TextTab) in \cref{fig:battle-scores-semantic},
    \item for numeric-heavy benchmarks (TabArena-Lite and TALENT-Tiny) in \cref{fig:battle-scores-numeric}.
\end{itemize}

Across all benchmarks investigated, FlexTab shows very strong performance, outperforming other baselines, even hyperparameter-optimized ones with inner CV ensembles. The ensembled FlexTab~[bag=8] variant achieves the highest Elo score across the full set of \num{158} datasets (\cref{fig:battle-scores-main}), followed closely by FlexTab~[bag=1], with the critical difference diagram confirming that both variants rank significantly above Random Forest, CatBoost, ConTextTab, and TabICLv2.

The per-benchmark-type analysis reveals a more nuanced picture. On the \num{71} semantic-heavy datasets from CARTE and TextTab (\cref{fig:battle-scores-semantic}), FlexTab's advantage is particularly pronounced: FlexTab~[bag=8] reaches an Elo of \num{1506}, more than \num{100} points ahead of the next-best in-context learner (ConTextTab) and more than \num{400} points ahead of TabPFNv2.6 and TabICLv2, whose handling of semantics is more limited. The corresponding critical difference diagram shows that both FlexTab variants significantly outperform all other methods. On the \num{87} numeric-heavy datasets from TabArena-Lite and TALENT-Tiny (\cref{fig:battle-scores-numeric}), the picture is tighter: TabPFNv2.6 and TabICLv2 take the top two Elo positions, with FlexTab ranking fourth but remaining competitive with all tuned and ensembled per-dataset baselines. Notably, the critical difference diagram indicates that the top three models in this regime are statistically indistinguishable.

These results show that FlexTab is the strongest method overall across the investigated benchmarks, with a clear and statistically significant edge on semantic-heavy tasks and competitive performance on numeric-heavy ones. This supports our central claim that target-agnostic row embeddings, combined with task-specific decoders, provide an effective and broadly applicable foundation without sacrificing performance on the well-studied classification and regression regimes.

\subsection{Anomaly Detection}
\label{app:add-results-outlier}

The checkpoint has been selected via early stopping based on a manually curated validation set of 89 datasets from standard classification (52) and regression (37) tasks (including CARTE, TabArena, cc18, ctr23, and TextTab). For classification data, rare classes were chosen as anomalies and common classes as normal points; for regression data, after plotting a distribution with density smoothing, an antimode between two local clusters (if present) is chosen as the split value. In both cases, datasets are retained only if one of the thus-constructed classes is substantially smaller than the other. The three scenarios (unsupervised, one-class, and semi-supervised) are all evaluated during validation, and the best checkpoint by average score is selected; this corresponds to 7.2M training steps. For the semi-supervised setup, we also test the original FlexTab version trained for classification and regression. To distinguish between the two, we denote the version trained for anomaly detection as FlexTab-AD.

\begin{figure*}
    \centering
    \includegraphics[width=.48\linewidth, valign=c]{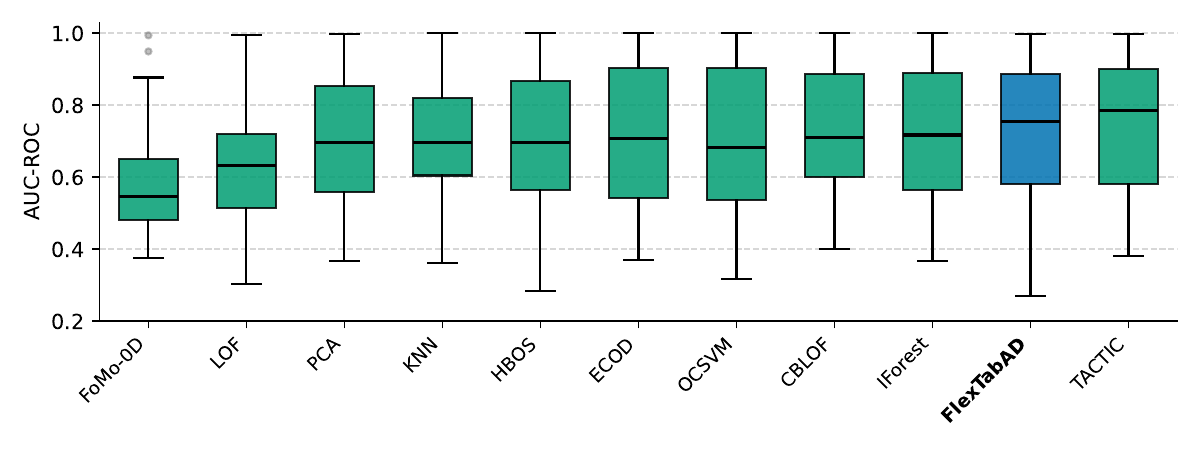}
    \includegraphics[width=.48\linewidth, valign=c]{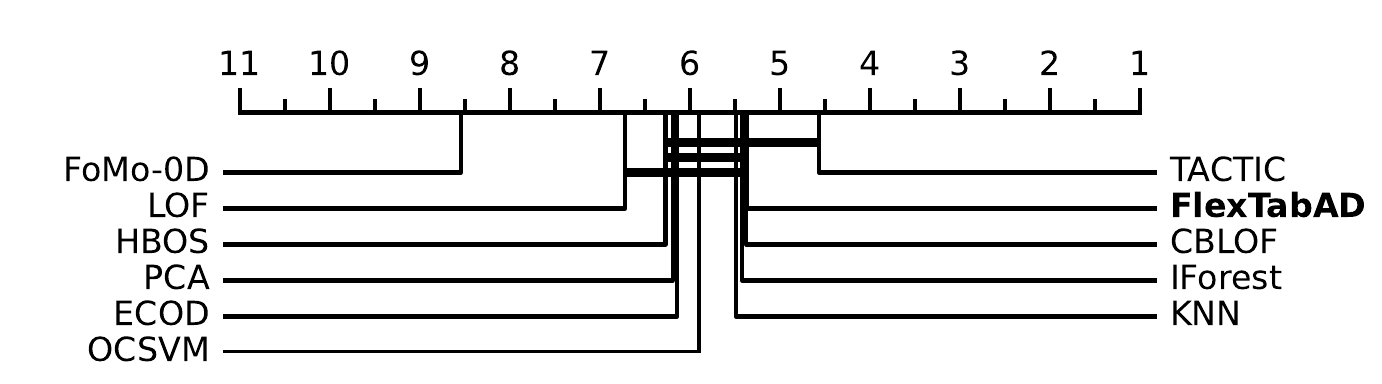}\\[1em]
    \includegraphics[width=.98\linewidth, valign=c]{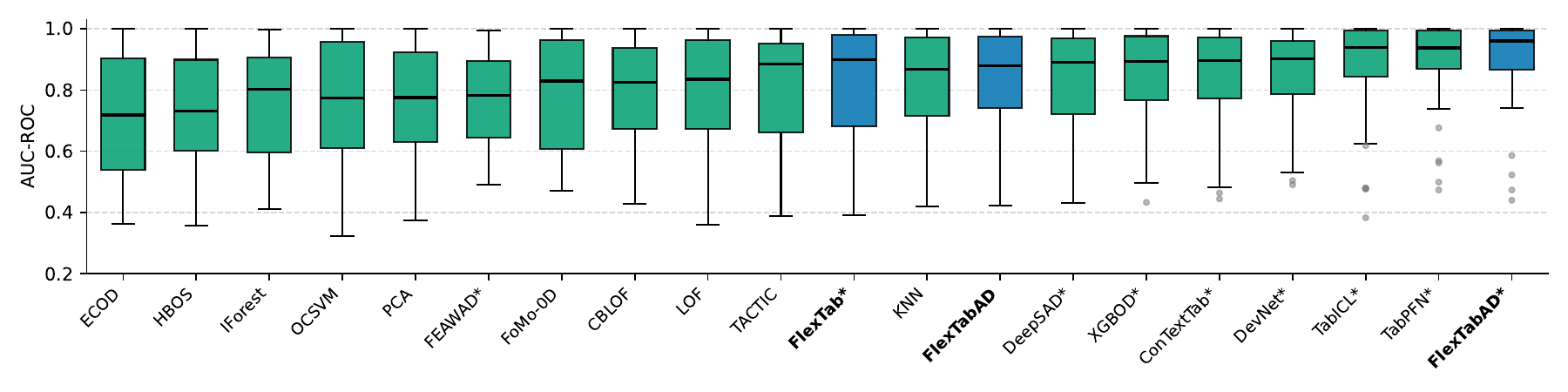}
    \includegraphics[width=.7\linewidth, valign=c]{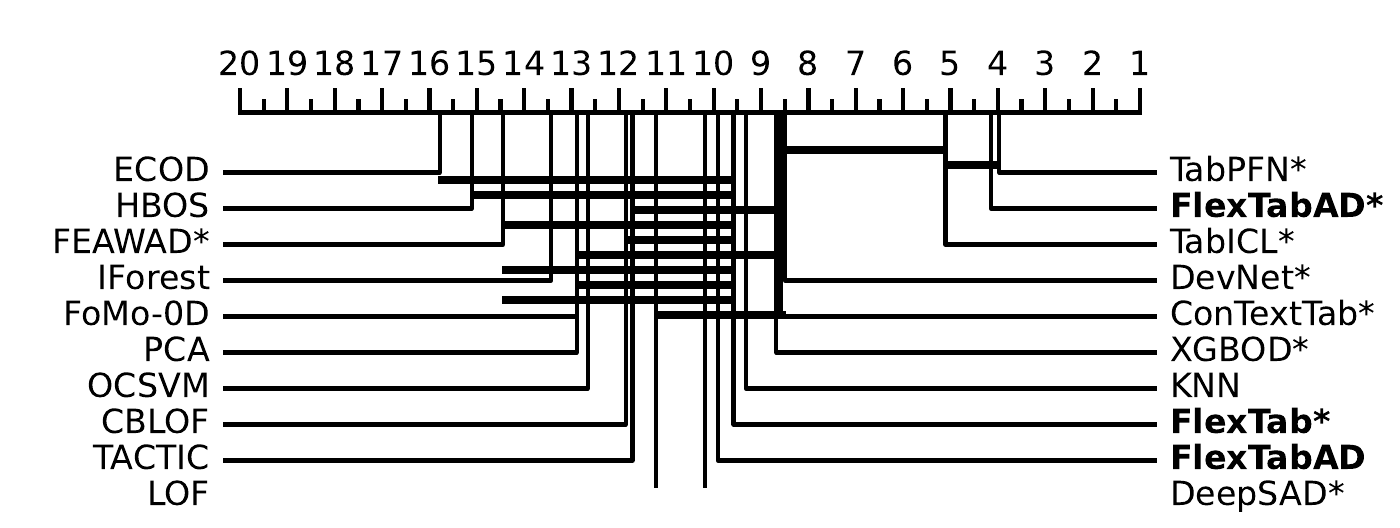}
    \caption{Results for unsupervised anomaly detection (top) and one-class + semi-supervised novelty detection (bottom). The models in the second group (i.e.\ those allowed to see the test set in a transductive manner) are marked with a (*). Left: box and whisker plots, sorted by average score. Right: critical difference diagrams based on \cite{autorank}.}
    \label{fig:unsupervised_oneclass_outlier_detection}
\end{figure*}

\custompar{Unsupervised}
The unsupervised setup (where no labels are provided to the model) is a notoriously challenging one. The results are reported in \cref{fig:unsupervised_oneclass_outlier_detection} (top). Overall, TACTIC slightly outperforms competing models across all considered AUROC-based metrics (average, median, and rank), with FlexTab scoring a close second. However, by and large, we share the same conclusion as in the original ADBench paper: in the unsupervised setting, essentially all methods end up being statistically equivalent, as the critical difference diagram in the top-right part of the figure highlights.

\custompar{One Class and semi-supervised novelty detection}
We group the results for these two settings, because the labeled data is the same in both cases; the only difference is that models in the latter category are allowed to operate in a transductive manner (i.e. to see all the test points at once and make use of that information). This changes the class of models that can be used entirely: in the inductive setting, only dedicated models can be used. Among these, the traditional KNN method appears as valid a choice as any more sophisticated approach, among which FlexTab-AD performs the best (with best overall median score in this class). Once test points can be exploited, though, one can obtain significantly higher accuracy by means of any binary classifier. In this setting we can see that the two best models are TabPFNv2.5 and FlexTab-AD. They have very close average rank, only rivaled by TabICLv2. FlexTab-AD scores first in terms of average and median AUC.

\begin{figure*}
    \centering
    \includegraphics[width=.48\linewidth, valign=c]{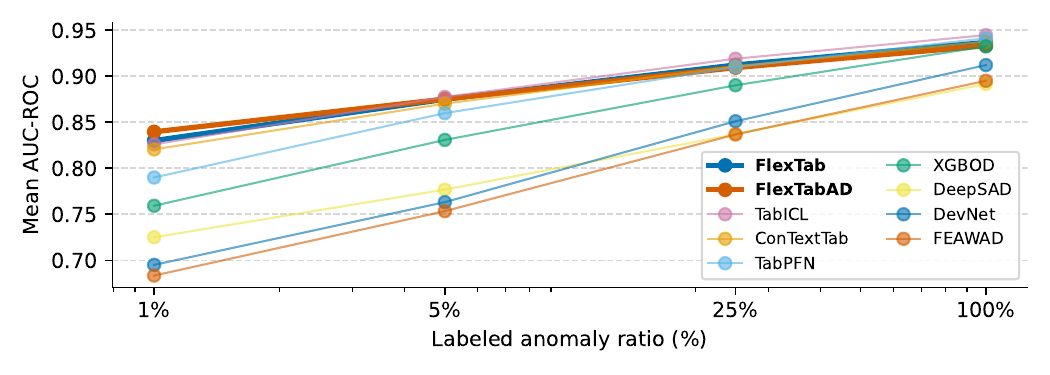}
    \includegraphics[width=.48\linewidth, valign=c]{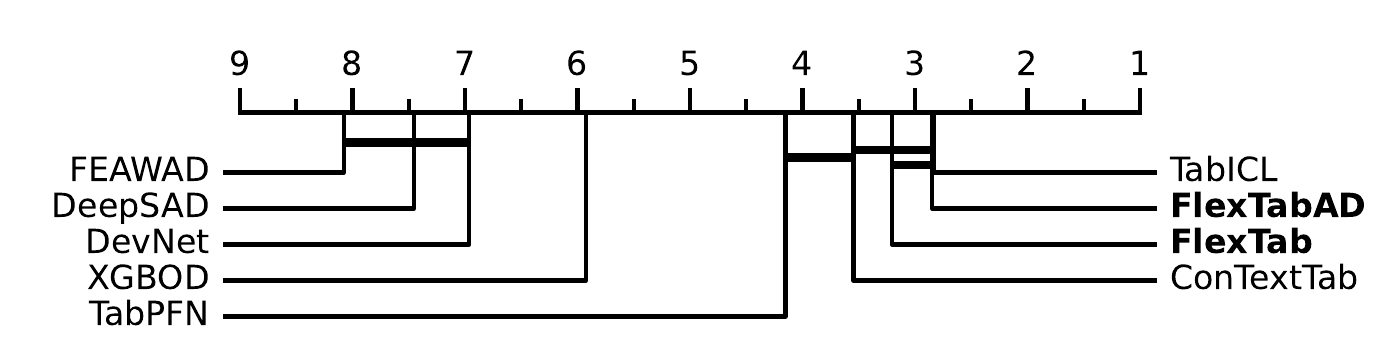}
    \caption{Results for semi-supervised anomaly detection. Left: Average AUROC scores per level of supervision. Right: critical difference diagram (all levels combined) based on \cite{autorank}.}
    \label{fig:semisupervised_outlier_detection}
\end{figure*}

\custompar{Semi-supervised}
In the semi-supervised setting, some of the anomalies are labeled as such, while all the other anomalies are labeled in the same way as normal points. Therefore, this setting can also be thought of as noisy binary classification. Looking at the original ADBench evaluation paper \cite{han2022adbench}, the best semi-supervised method turned out to be XGBOD. In fact, since our metric of choice (AUROC) only requires the model to rank the points according to their likelihood of being anomalies, any classification algorithm can be used for this task; XGBOD itself is only a small addition on top of a traditional XGBoost classifier. For this reason, to provide a challenging baseline, we evaluate against state-of-the-art (classification) tabular foundation models that are able to pick up the signal starting from very few labeled anomalies. As already mentioned in \cref{tab:outlier_detection_semisup} and further detailed in \cref{fig:semisupervised_outlier_detection}, TabICLv2 in particular performs well with just 5\% or 10\% of the points being labeled, as does our FlexTab model trained for classification and regression (with TabPFNv2 and ConTextTab following closely). The FlexTab-AD model trained for anomaly detection has a lead in the nearly-unsupervised setup of just 1\% of labeled anomalies, but falls slightly behind as the setting approaches full supervision (which is just a standard, if imbalanced, binary classification task).

\subsection{Clustering}\label{sec:clustering_results}

To evaluate FlexTab for clustering, we use the same codebase and data as TabClustPFN \cite{tabclustpfn}. This consists of a total of 44 datasets of small size (at most 1000 rows, 64 features, and 10 clusters), manually collected from UCI and OpenML sources (22 each). We evaluate against the same baselines presented there (most of which are based on the scikit-learn implementation~\cite{sklearn}, with the exception of ZEUS~\cite{marszalek2025zeus} for which we use the official implementation), as well as TabClustPFN itself. The metric of choice is ARI.

We remark that clustering evaluation in general, and this evaluation benchmark specifically, is very noisy. Being totally unsupervised, with (classification) labels withheld from the model, it is entirely possible that multiple valid classification tasks could be defined over the same features, and the evaluated metric would change dramatically depending on which one is picked by the model. For this reason, we chose not to present these results in the main body of the paper, because multiple trainings differing in apparently minor details could lead to significantly different results. Similarly, when retraining TabClustPFN with the officially provided training code, results oscillate significantly (though to a lesser extent than when training FlexTab; probably due to synthetic data being more regular than real training data) between consecutive checkpoints.

For the same reason, the choice of validation set and checkpoint can affect results significantly. We chose to select classification tasks from various benchmarks (TabArena, CC18, TALENT-Tiny) with the only filter that, after applying skrub's TableVectorizer~\cite{skrub} to handle non-numeric fields, the k-means algorithm resulted in an ARI score of at least $0.1$. After removing duplicates and any dataset also appearing in the TabClustPFN evaluation set, this resulted in a total of 38 tasks. The best checkpoint based on median ARI was selected after 15.6M steps.

\begin{figure*}
    \centering
    \includegraphics[width=.48\linewidth, valign=c]{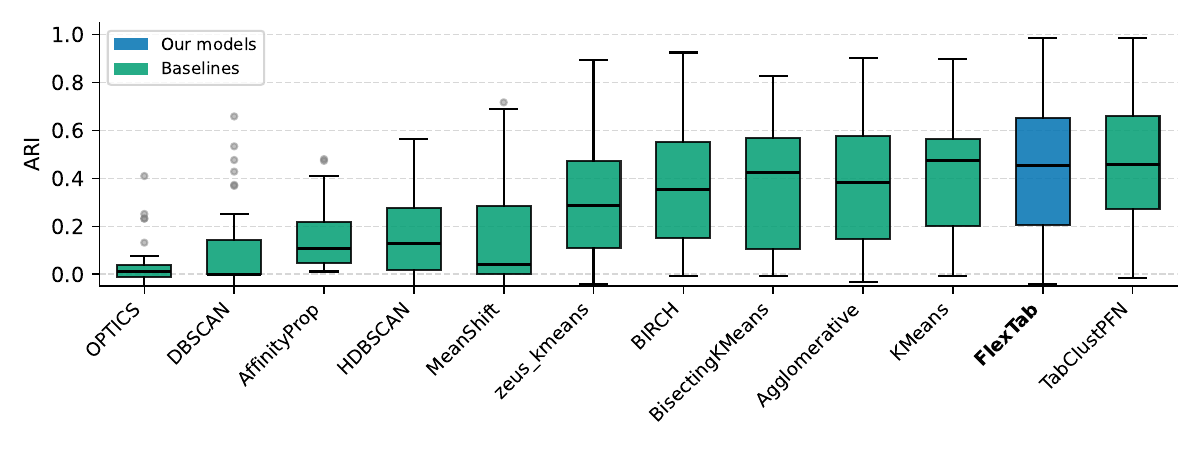}
    \includegraphics[width=.48\linewidth, valign=c]{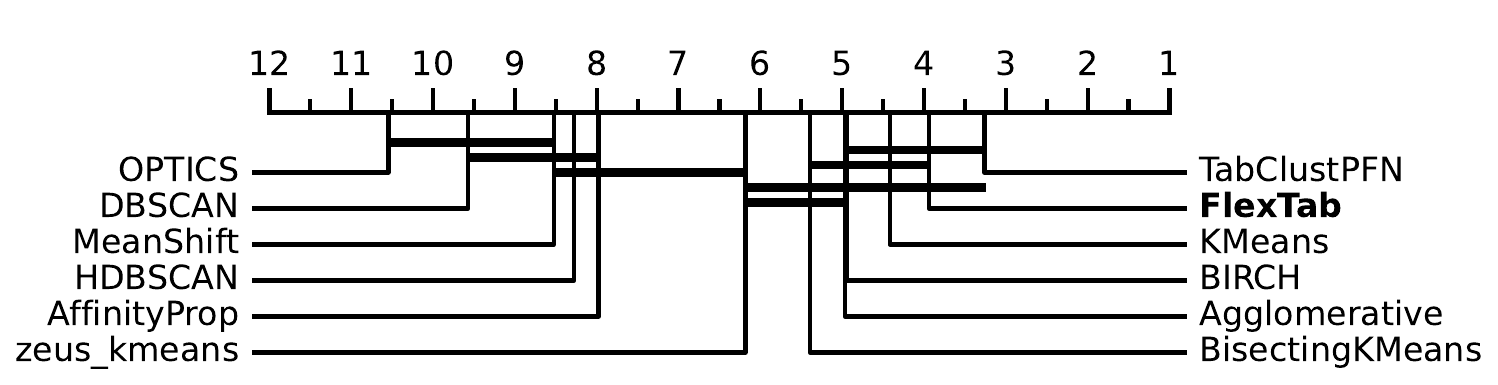}
    \caption{Results for clustering. Left: box and whisker plots, sorted by average score. Right: critical difference diagrams.}
    \label{fig:clustering}
\end{figure*}

The results are reported in \cref{tab:clustering} (separately for UCI and OpenML) and \cref{fig:clustering} (jointly). Overall, TabClustPFN shows a small edge in both rank and average, although differences are minimal, not uniform across datasets, and not statistically significant. In fact, several methods, including at least FlexTab and k-means, end up being equivalent, and these results suggest that the clustering task, at least on this benchmark, does not yet clearly benefit from foundation models. FlexTab does have the advantage of being natively able to support non-numeric data; however, this kind of data is absent in the clustering datasets evaluated here.

\input{tables/clustering}

\subsection{Matching}
\label{app:add-results-matching}
We provide additional results in \cref{tab:matching_f1_extended}.
These include TabPFN and TabICL combined with TF-IDF features obtained for high-cardinality columns via Skrub's \texttt{TableVectorizer}, as well as the complete model list of the DeepMatcher evaluation~\cite{mudgal2018deepmatcher}. 
Finally, we also report results using the \texttt{TableVectorizer} embeddings directly for matching.

Overall, the results are in line with the distilled ones presented in the main paper.
The matching-native FlexTab-Match performs best across all models considered---both in terms of average F1 and rank.
Notably, due to the increased number of evaluated baselines, the rank of the DITTO model has worsened slightly, falling behind our approach.

Lastly, we provide pretraining strategy ablation results in \cref{tab:matching_ablation}.
We see that, depending on the task, the individual task generation strategies yield different outcomes, with no clear superior strategy. 
This underlines the different matching notions inherent to these datasets, ranging from simple exact (or fuzzy) column value matching, to a more nuanced semantic matching.
Combining the two pretraining strategies, equally weighted, yields a hybrid approach that outperforms the corresponding individual ones on almost all datasets considered, and ranks best overall by a clear margin.

\input{tables/matching_ablation}

\input{tables/matching_extended}

\subsection{Relational}
\label{app:add-results-relational}

\begin{figure*}
    \centering
    \begin{subfigure}[t]{0.45\linewidth}
        \centering
        \includegraphics[width=\linewidth]{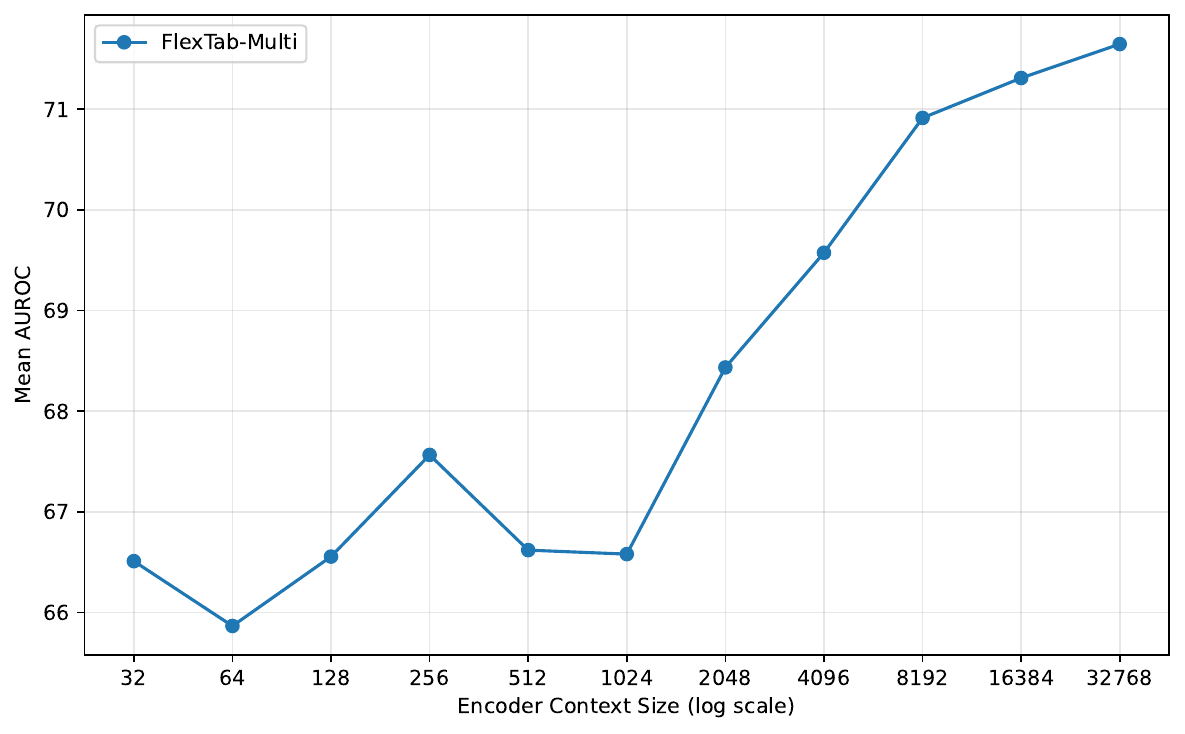}
    \end{subfigure}
    \hspace{1em}
    \begin{subfigure}[t]{0.45\linewidth}
        \centering
        \includegraphics[width=\linewidth]{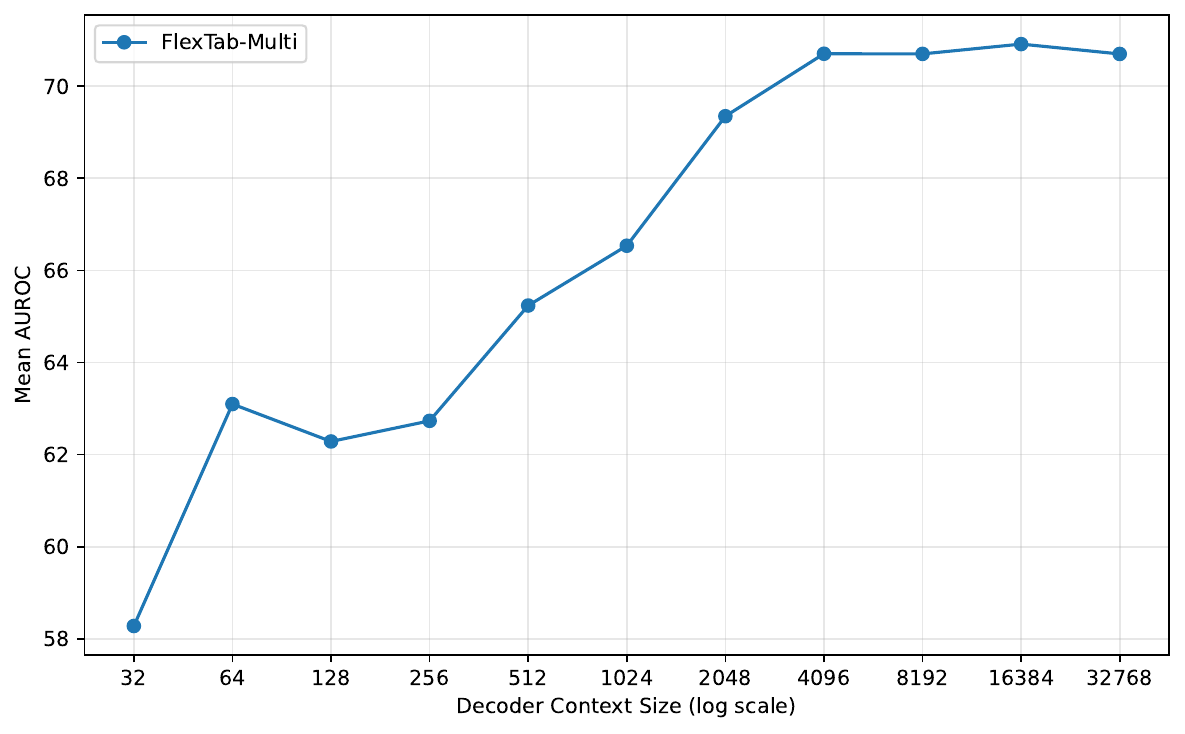}
    \end{subfigure}
    \\[1ex]
    \begin{subfigure}[t]{0.49\linewidth}
        \centering
        \includegraphics[width=0.9\linewidth]{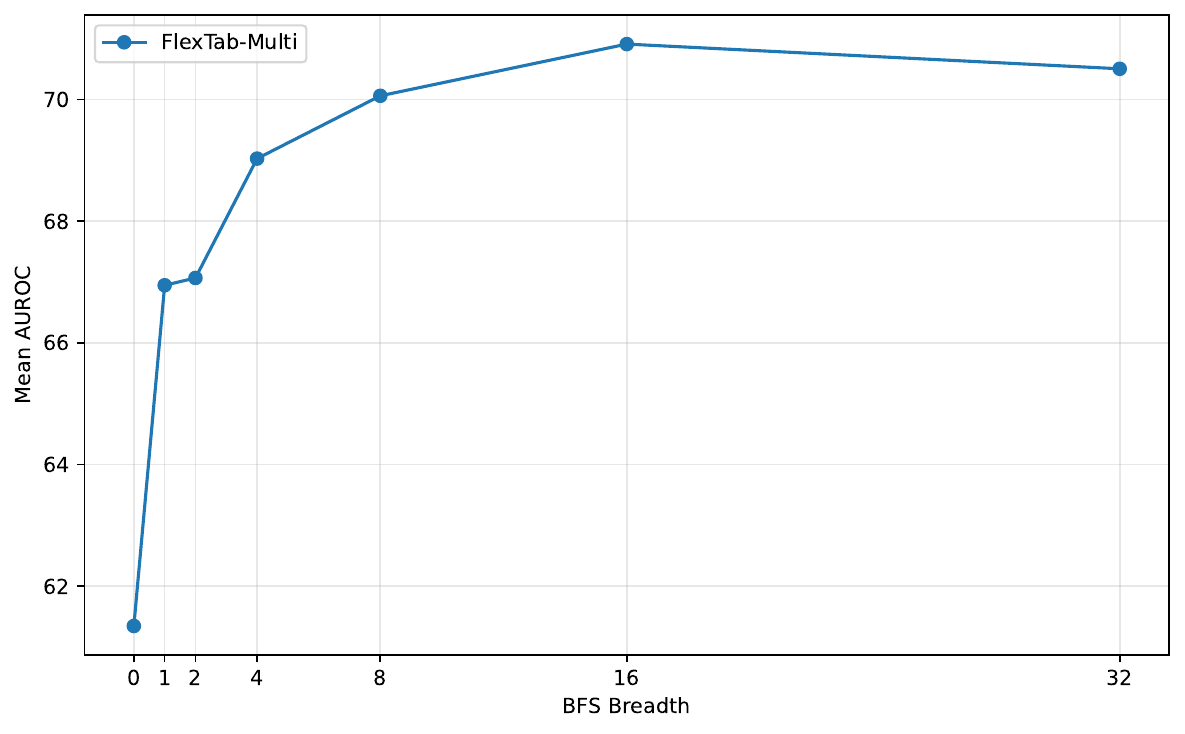}
    \end{subfigure}
    \caption{FlexTab-Multi ablation across 12 RelBench tasks showing how scaling input data affects mean AUROC. (Top left) Effect of encoder context size. (Top right) Effect of decoder context size. (Bottom) Effect of BFS breadth. X-axes in the top left and top right plots are shown in log scale.}
    \label{fig:relational_ablations}
\end{figure*}

To better understand how FlexTab-Multi leverages information from auxiliary tables, we conduct a series of ablation studies across the 12 RelBench classification tasks. We investigate three key factors that control the amount of information available to the model: the encoder context size (number of context rows per table), the decoder context size (number of root-table examples with each example represented by multiple \ROWT tokens), and the BFS breadth (the maximum number of related rows retrieved per hop during breadth-first search). Mean AUROC across the 12 tasks is reported in \cref{fig:relational_ablations}.

\custompar{Encoder context size} As shown in the top-left plot of \cref{fig:relational_ablations}, performance improves as the encoder context size grows from \num{32} to \num{32768} rows while keeping the decoder context size fixed at \num{16384} and BFS breadth fixed at 16. This indicates that the encoder effectively leverages larger intra-table context to produce richer row embeddings. Although we observe gains as the context size is increased, \num{8192} is used as the default value in our experiments to obtain a good trade-off between predictive performance and computational cost, since every table in the schema must be independently embedded by the encoder and the cost of this step grows with the context size.

\custompar{Decoder Context Size} The top-right plot shows a similar monotonic trend as the decoder context is expanded from \num{32} to \num{32768} rows, with mean AUROC rising steadily up to \num{16384} rows. In this experiment, the encoder context size is fixed at \num{8192} and the BFS breadth to 16. This confirms that the decoder can successfully aggregate signal from a growing set of \ROWT embeddings up to a saturation point where additional context does not provide further improvement.

\custompar{BFS Breadth} The bottom plot examines the effect of capping the number of related rows retrieved per hop during BFS across the database. Increasing the BFS breadth from \num{0} (no auxiliary rows, i.e., root-table-only prediction) to \num{16} yields a substantial improvement from \textasciitilde \num{61} AUROC at breadth \num{0} to \textasciitilde \num{71} AUROC at breadth \num{16}. This clearly demonstrates that the FlexTab-Multi decoder effectively leverages the additional \ROWT tokens from auxiliary tables retrieved via BFS, validating our design choice of decoupling per-table encoding from cross-table relational reasoning at the decoder level.

Overall, these ablations confirm that FlexTab-Multi's performance scales favorably with both the intra-table context provided to the encoder and the number of auxiliary rows aggregated by the decoder. Notably, the monotonic improvement with BFS breadth highlights that, despite being pretrained only on chunks derived from single tables, the decoder generalizes to multi-table relational reasoning at inference time.

\subsection{Retrieval}
\label{app:retrieval-results}

To further examine the meaningfulness of embeddings, in~\cref{tab:retrieval-extended-res} we show that retrieval with FlexTab embeddings outperforms random retrieval across various context sizes.

\input{tables/retrieval_extended}

\subsection{Runtime Analysis}
\label{app:runtime-analysis}

\begin{figure*}
    \centering
    \includegraphics[width=1.0\linewidth]{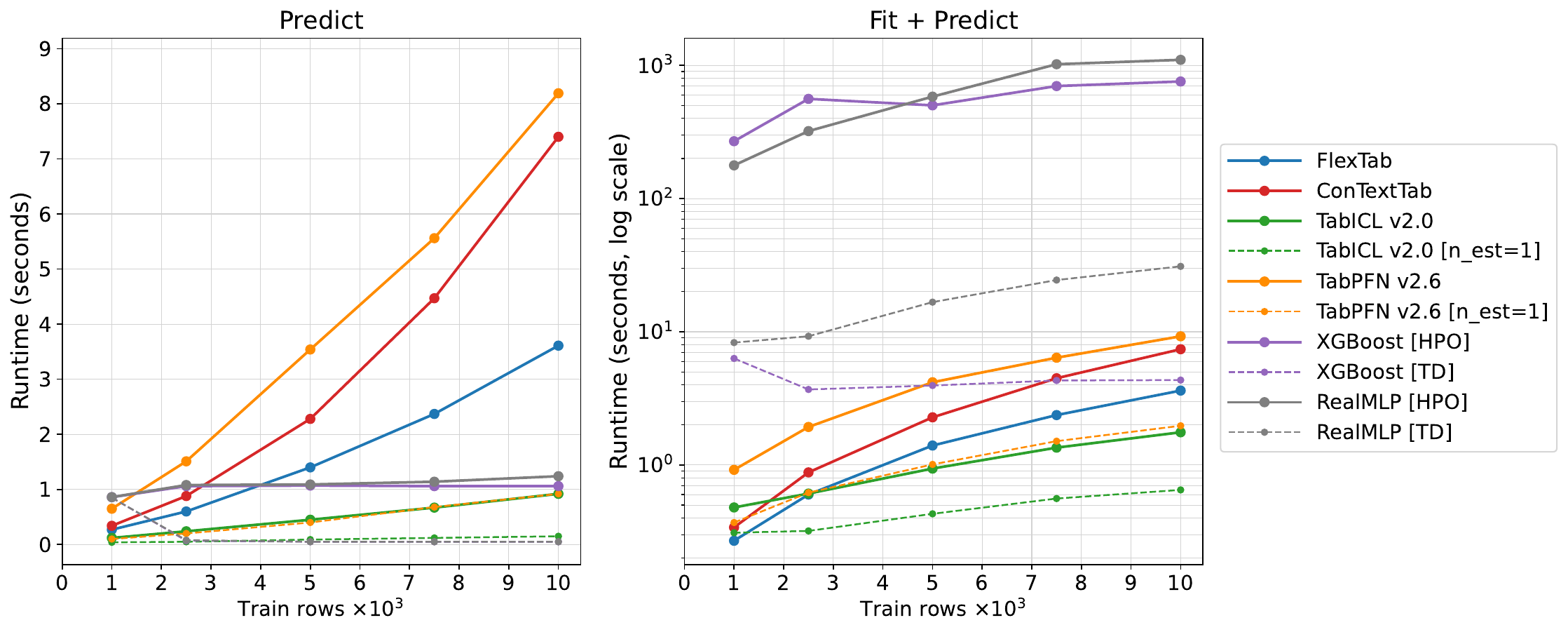}
    \caption{Runtime comparison of different models across varying numbers of training rows. (Left) Runtime for model prediction. (Right) Runtime for model prediction, including fitting. Note the linear y-axis scale on the left side as opposed to the log-scale on the right.}
    \label{fig:runtime_comparison}
\end{figure*}

We conducted a runtime analysis comparing FlexTab with TabPFN, TabICL, ConTextTab, and two trainable baselines, XGBoost and RealMLP, with results presented in \cref{fig:runtime_comparison}. To support this analysis, we generated a synthetic dataset with \num{100} columns split evenly between categorical/textual and numerical features, while varying the number of training/context rows from \num{1000} to \num{10000}. Each test was performed on a single query row at a time across different context sizes, averaging over \num{10} repetitions for each table size. All experiments ran on a compute node equipped with \num{40} CPU cores, \num{320}\,GB of RAM, and an H100 GPU with \num{96}\,GB of VRAM.

Depending on the table size, FlexTab runtimes range from \textasciitilde \num{0.1}s to \textasciitilde \num{4}s. Our findings indicate that FlexTab is approximately twice as fast as TabPFNv2.6, which is mostly attributable to TabPFNv2.6 using \num{24} layers with \num{8} estimators, while our model uses a \num{12}-layer encoder with a larger hidden dimension of \num{768} and bagging of \num{1}. We also achieve a \textasciitilde 2× speedup over ConTextTab for a larger context size of \num{10}k, which is a consequence of our use of flash attention, whereas ConTextTab relies on memory-efficient attention combined with a custom attention mask. As a result, ConTextTab exhibits quadratic scaling, while this trend is far less pronounced for the other models. TabICL is clearly the fastest ICL model in this comparison, as it uses a different architecture. Our experiment reveals that most of FlexTab's runtime is spent in the encoder, which processes all cells in the table, whereas the decoder is notably faster since it operates only on \ROWT tokens and the target column.

We additionally report runtimes for TabPFN and TabICL using a single estimator instead of the default of \num{8}. While this yields an almost linear reduction in runtime, we cannot guarantee comparable predictive performance in this configuration. In contrast, FlexTab employs bagging with 8 estimators only for larger contexts exceeding \num{8192} rows, and in this experiment, for none of the reported runtimes. Inference runtime could be further reduced through complementary techniques such as KV caching or model compilation.

Note that, while non-ICL models offer significantly faster prediction times, they require extensive training. In particular, the hyperparameter-optimized (HPO) versions of XGBoost and RealMLP exhibit fit+predict times that are two orders of magnitude longer.

\subsection{Pretraining Ablations}

In this section we briefly list some experiments we ran to evaluate the impact of our curriculum-learning approach against more direct approaches that do not reuse a frozen encoder pre-trained for classification and regression tasks.

\textbf{No encoder} As a first experiment, we replace the encoder part of the architecture by a simple average of the tokens in each row. Therefore, only the embedding and decoder layers (cross-attending to the average of the embeddings) are maintained. This training run was quickly aborted: after around 4 million micro-steps (15 thousand optimizer steps) the training collapsed and was unable to recover.

\textbf{Shuffled weights} As a further sanity check that learned encoder weights are needed, we tried shuffling them (while maintaining embedding and decoder weights frozen), to see if the inductive bias of the architecture, plus a learned decoder, could be enough to retrieve some prediction quality. Unsurprisingly, the answer is negative, with the overall model collapsing to random-level predictions (or worse).

\textbf{Shuffled weights and fine-tune} In the same setup of the previous point, we then tried fine-tuning the decoder, to see if the performance could be retrieved. This training failed even faster than the first one, with performance collapsing before 1 million micro-steps, never to recover.

%% file: tables/classif_reg_extended.tex
\begin{table}\centering\footnotesize
\caption{Extended classification and regression performance, depicting mean accuracy (Acc) for classification and (soft-clipped) $R^2$ score for regression tasks, in percent, as well as mean per-task rank.}
\label{tab:tabular-benchmarks-extended}
\begin{tabular}{lS[table-format=2.1]S[table-format=2.1]S[table-format=2.1]S[table-format=2.1]S[table-format=2.1]S[table-format=2.1]S[table-format=2.1]S[table-format=2.1]S[table-format=2.1]S[table-format=2.1]S[table-format=2.1]S[table-format=2.1]S[table-format=2.1]}
\toprule
Model Name & \multicolumn{1}{c}{All} & \multicolumn{3}{c}{CARTE} & \multicolumn{3}{c}{TabArena-Lite} & \multicolumn{3}{c}{TALENT-Tiny} & \multicolumn{3}{c}{TextTab} \\
  \cmidrule(lr){2-2} \cmidrule(lr){3-5} \cmidrule(lr){6-8} \cmidrule(lr){9-11} \cmidrule(lr){12-14}
 & {Rank} & {Rank} & {Acc} & {R2} & {Rank} & {Acc} & {R2} & {Rank} & {Acc} & {R2} & {Rank} & {Acc} & {R2} \\
\midrule
AutoGluon & \NAN & \NAN & 78.8 & 74.0 & \NAN & 88.6 & 80.7 & \NAN & 88.9 & 84.7 & \NAN & 83.5 & 67.0 \\\arrayrulecolor{mygrey}\midrule\arrayrulecolor{black}
\bfseries FlexTab [bag=8] & \bfseries 4.5 & \bfseries 2.0 & \bfseries 77.3 & \bfseries 74.4 & 6.1 & 87.1 & 79.5 & 5.5 & 88.0 & \bfseries 85.8 & 5.3 & 83.8 & 62.7 \\
\bfseries FlexTab [bag=1] & 5.2 & 3.0 & 77.0 & 73.6 & 6.8 & 87.0 & 79.1 & 5.9 & 87.9 & 85.5 & 5.9 & 83.7 & 61.4 \\
RealMLP [HPO, CV=5] & 5.3 & 5.7 & 74.3 & 69.0 & 5.0 & 88.3 & 79.9 & 5.2 & 88.6 & 85.0 & 5.0 & 82.3 & \bfseries 67.7 \\
LimiX-16M [etrieval] & 5.8 & 6.2 & 75.5 & 69.5 & 5.9 & 86.8 & 78.1 & 5.9 & \bfseries 88.9 & 84.4 & \bfseries 4.4 & 83.8 & 65.0 \\
ConTextTab & 6.1 & 4.0 & 77.1 & 72.4 & 7.3 & 87.6 & 77.8 & 7.1 & 87.6 & 83.2 & 6.4 & \bfseries 84.4 & 58.8 \\
CatBoost [HPO, CV=5] & 6.1 & 6.7 & 76.3 & 68.3 & 5.7 & 88.2 & 79.0 & 6.4 & 87.1 & 83.5 & 5.0 & 83.7 & 65.4 \\
TabPFNv2.6 & 6.6 & 12.6 & 70.7 & 59.9 & \bfseries 2.5 & 88.7 & \bfseries 80.5 & 3.5 & 88.4 & 84.9 & 7.2 & 81.5 & 63.8 \\
TabICLv2 & 6.9 & 13.3 & 70.5 & 55.3 & 3.0 & \bfseries 88.7 & 79.9 & \bfseries 3.2 & 88.3 & 85.3 & 7.2 & 82.5 & 60.2 \\
LightGBM [HPO, CV=5] & 7.1 & 7.7 & 73.6 & 67.4 & 6.6 & 88.0 & 78.7 & 7.5 & 86.9 & 81.5 & 6.3 & 81.5 & 65.5 \\
XGBoost [HPO, CV=5] & 7.7 & 8.6 & 73.5 & 66.9 & 7.4 & 87.9 & 78.8 & 7.4 & 87.0 & 81.1 & 7.1 & 81.4 & 65.4 \\
TabDPT & 8.2 & 9.8 & 73.0 & 65.7 & 5.4 & 88.1 & 80.1 & 8.6 & 87.0 & 82.7 & 10.6 & 80.8 & 53.7 \\
LimiX-2M [retrieval] & 8.2 & 8.5 & 74.9 & 68.2 & 7.5 & 84.4 & 77.5 & 8.0 & 88.1 & 81.5 & 9.8 & 77.0 & 59.4 \\
CatBoost [TD] & 10.3 & 10.5 & 74.7 & 65.9 & 10.3 & 87.6 & 76.4 & 10.3 & 86.2 & 82.3 & 9.4 & 83.3 & 53.2 \\
LightGBM [TD] & 10.4 & 11.5 & 72.7 & 64.9 & 10.0 & 87.4 & 76.8 & 9.4 & 86.4 & 78.3 & 10.5 & 80.9 & 60.8 \\
XGBoost [TD] & 11.6 & 12.8 & 72.3 & 64.6 & 11.0 & 87.4 & 76.6 & 11.4 & 86.3 & 78.0 & 10.3 & 81.1 & 60.4 \\
TabM  & 11.7 & 15.4 & 70.7 & 60.7 & 9.1 & 87.2 & 76.8 & 10.1 & 85.7 & 82.0 & 12.1 & 79.6 & 55.9 \\
Random forest & 11.8 & 13.8 & 71.4 & 63.4 & 10.7 & 87.6 & 76.1 & 11.4 & 85.6 & 78.0 & 10.4 & 79.8 & 60.7 \\
RealMLP [TD] & 12.1 & 15.6 & 70.6 & 59.7 & 10.7 & 87.3 & 75.4 & 9.8 & 86.0 & 79.6 & 11.2 & 80.0 & 57.8 \\
KNN [k=5] & 18.1 & 18.3 & 65.6 & 34.5 & 17.3 & 81.8 & 57.6 & 18.4 & 80.3 & 66.9 & 19.4 & 72.5 & 4.8 \\
Naive & 18.7 & 19.9 & 53.0 & -1.7 & 17.3 & 75.0 & -7.3 & 19.5 & 53.4 & -21.0 & 17.6 & 70.4 & -5.5 \\
\bottomrule
\end{tabular}
\end{table}

%% file: tables/clustering.tex
\begin{table}
\centering
\small
\caption{Results for clustering: Mean rank and average ARI per dataset.}
\label{tab:clustering}
\setlength{\tabcolsep}{3pt}
\begin{tabular}{l
S[table-format=2.2, detect-weight=true]
S[table-format=2.2, detect-weight=true]
S[table-format=2.2, detect-weight=true]
S[table-format=2.2, detect-weight=true]}
\toprule
Model & \multicolumn{2}{c}{UCI} & \multicolumn{2}{c}{OpenML} \\
  \cmidrule(lr){2-3} \cmidrule(lr){4-5}
 & {Rank} & {Avg ARI} & {Rank} & {Avg ARI} \\
\midrule
TabClustPFN & \bfseries 2.91 & 0.44 & \bfseries 3.61 & \bfseries 0.47 \\
\bfseries{FlexTab} & 3.27 & \bfseries 0.47 & 4.61 & 0.44 \\
KMeans & 4.86 & 0.40 & 3.98 & 0.45 \\
BIRCH & 4.30 & 0.40 & 5.59 & 0.34 \\
Agglomerative & 5.30 & 0.38 & 4.64 & 0.38 \\
BisectingKMeans & 5.77 & 0.36 & 5.02 & 0.39 \\
ZEUS + KMeans & 5.91 & 0.36 & 6.45 & 0.26 \\
AffinityProp & 7.95 & 0.15 & 8.00 & 0.13 \\
HDBSCAN & 8.41 & 0.15 & 8.14 & 0.15 \\
MeanShift & 8.84 & 0.15 & 8.20 & 0.17 \\
DBSCAN & 10.02 & 0.08 & 9.11 & 0.10 \\
OPTICS & 10.45 & 0.04 & 10.64 & 0.02 \\
\bottomrule
\end{tabular}
\end{table}

%% file: tables/matching_ablation.tex
\begin{table}[h]
\footnotesize
\centering
\caption{FlexTab-Match ablation: training task generation mix results (\%, positive-class F1). The hybrid task generation and pretraining objective performs best, combining the best of each of the individual ones. Each dataset is evaluated via 5-fold cross-validation.}
\label{tab:matching_ablation}
\begin{tabular}{l S[table-format=3.1, detect-weight]  S[table-format=3.1, detect-weight]  S[table-format=3.1, detect-weight]  S[table-format=3.1, detect-weight]  S[table-format=3.1, detect-weight]  S[table-format=3.1, detect-weight]  S[table-format=2.1, detect-weight]}
\toprule
 & \multicolumn{5}{c}{Benchmark} & \multicolumn{2}{c}{Average} \\
\cmidrule(lr){2-6} \cmidrule(lr){7-8}
Model & {Febrl4} & {F.-Zagat} & {Bikes} & {eBooks} & {Movies} & {F1} & {Rk} \\
\midrule
Hidden-Target (100\%) & 92.3 & 58.6 & 85.9 & 60.1 & 49.6 & 69.3 & 3.0 \\
Same-Row-Match (100\%) & 99.2 & 90.3 & \bfseries 86.9 & 84.7 & \bfseries 91.5 & 90.5 & 1.4 \\
\textbf{SRM 50\% + HT 50\%} & \bfseries 99.3 & \bfseries 92.5 & 86.8 & \bfseries 87.3 & 90.2 & \bfseries 91.2 & \bfseries 1.2 \\
\bottomrule
\end{tabular}
\end{table}

%% file: tables/matching_extended.tex
\begin{table}
\footnotesize
\centering
\caption{Matching benchmark F1 scores (\%, positive class) obtained as 5-fold cross-validation averages for each dataset. The average per-group rank is denoted as GRk whereas the average rank across all investigated models is denoted as Rk.}
\label{tab:matching_f1_extended}
\begin{tabular}{l S[table-format=3.1, detect-weight]  S[table-format=3.1, detect-weight]  S[table-format=3.1, detect-weight]  S[table-format=3.1, detect-weight]  S[table-format=3.1, detect-weight]  S[table-format=3.1, detect-weight]  S[table-format=2.1, detect-weight]  S[table-format=2.1, detect-weight]}
\toprule
 & \multicolumn{5}{c}{Benchmark} & \multicolumn{3}{c}{Average} \\
\cmidrule(lr){2-6} \cmidrule(lr){7-9}
Model & {Febrl4} & {F.-Zagats} & {Bikes} & {eBooks} & {Movies} & {F1} & {GRk} & {Rk} \\
\midrule
\multicolumn{9}{l}{\textcolor{gray}{\textit{In-context learners}}} \\
TabPFN & 97.3 & 29.4 & 81.8 & 67.9 & 0.0 & 55.3 & 3.6 & 10.0 \\
TabPFN (TF-IDF) & 94.8 & 0.0 & 79.6 & 67.7 & 31.2 & 54.7 & 5.0 & 11.0 \\
TabICL & 95.0 & 36.9 & 82.4 & 73.1 & 0.0 & 57.5 & 3.2 & 9.6 \\
TabICL (TF-IDF) & 76.0 & 4.7 & 82.4 & 67.4 & 5.9 & 47.3 & 5.0 & 11.4 \\
ConTextTab & 83.2 & 23.2 & 82.3 & 60.3 & 79.0 & 65.6 & 4.2 & 9.6 \\
\textbf{FlexTab (ours)} & 79.8 & 12.7 & 84.3 & 72.6 & 61.0 & 62.1 & 3.8 & 9.4 \\
\textbf{FlexTab-Match (ours)} & \bfseries 99.3 & \bfseries 92.5 & \bfseries 86.8 & \bfseries 87.3 & \bfseries 90.2 & \bfseries 91.2 & \bfseries 1.0 & \bfseries 2.4 \\
\arrayrulecolor{gray}\midrule\arrayrulecolor{black}
\multicolumn{9}{l}{\textcolor{gray}{\textit{Embedding similarity}}} \\
NGramHash Embeddings & \bfseries 99.7 & \bfseries 88.0 & \bfseries 69.8 & \bfseries 88.1 & 22.9 & 73.7 & \bfseries 1.6 & \bfseries 6.4 \\
Skrub Embeddings & 45.0 & 61.4 & 47.6 & 53.1 & 31.4 & 47.7 & 3.4 & 12.6 \\
TabPFN Embeddings & 71.8 & 41.2 & 50.8 & 56.4 & 24.0 & 48.9 & 3.2 & 12.6 \\
\textbf{FlexTab Embeddings (ours)} & 98.5 & 81.5 & 54.4 & 84.9 & \bfseries 78.9 & \bfseries 79.6 & 1.8 & 7.8 \\
\arrayrulecolor{gray}\midrule\arrayrulecolor{black}
\multicolumn{9}{l}{\textcolor{gray}{\textit{Per-dataset trained}}} \\
Deepmatch SIF & 99.5 & 0.0 & 0.0 & 35.6 & 0.0 & 27.0 & 4.2 & 12.4 \\
Deepmatch RNN & 99.9 & 89.5 & 86.0 & 88.1 & 80.2 & 88.8 & 2.2 & 2.8 \\
Deepmatch Attn & 99.9 & 89.5 & 82.1 & 84.0 & 83.4 & 87.8 & 2.6 & 4.0 \\
Deepmatch Hybrid & 99.8 & 87.1 & 87.7 & 88.5 & \bfseries 87.6 & \bfseries 90.2 & 2.0 & 2.6 \\
DITTO & \bfseries 100.0 & \bfseries 94.7 & \bfseries 95.5 & \bfseries 89.5 & 44.9 & 84.9 & \bfseries 1.6 & \bfseries 2.4 \\
\bottomrule
\end{tabular}
\end{table}

%% file: tables/retrieval_extended.tex
\begin{table}
\centering
\footnotesize
\caption{Retrieval influence on classification and regression performance, depicting mean accuracy (Acc) and (soft-clipped) $R^2$ score, in percent, as well as mean per-task rank.}
\label{tab:retrieval-extended-res}
\setlength{\tabcolsep}{3.5pt}
\begin{tabular}{lS[table-format=2.1]S[table-format=2.1]S[table-format=2.1]S[table-format=2.1]S[table-format=2.1]S[table-format=2.1]S[table-format=2.1]S[table-format=2.1]S[table-format=2.1]S[table-format=2.1]S[table-format=2.1]S[table-format=2.1]S[table-format=2.1]}
\toprule
Model Name & \multicolumn{1}{c}{All} & \multicolumn{3}{c}{CARTE} & \multicolumn{3}{c}{TabArena} & \multicolumn{3}{c}{TALENT} & \multicolumn{3}{c}{TextTab} \\
  \cmidrule(lr){2-2} \cmidrule(lr){3-5} \cmidrule(lr){6-8} \cmidrule(lr){9-11} \cmidrule(lr){12-14}
 & {Rank} & {Rank} & {Acc} & {R2} & {Rank} & {Acc} & {R2} & {Rank} & {Acc} & {R2} & {Rank} & {Acc} & {R2} \\
\midrule
FlexTab [k=1024, retrieval] & \bfseries 2.0 & \bfseries 1.9 & \bfseries 77.3 & \bfseries 75.5 & \bfseries 1.7 & \bfseries 87.6 & \bfseries 80.4 & 2.7 & 87.7 & \bfseries 86.3 & \bfseries 1.9 & \bfseries 84.2 & 70.0 \\
FlexTab [k=8192, random] & 2.7 & 2.8 & 77.0 & 73.6 & 2.4 & 87.0 & 79.1 & \bfseries 2.2 & \bfseries 87.9 & 85.5 & 3.5 & 83.7 & 61.4 \\
FlexTab [k=512, retrieval] & 3.0 & 2.8 & 77.0 & 74.9 & 3.3 & 87.2 & 79.9 & 3.2 & 87.1 & 85.0 & 2.2 & 83.9 & 72.1 \\
FlexTab [k=4096, random] & 3.3 & 4.1 & 76.5 & 72.1 & 2.7 & 87.3 & 78.5 & 2.7 & 87.7 & 85.0 & 3.7 & 83.5 & 60.0 \\
FlexTab [k=256, retrieval] & 3.7 & 4.0 & 77.0 & 74.1 & 3.5 & 87.4 & 79.6 & 3.7 & 86.7 & 84.4 & 3.8 & 83.4 & \bfseries 72.7 \\
FlexTab [k=2048, random] & 4.6 & 5.9 & 75.7 & 69.8 & 3.8 & 87.1 & 76.9 & 4.1 & 86.5 & 84.2 & 4.3 & 82.6 & 59.3 \\
FlexTab [k=128, retrieval] & 4.9 & 5.2 & 76.4 & 72.9 & 4.6 & 87.3 & 78.6 & 5.2 & 86.4 & 82.4 & 4.6 & 82.6 & 71.2 \\
FlexTab [k=1024, random] & 6.0 & 7.5 & 74.3 & 66.8 & 4.7 & 86.7 & 75.0 & 6.0 & 85.4 & 83.2 & 5.5 & 81.3 & 57.1 \\
\bottomrule
\end{tabular}
\end{table}

%% file: main.bbl
\begin{thebibliography}{71}
\expandafter\ifx\csname natexlab\endcsname\relax\def\natexlab#1{#1}\fi

\bibitem[{Arbel et~al.(2025)Arbel, Salinas, and Hutter}]{equitabpfn}
Michael Arbel, David Salinas, and Frank Hutter. 2025.
\newblock {EquiTabPFN}: A target-permutation equivariant prior fitted network.
\newblock In \emph{Advances in Neural Information Processing Systems}.

\bibitem[{Baevski et~al.(2022)Baevski, Hsu, Xu, Babu, Gu, and
  Auli}]{baevski2022data2vec}
Alexei Baevski, Wei-Ning Hsu, Qiantong Xu, Arun Babu, Jiatao Gu, and Michael
  Auli. 2022.
\newblock Data2vec: A general framework for self-supervised learning in speech,
  vision and language.
\newblock In \emph{International Conference on Machine Learning}.

\bibitem[{Balazadeh et~al.(2025)Balazadeh, Kamkari, Thomas, Li, Ma, Cresswell,
  and Krishnan}]{causalpfn}
Vahid Balazadeh, Hamidreza Kamkari, Valentin Thomas, Benson Li, Junwei Ma,
  Jesse~C Cresswell, and Rahul~G Krishnan. 2025.
\newblock {CausalPFN}: Amortized causal effect estimation via in-context
  learning.
\newblock In \emph{Advances in Neural Information Processing Systems}.

\bibitem[{Balestriero and LeCun(2025)}]{balestriero2025lejepa}
Randall Balestriero and Yann LeCun. 2025.
\newblock {LeJEPA}: Provable and scalable self-supervised learning without the
  heuristics.
\newblock \emph{arXiv preprint arXiv:2511.08544}.

\bibitem[{Blanchard et~al.(2010)Blanchard, Lee, and
  Scott}]{JMLR:v11:blanchard10a}
Gilles Blanchard, Gyemin Lee, and Clayton Scott. 2010.
\newblock Semi-supervised novelty detection.
\newblock \emph{Journal of Machine Learning Research}, 11(99).

\bibitem[{Breunig et~al.(2000)Breunig, Kriegel, Ng, and Sander}]{LOF}
Markus~M Breunig, Hans-Peter Kriegel, Raymond~T Ng, and Jörg Sander. 2000.
\newblock {LOF}: Identifying density-based local outliers.
\newblock In \emph{SIGMOD}.

\bibitem[{Cerda et~al.(2018)Cerda, Varoquaux, and K{\'e}gl}]{skrub}
Patricio Cerda, Ga{\"e}l Varoquaux, and Bal{\'a}zs K{\'e}gl. 2018.
\newblock Similarity encoding for learning with dirty categorical variables.
\newblock \emph{Machine Learning}, 107(8).

\bibitem[{Chen and Guestrin(2016)}]{xgboost}
Tianqi Chen and Carlos Guestrin. 2016.
\newblock {XGBoost}: A scalable tree boosting system.
\newblock In \emph{Proceedings of the 22nd ACM SIGKDD International Conference
  on Knowledge Discovery and Data Mining}, KDD '16.

\bibitem[{Chiang et~al.(2024)Chiang, Zheng, Sheng, Angelopoulos, Li, Li, Zhang,
  Zhu, Jordan, Gonzalez et~al.}]{chiang2024chatbot}
Wei-Lin Chiang, Lianmin Zheng, Ying Sheng, Anastasios~Nikolas Angelopoulos,
  Tianle Li, Dacheng Li, Hao Zhang, Banghua Zhu, Michael Jordan, Joseph~E
  Gonzalez, et~al. 2024.
\newblock {Chatbot Arena}: An open platform for evaluating {LLMs} by human
  preference.
\newblock \emph{arXiv preprint arXiv:2403.04132}.

\bibitem[{Christen(2008)}]{febrl}
Peter Christen. 2008.
\newblock Febrl -- an open source data cleaning, deduplication and record
  linkage system with a graphical user interface.
\newblock In \emph{Proceedings of the ACM SIGMOD International Conference on
  Knowledge Discovery and Data Mining}.

\bibitem[{Collet(2024)}]{xxhash_github}
Yann Collet. 2024.
\newblock xxhash: Extremely fast non-cryptographic hash algorithm.
\newblock \url{https://github.com/Cyan4973/xxHash}.
\newblock GitHub repository; accessed 2026-05-04.

\bibitem[{Das et~al.()Das, Doan, G.~C., Gokhale, Konda, Govind, and
  Paulsen}]{magellandata}
Sanjib Das, AnHai Doan, Paul~Suganthan G.~C., Chaitanya Gokhale, Pradap Konda,
  Yash Govind, and Derek Paulsen.
\newblock The magellan data repository.
\newblock \url{https://sites.google.com/site/anhaidgroup/projects/data}.

\bibitem[{Devlin et~al.(2019)Devlin, Chang, Lee, and Toutanova}]{bert}
Jacob Devlin, Ming-Wei Chang, Kenton Lee, and Kristina Toutanova. 2019.
\newblock {BERT}: Pre-training of deep bidirectional transformers for language
  understanding.
\newblock In \emph{Proceedings of the 2019 Conference of the North American
  Chapter of the Association for Computational Linguistics}, pages 4171--4186.

\bibitem[{Dwivedi et~al.(2025)Dwivedi, Jaladi, Shen, L{\'o}pez, Kanatsoulis,
  Puri, Fey, and Leskovec}]{dwivedi2025relationalgraphtransformer}
Vijay~Prakash Dwivedi, Sri Jaladi, Yangyi Shen, Federico L{\'o}pez, Charilaos~I
  Kanatsoulis, Rishi Puri, Matthias Fey, and Jure Leskovec. 2025.
\newblock Relational graph transformer.
\newblock \emph{arXiv preprint arXiv:2505.10960}.

\bibitem[{Eremeev et~al.(2025)Eremeev, Bazhenov, Platonov, Babenko, and
  Prokhorenkova}]{yandexturningtfmsintogfms}
Dmitry Eremeev, Gleb Bazhenov, Oleg Platonov, Artem Babenko, and Liudmila
  Prokhorenkova. 2025.
\newblock Turning tabular foundation models into graph foundation models.
\newblock In \emph{NeurIPS Workshop on New Perspectives in Advancing Graph
  Machine Learning}.

\bibitem[{Eremeev et~al.(2026)Eremeev, Platonov, Bazhenov, Babenko, and
  Prokhorenkova}]{graphpfn}
Dmitry Eremeev, Oleg Platonov, Gleb Bazhenov, Artem Babenko, and Liudmila
  Prokhorenkova. 2026.
\newblock Graphpfn: A prior-data fitted graph foundation model.
\newblock \emph{International Conference on Machine Learning}.

\bibitem[{Erickson et~al.(2020)Erickson, Mueller, Shirkov, Zhang, Larroy, Li,
  and Smola}]{autogluon}
Nick Erickson, Jonas Mueller, Alexander Shirkov, Hang Zhang, Pedro Larroy,
  Mu~Li, and Alexander Smola. 2020.
\newblock {AutoGluon-Tabular}: Robust and accurate {AutoML} for structured
  data.
\newblock \emph{arXiv preprint arXiv:2003.06505}.

\bibitem[{Erickson et~al.(2025)Erickson, Purucker, Tschalzev, Holzm{\"u}ller,
  Desai, Hutter et~al.}]{tabarena}
Nick Erickson, Lennart Purucker, Andrej Tschalzev, David Holzm{\"u}ller,
  Prateek~Mutalik Desai, Frank Hutter, et~al. 2025.
\newblock {TabArena}: A living benchmark for machine learning on tabular data.
\newblock In \emph{Advances in Neural Information Processing Systems}.

\bibitem[{Fey et~al.(2025)Fey, Kocijan, Lopez, Lenssen, and Leskovec}]{kumorfm}
Matthias Fey, Vid Kocijan, Federico Lopez, Jan~Eric Lenssen, and Jure Leskovec.
  2025.
\newblock \href {https://kumo.ai/research/kumo_relational_foundation_model.pdf}
  {{KumoRFM}: A foundation model for in-context learning on relational data.}

\bibitem[{Gardner et~al.(2024)Gardner, Perdomo, and Schmidt}]{tabula8b}
Josh Gardner, Juan~C Perdomo, and Ludwig Schmidt. 2024.
\newblock Large scale transfer learning for tabular data via language modeling.
\newblock volume~37.

\bibitem[{Garnelo et~al.(2018)Garnelo, Rosenbaum, Maddison, Ramalho, Saxton,
  Shanahan, Yee Whye~Teh, and Eslami}]{neuralprocesses}
Marta Garnelo, Dan Rosenbaum, Christopher Maddison, Tiago Ramalho, David
  Saxton, Murray Shanahan, Danilo~Rezende Yee Whye~Teh, and SM~Ali Eslami.
  2018.
\newblock Conditional neural processes.
\newblock In \emph{International Conference on Machine Learning}.

\bibitem[{Gorishniy et~al.(2025)Gorishniy, Kotelnikov, and Babenko}]{tabm}
Yury Gorishniy, Akim Kotelnikov, and Artem Babenko. 2025.
\newblock {TabM}: Advancing tabular deep learning with parameter-efficient
  ensembling.
\newblock In \emph{International Conference on Learning Representations}.

\bibitem[{Han et~al.(2022)Han, Hu, Huang, Jiang, and Zhao}]{han2022adbench}
Songqiao Han, Xiyang Hu, Hailiang Huang, Minqi Jiang, and Yue Zhao. 2022.
\newblock {ADB}ench: Anomaly detection benchmark.
\newblock In \emph{Thirty-sixth Conference on Neural Information Processing
  Systems Datasets and Benchmarks Track}.

\bibitem[{Hayler et~al.(2025)Hayler, Huang, Ceylan, Bronstein, and
  Finkelshtein}]{ofgraphsandtables}
Adrian Hayler, Xingyue Huang, Ismail~Ilkan Ceylan, Michael~M. Bronstein, and
  Ben Finkelshtein. 2025.
\newblock Of graphs and tables: Zero-shot node classification with tabular
  foundation models.
\newblock In \emph{NeurIPS Workshop on New Perspectives in Advancing Graph
  Machine Learning}.

\bibitem[{Herbold(2020)}]{autorank}
Steffen Herbold. 2020.
\newblock \href {https://doi.org/10.21105/joss.02173} {Autorank: A {Python}
  package for automated ranking of classifiers}.
\newblock \emph{Journal of Open Source Software}, 5(48).

\bibitem[{Hollmann et~al.(2023)Hollmann, M{\"u}ller, Eggensperger, and
  Hutter}]{tabpfnv1}
Noah Hollmann, Samuel M{\"u}ller, Katharina Eggensperger, and Frank Hutter.
  2023.
\newblock Tab{PFN}: A transformer that solves small tabular classification
  problems in a second.
\newblock In \emph{The Eleventh International Conference on Learning
  Representations}.

\bibitem[{Hollmann et~al.(2025)Hollmann, M{\"u}ller, Purucker, Krishnakumar,
  K{\"o}rfer, Hoo, Schirrmeister, and Hutter}]{tabpfnv2}
Noah Hollmann, Samuel M{\"u}ller, Lennart Purucker, Arjun Krishnakumar, Max
  K{\"o}rfer, Shi~Bin Hoo, Robin~Tibor Schirrmeister, and Frank Hutter. 2025.
\newblock Accurate predictions on small data with a tabular foundation model.
\newblock \emph{Nature}, 637(8045):319--326.

\bibitem[{Holzm{\"u}ller et~al.(2024)Holzm{\"u}ller, Grinsztajn, and
  Steinwart}]{realmlp}
David Holzm{\"u}ller, L{\'e}o Grinsztajn, and Ingo Steinwart. 2024.
\newblock Better by default: Strong pre-tuned {MLP}s and boosted trees on
  tabular data.
\newblock \emph{Advances in Neural Information Processing Systems}, 37.

\bibitem[{Hoo et~al.(2024)Hoo, Müller, Salinas, and Hutter}]{timeseriespfn}
Shi~Bin Hoo, Samuel Müller, David Salinas, and Frank Hutter. 2024.
\newblock The tabular foundation model {TabPFN} outperforms specialized time
  series forecasting models based on simple features.
\newblock In \emph{NeurIPS Workshop on Time Series in the Age of Large Models}.

\bibitem[{Hudovernik et~al.(2026)Hudovernik, L{\'o}pez, Kocijan, Nitta,
  Lenssen, Leskovec, and Fey}]{kumorfm2}
Valter Hudovernik, Federico L{\'o}pez, Vid Kocijan, Akihiro Nitta, Jan~Eric
  Lenssen, Jure Leskovec, and Matthias Fey. 2026.
\newblock \href {http://arxiv.org/abs/2604.12596} {{KumoRFM-2}: Scaling
  foundation models for relational learning}.

\bibitem[{Ke et~al.(2017)Ke, Meng, Finley, Wang, Chen, Ma, Ye, and
  Liu}]{lightgbm}
Guolin Ke, Qi~Meng, Thomas Finley, Taifeng Wang, Wei Chen, Weidong Ma, Qiwei
  Ye, and Tie-Yan Liu. 2017.
\newblock {LightGBM}: A highly efficient gradient boosting decision tree.
\newblock In \emph{Advances in Neural Information Processing Systems}.

\bibitem[{Kim et~al.(2024)Kim, Grinsztajn, and Varoquaux}]{CARTE}
Myung~Jun Kim, Leo Grinsztajn, and Gael Varoquaux. 2024.
\newblock {CARTE}: Pretraining and transfer for tabular learning.
\newblock In \emph{Forty-first International Conference on Machine Learning}.

\bibitem[{Konda et~al.(2016)Konda, Das, C, Doan, Ardalan, Ballard, Li, Panahi,
  Zhang, Naughton et~al.}]{konda2016magellan}
Pradap Konda, Sanjib Das, Paul Suganthan~G C, AnHai Doan, Adel Ardalan,
  Jeffrey~R Ballard, Han Li, Fatemah Panahi, Haojun Zhang, Jeff Naughton,
  et~al. 2016.
\newblock {Magellan}: Toward building entity matching management systems over
  data science stacks.
\newblock \emph{Proceedings of the VLDB Endowment}, 9(13).

\bibitem[{Li et~al.(2020)Li, Li, Suhara, Doan, and Tan}]{ditto}
Yuliang Li, Jinfeng Li, Yoshihiko Suhara, AnHai Doan, and Wang-Chiew Tan. 2020.
\newblock Deep entity matching with pre-trained language models.
\newblock \emph{Proceedings of the VLDB Endowment}, 14(1):50--60.

\bibitem[{Li et~al.(2022)Li, Zhao, Hu, Botta, Ionescu, and Chen}]{ECOD}
Zheng Li, Yue Zhao, Xiyang Hu, Nicola Botta, Cezar Ionescu, and George Chen.
  2022.
\newblock {ECOD}: Unsupervised outlier detection using empirical cumulative
  distribution functions.
\newblock \emph{IEEE Transactions on Knowledge and Data Engineering}.

\bibitem[{Liu et~al.(2008)Liu, Ting, and Zhou}]{liu2008isolation}
Fei~Tony Liu, Kai~Ming Ting, and Zhi-Hua Zhou. 2008.
\newblock Isolation forest.
\newblock In \emph{IEEE International Conference on Data Mining}.

\bibitem[{Liu et~al.(2019)Liu, Ott, Goyal, Du, Joshi, Chen, Levy, Lewis,
  Zettlemoyer, and Stoyanov}]{RoBERTa}
Yinhan Liu, Myle Ott, Naman Goyal, Jingfei Du, Mandar Joshi, Danqi Chen, Omer
  Levy, Mike Lewis, Luke Zettlemoyer, and Veselin Stoyanov. 2019.
\newblock {RoBERTa}: {A} robustly optimized {BERT} pretraining approach.
\newblock \emph{arXiv preprint arXiv:1907.11692}.

\bibitem[{Ma et~al.(2025{\natexlab{a}})Ma, Shaheen, Labach, Mhedhbi, Hutter,
  Caterini, and Thomas}]{GeneralizationCanEmerge}
Junwei Ma, Nour Shaheen, Alex Labach, Amine Mhedhbi, Frank Hutter, Anthony~L
  Caterini, and Valentin Thomas. 2025{\natexlab{a}}.
\newblock Generalization can emerge in tabular foundation models from a single
  table.
\newblock In \emph{EurIPS Workshop on AI for Tabular Data}.

\bibitem[{Ma et~al.(2025{\natexlab{b}})Ma, Thomas, Hosseinzadeh, Kamkari,
  Labach, Cresswell, Golestan, Yu, Volkovs, and Caterini}]{tabdpt}
Junwei Ma, Valentin Thomas, Rasa Hosseinzadeh, Hamidreza Kamkari, Alex Labach,
  Jesse~C Cresswell, Keyvan Golestan, Guangwei Yu, Maksims Volkovs, and
  Anthony~L Caterini. 2025{\natexlab{b}}.
\newblock {TabDPT}: Scaling tabular foundation models.
\newblock \emph{Advances in Neural Information Processing Systems}.

\bibitem[{Marsza{\l}ek et~al.(2026)Marsza{\l}ek, Ku{\'s}mierczyk, and
  {\'S}mieja}]{marszalek2026tactic}
Patryk Marsza{\l}ek, Tomasz Ku{\'s}mierczyk, and Marek {\'S}mieja. 2026.
\newblock {TACTIC} for navigating the unknown: Tabular anomaly detection via
  in-context inference.
\newblock \emph{arXiv preprint arXiv:2603.14171}.

\bibitem[{Marsza{\l}ek et~al.(2025)Marsza{\l}ek, Ku{\'s}mierczyk,
  Wydma{\'n}ski, Tabor, and {\'S}mieja}]{marszalek2025zeus}
Patryk Marsza{\l}ek, Tomasz Ku{\'s}mierczyk, Witold Wydma{\'n}ski, Jacek Tabor,
  and Marek {\'S}mieja. 2025.
\newblock Zeus: Zero-shot embeddings for unsupervised separation of tabular
  data.
\newblock In \emph{Advances in Neural Information Processing Systems}.

\bibitem[{Meyer et~al.(2025)Meyer, Lachi, Mohammadi, Upendra, Dyer, Li, and
  Palczewski}]{meyer2025relate}
Joe Meyer, Divyansha Lachi, Mahmoud Mohammadi, Roshan~Reddy Upendra, Eva~L
  Dyer, Mark Li, and Tom Palczewski. 2025.
\newblock {R}{E}{L}{A}{T}{E}: A schema-agnostic perceiver encoder for
  multimodal relational graphs.
\newblock \emph{arXiv preprint arXiv:2510.19954}.

\bibitem[{Mr{\'a}z et~al.(2025)Mr{\'a}z, Das, Gupta, Purucker, and
  Hutter}]{texttabbench}
Martin Mr{\'a}z, Breenda Das, Anshul Gupta, Lennart Purucker, and Frank Hutter.
  2025.
\newblock Towards benchmarking foundation models for tabular data with text.
\newblock In \emph{ICML 2025 Workshop on Foundation Models for Structured Data
  (FMSD)}.

\bibitem[{Mudgal et~al.(2018)Mudgal, Li, Rekatsinas, Doan, Park, Krishnan,
  Deep, Arcaute, and Raghavendra}]{mudgal2018deepmatcher}
Sidharth Mudgal, Han Li, Theodoros Rekatsinas, AnHai Doan, Youngchoon Park,
  Ganesh Krishnan, Rohit Deep, Esteban Arcaute, and Vijay Raghavendra. 2018.
\newblock Deep learning for entity matching: A design space exploration.
\newblock In \emph{Proceedings of the ACM SIGMOD International Conference on
  Management of Data}, pages 19--34.

\bibitem[{M{\"u}ller et~al.(2022)M{\"u}ller, Hollmann, Arango, Grabocka, and
  Hutter}]{pfn}
Samuel M{\"u}ller, Noah Hollmann, Sebastian~Pineda Arango, Josif Grabocka, and
  Frank Hutter. 2022.
\newblock Transformers can do {Bayesian} inference.
\newblock In \emph{International Conference on Learning Representations}.

\bibitem[{Pang et~al.(2019)Pang, Shen, and van~den Hengel}]{devnet}
Guansong Pang, Chunhua Shen, and Anton van~den Hengel. 2019.
\newblock Deep anomaly detection with deviation networks.
\newblock In \emph{KDD}, pages 353--362.

\bibitem[{Pedregosa et~al.(2011)Pedregosa, Varoquaux, Gramfort, Michel,
  Thirion, Grisel, Blondel, Prettenhofer, Weiss, Dubourg et~al.}]{sklearn}
Fabian Pedregosa, Ga{\"e}l Varoquaux, Alexandre Gramfort, Vincent Michel,
  Bertrand Thirion, Olivier Grisel, Mathieu Blondel, Peter Prettenhofer, Ron
  Weiss, Vincent Dubourg, et~al. 2011.
\newblock Scikit-learn: Machine learning in python.
\newblock \emph{the Journal of Machine Learning Research}, 12:2825--2830.

\bibitem[{Prokhorenkova et~al.(2018)Prokhorenkova, Gusev, Vorobev, Dorogush,
  and Gulin}]{catboost}
Liudmila Prokhorenkova, Gleb Gusev, Aleksandr Vorobev, Anna~Veronika Dorogush,
  and Andrey Gulin. 2018.
\newblock {CatBoost}: Unbiased boosting with categorical features.
\newblock \emph{Advances in Neural Information Processing Systems}, 31.

\bibitem[{Pugnaloni et~al.(2025)Pugnaloni, Zecchini, Paganelli, Lissandrini,
  Naumann, and Simonini}]{pugnaloni2025armadillo}
Francesco Pugnaloni, Luca Zecchini, Matteo Paganelli, Matteo Lissandrini, Felix
  Naumann, and Giovanni Simonini. 2025.
\newblock Table overlap estimation through graph embeddings.
\newblock \emph{Proceedings of the ACM on Management of Data}, 3(3):1--25.

\bibitem[{Qu et~al.(2025)Qu, Holzm{\"u}ller, Varoquaux, and Le~Morvan}]{tabicl}
Jingang Qu, David Holzm{\"u}ller, Ga{\"e}l Varoquaux, and Marine Le~Morvan.
  2025.
\newblock {TabICL}: A tabular foundation model for in-context learning on large
  data.
\newblock In \emph{International Conference on Machine Learning}.

\bibitem[{Qu et~al.(2026)Qu, Holzm{\"u}ller, Varoquaux, and
  Le~Morvan}]{qu2026tabiclv2}
Jingang Qu, David Holzm{\"u}ller, Ga{\"e}l Varoquaux, and Marine Le~Morvan.
  2026.
\newblock {TabICLv2}: {A} better, faster, scalable, and open tabular foundation
  model.
\newblock In \emph{Proceedings of the 43rd International Conference on Machine
  Learning}.

\bibitem[{Ramaswamy et~al.(2000)Ramaswamy, Rastogi, and
  Shim}]{ramaswamy2000efficient}
Sridhar Ramaswamy, Rajeev Rastogi, and Kyuseok Shim. 2000.
\newblock Efficient algorithms for mining outliers from large data sets.
\newblock In \emph{SIGMOD}, pages 427--438.

\bibitem[{Ranjan et~al.(2026)Ranjan, Hudovernik, Znidar, Kanatsoulis, Upendra,
  Mohammadi, Meyer, Palczewski, Guestrin, and
  Leskovec}]{relational_transformer}
Rishabh Ranjan, Valter Hudovernik, Mark Znidar, Charilaos~I. Kanatsoulis,
  Roshan~Reddy Upendra, Mahmoud Mohammadi, Joe Meyer, Tom Palczewski, Carlos
  Guestrin, and Jure Leskovec. 2026.
\newblock Relational transformer: Toward zero-shot foundation models for
  relational data.
\newblock In \emph{The Fourteenth International Conference on Learning
  Representations}.

\bibitem[{Robertson et~al.(2025)Robertson, Reuter, Guo, Hollmann, Hutter, and
  Sch{\"o}lkopf}]{dopfn}
Jake Robertson, Arik Reuter, Siyuan Guo, Noah Hollmann, Frank Hutter, and
  Bernhard Sch{\"o}lkopf. 2025.
\newblock {Do-PFN}: In-context learning for causal effect estimation.
\newblock In \emph{Advances in Neural Information Processing Systems}.

\bibitem[{Robinson et~al.(2024)Robinson, Ranjan, Hu, Huang, Han, Dobles, Fey,
  Lenssen, Yuan, Zhang et~al.}]{relbench}
Joshua Robinson, Rishabh Ranjan, Weihua Hu, Kexin Huang, Jiaqi Han, Alejandro
  Dobles, Matthias Fey, Jan~E Lenssen, Yiwen Yuan, Zecheng Zhang, et~al. 2024.
\newblock {Relbench}: {A} benchmark for deep learning on relational databases.
\newblock \emph{Advances in Neural Information Processing Systems}, 37.

\bibitem[{Ruff et~al.(2020)Ruff, Vandermeulen, Görnitz, Binder, Müller,
  Müller, and Kloft}]{DeepSAD}
Lukas Ruff, Robert~A. Vandermeulen, Nico Görnitz, Alexander Binder, Emmanuel
  Müller, Klaus-Robert Müller, and Marius Kloft. 2020.
\newblock Deep semi-supervised anomaly detection.
\newblock In \emph{International Conference on Learning Representations}.

\bibitem[{Schiff et~al.(2026)Schiff, Lindenbaum, and Efroni}]{tabpfnRL}
David Schiff, Ofir Lindenbaum, and Yonathan Efroni. 2026.
\newblock {ICR-RL}: Deep reinforcement learning via in-context-regression.
\newblock In \emph{International Conference on Machine Learning}.

\bibitem[{Sch{\"o}lkopf et~al.(1999)Sch{\"o}lkopf, Williamson, Smola,
  Shawe-Taylor, and Platt}]{scholkopf1999support}
Bernhard Sch{\"o}lkopf, Robert~C Williamson, Alex Smola, John Shawe-Taylor, and
  John Platt. 1999.
\newblock Support vector method for novelty detection.
\newblock \emph{Advances in Neural Information Processing Systems}, 12.

\bibitem[{Shen et~al.(2025)Shen, Wen, and Akoglu}]{shen2025fomod}
Yuchen Shen, Haomin Wen, and Leman Akoglu. 2025.
\newblock Fomo-0d: A foundation model for zero-shot tabular outlier detection.
\newblock \emph{Transactions on Machine Learning Research}.

\bibitem[{Shyu et~al.(2003)Shyu, Chen, Sarinnapakorn, and
  Chang}]{shyu2003novel}
Mei-Ling Shyu, Shu-Ching Chen, Kanoksri Sarinnapakorn, and LiWu Chang. 2003.
\newblock A novel anomaly detection scheme based on principal component
  classifier.
\newblock In \emph{Proceedings of the IEEE Foundations and New Directions of
  Data Mining Workshop}.

\bibitem[{Spinaci et~al.(2025)Spinaci, Polewczyk, Schambach, and
  Thelin}]{contexttab}
Marco Spinaci, Marek Polewczyk, Maximilian Schambach, and Sam Thelin. 2025.
\newblock {ConTextTab}: A semantics-aware tabular in-context learner.
\newblock In \emph{Advances in Neural Information Processing Systems}.

\bibitem[{Thomas et~al.(2024)Thomas, Ma, Hosseinzadeh, Golestan, Yu, Volkovs,
  and Caterini}]{localpfn}
Valentin Thomas, Junwei Ma, Rasa Hosseinzadeh, Keyvan Golestan, Guangwei Yu,
  Maksims Volkovs, and Anthony Caterini. 2024.
\newblock Retrieval \& fine-tuning for in-context tabular models.
\newblock In \emph{Advances in Neural Information Processing Systems}.

\bibitem[{Vaswani et~al.(2017)Vaswani, Shazeer, Parmar, Uszkoreit, Jones,
  Gomez, Kaiser, and Polosukhin}]{vaswani2017attention}
Ashish Vaswani, Noam Shazeer, Niki Parmar, Jakob Uszkoreit, Llion Jones,
  Aidan~N Gomez, {\L}ukasz Kaiser, and Illia Polosukhin. 2017.
\newblock Attention is all you need.
\newblock \emph{Advances in Neural Information Processing Systems}, 30.

\bibitem[{Wang et~al.(2025)Wang, Wang, Gan, Wang, Yang, Wipf, and
  Zhang}]{griffin}
Yanbo Wang, Xiyuan Wang, Quan Gan, Minjie Wang, Qibin Yang, David Wipf, and
  Muhan Zhang. 2025.
\newblock Griffin: Towards a graph-centric relational database foundation
  model.
\newblock In \emph{Forty-second International Conference on Machine Learning}.

\bibitem[{Wei and Armanfard(2026)}]{wei2026iclad}
Jack~Yi Wei and Narges Armanfard. 2026.
\newblock {ICLAD}: In-context learning for unified tabular anomaly detection
  across supervision regimes.
\newblock \emph{arXiv preprint arXiv:2603.19497}.

\bibitem[{Xu et~al.(2026)Xu, Zhang, Gan, Wang, and Wipf}]{rdblearn}
Linjie Xu, Yanlin Zhang, Quan Gan, Minjie Wang, and David Wipf. 2026.
\newblock No need to train your {RDB} foundation model.
\newblock In \emph{International Conference on Machine Learning}.

\bibitem[{Ye et~al.(2024)Ye, Liu, Cai, Zhou, and Zhan}]{talent}
Han-Jia Ye, Si-Yang Liu, Hao-Run Cai, Qi-Le Zhou, and De-Chuan Zhan. 2024.
\newblock A closer look at deep learning methods on tabular datasets.
\newblock \emph{arXiv preprint arXiv:2407.00956}.

\bibitem[{Zhang et~al.(2025)Zhang, Maddix, Yin, Erickson, Ansari, Han, Zhang,
  Akoglu, Faloutsos, Mahoney et~al.}]{mitra}
Xiyuan Zhang, Danielle~C Maddix, Junming Yin, Nick Erickson, Abdul~Fatir
  Ansari, Boran Han, Shuai Zhang, Leman Akoglu, Christos Faloutsos, Michael~W
  Mahoney, et~al. 2025.
\newblock Mitra: Mixed synthetic priors for enhancing tabular foundation
  models.
\newblock In \emph{Advances in Neural Information Processing Systems}.

\bibitem[{Zhao et~al.(2026)Zhao, Wang, Tan, and Zhang}]{tabclustpfn}
Tianqi Zhao, Guanyang Wang, Yan~Shuo Tan, and Qiong Zhang. 2026.
\newblock {TabClustPFN}: A prior-fitted network for tabular data clustering.
\newblock \emph{arXiv preprint arXiv:2601.21656}.

\bibitem[{Zhao and Hryniewicki(2018)}]{zhao2018xgbod}
Yue Zhao and Maciej~K Hryniewicki. 2018.
\newblock {XGBOD}: Improving supervised outlier detection with unsupervised
  representation learning.
\newblock In \emph{International Joint Conference on Neural Networks (IJCNN)}.

\bibitem[{Zhou et~al.(2021)Zhou, Song, Zhang, Liu, Zhu, and Liu}]{FEAWAD}
Yingjie Zhou, Xucheng Song, Yanru Zhang, Fanxing Liu, Ce~Zhu, and Lingqiao Liu.
  2021.
\newblock Feature encoding with autoencoders for weakly supervised anomaly
  detection.
\newblock \emph{IEEE Transactions on Neural Networks and Learning Systems},
  33(6).

\end{thebibliography}
